\documentclass[10pt,twocolumn,letterpaper]{article}

\newcommand{\dataset}{RealBokeh\xspace}
\newcommand{\sdataset}{\dataset}
\newcommand{\bindataset}{$\text{\dataset}_{\text{\textit{bin}}}$}

\newcommand{\datasetsizescene}{4400\xspace}

\newcommand{\method}{Bokehlicious\xspace}
\newcommand{\smethod}{\method}

\newcommand{\fnum}[1]{\emph{f/#1}}
\newcommand{\fstop}{\emph{f}-stop\xspace}
\newcommand{\fstops}{\emph{f}-stops\xspace}
\usepackage[pagenumbers]{iccv}

\usepackage{float}
\usepackage{placeins}
\usepackage{multirow}
\usepackage{tikz}
\usepackage{pgfplots}
\usepackage{array, makecell}
\usepackage{mwe}

\usepackage[pagebackref,breaklinks,colorlinks]{hyperref}
\def\confName{ArXiv}
\def\confYear{2025}
\title{Bokehlicious: Photorealistic Bokeh Rendering with Controllable Apertures}

\author{Tim Seizinger, Florin-Alexandru Vasluianu, Marcos V. Conde, Zongwei Wu, Radu Timofte \\
Computer Vision Lab, CAIDAS, University of Wurzburg, Germany\\
}

\begin{document}

\twocolumn[{
\renewcommand\twocolumn[1][]{#1}%
\maketitle
\begin{center}
\setlength{\tabcolsep}{2pt}
\renewcommand{\arraystretch}{0.3}
\def\scalefact{1.0}
\def\widthimga{0.158}
\def\widthimgb{0.192}
\def\widthimgc{0.200}
\def\widthimgd{0.1475}
\begin{tabular}{cccccc}
    \rotatebox{90}{\hspace{10mm}Before} &
        \includegraphics[width=\widthimga\linewidth, trim={0 320px 0 0},clip]{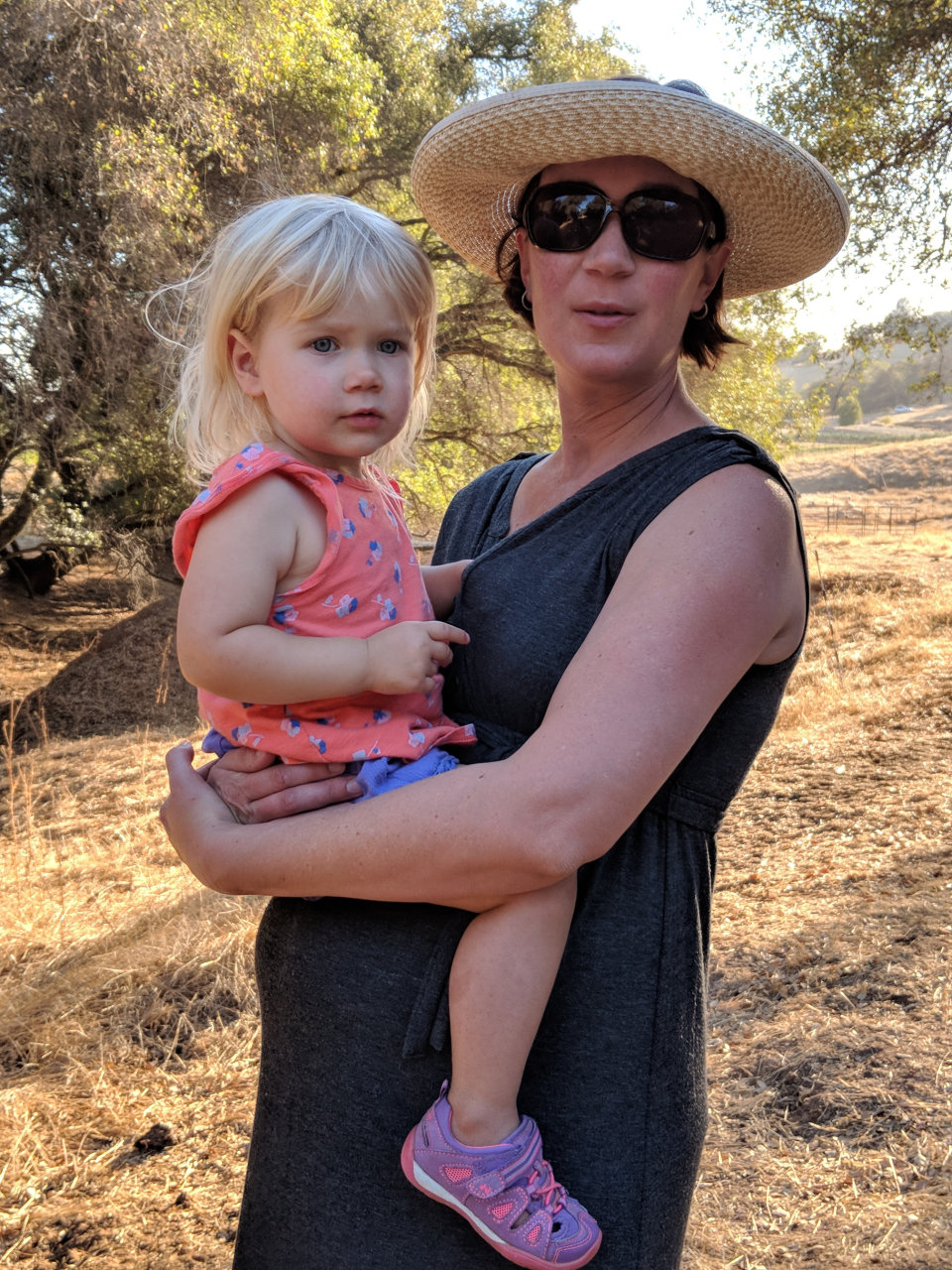} &
        \includegraphics[width=\widthimgc\linewidth]{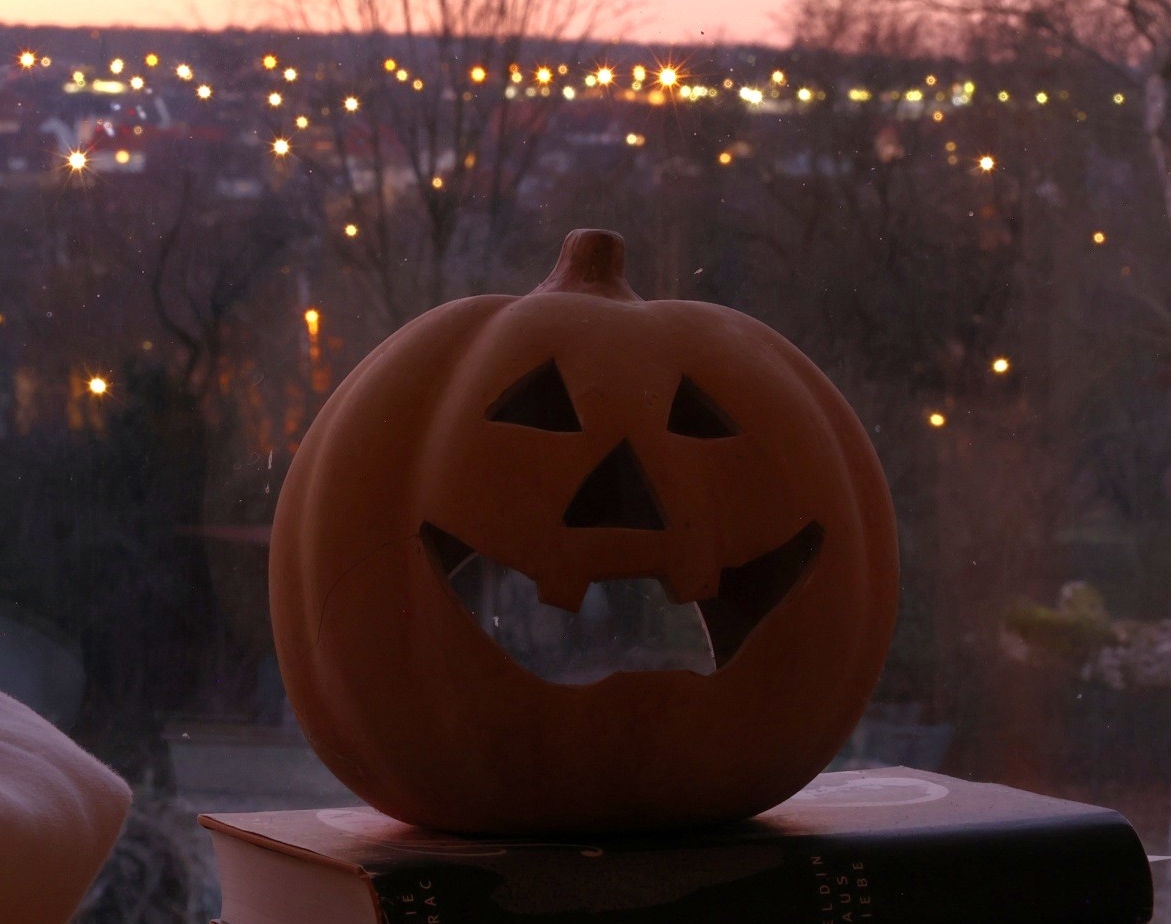} &
        \includegraphics[width=\widthimgb\linewidth, trim={0 0 0 392px},clip]{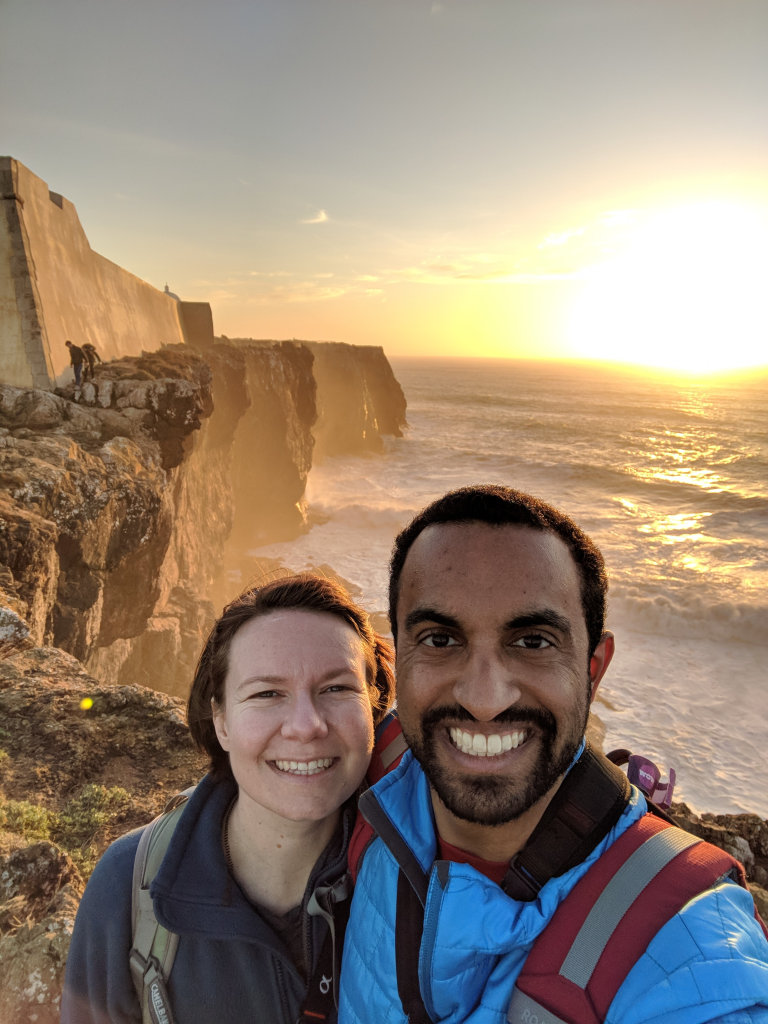} &
        \includegraphics[width=\widthimgb\linewidth, trim={158px 20px 124px 120px},clip]{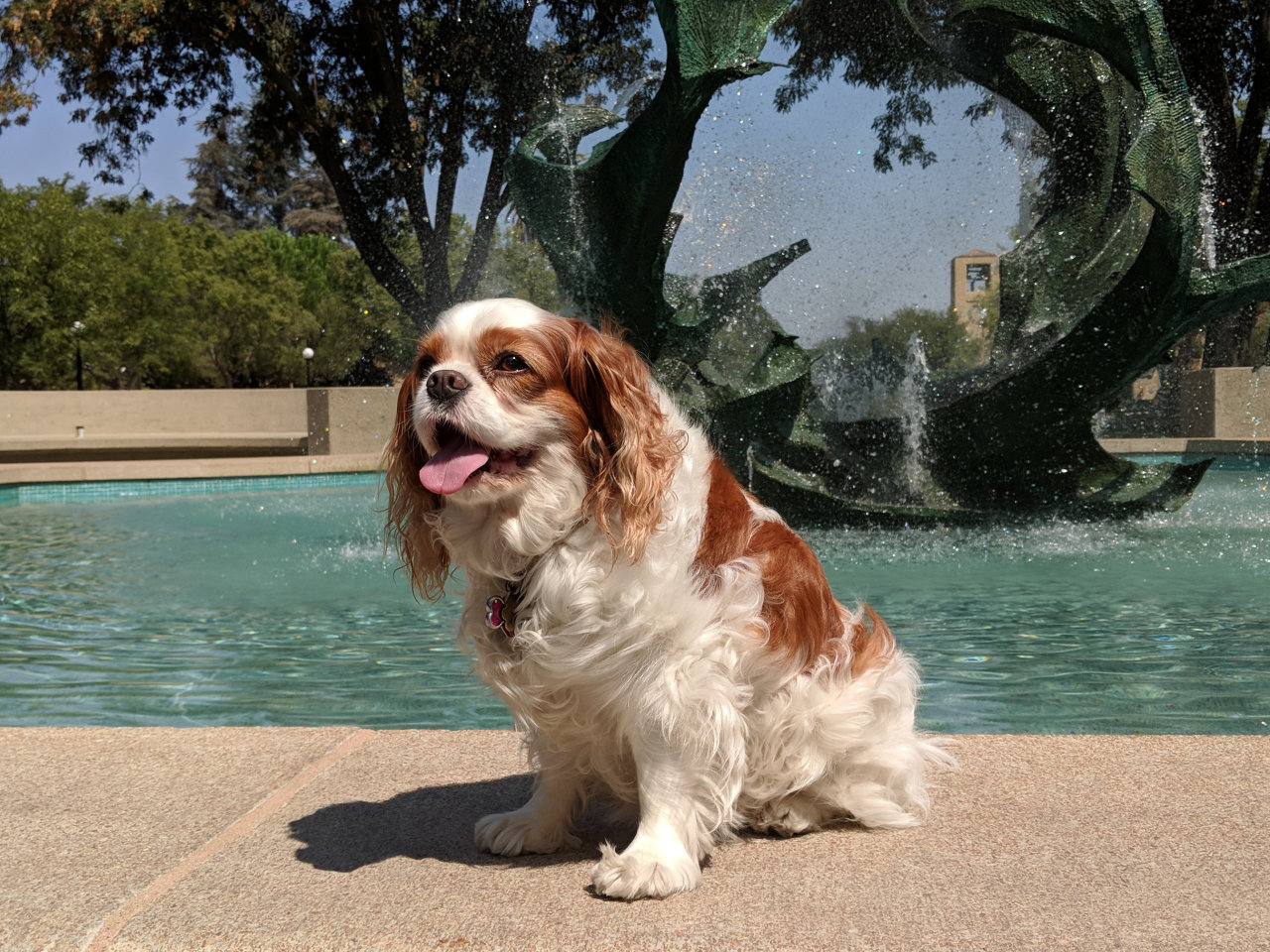} &
        \includegraphics[width=\widthimgd\linewidth, trim={600px 0 0 0},clip]{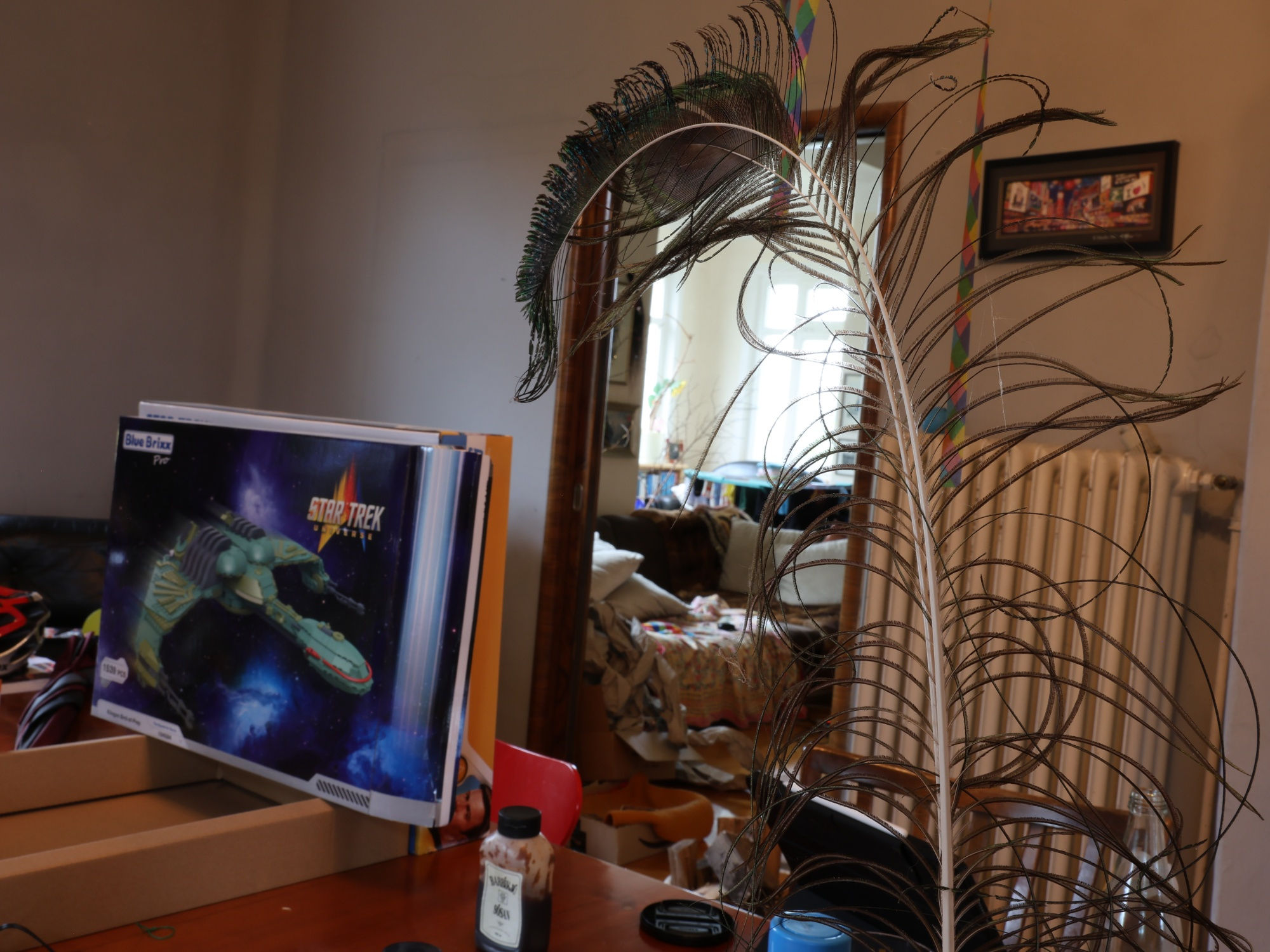}
        \\
        \rotatebox{90}{\hspace{10mm}After} &
        \includegraphics[width=\widthimga\linewidth, trim={0 320px 0 0},clip]{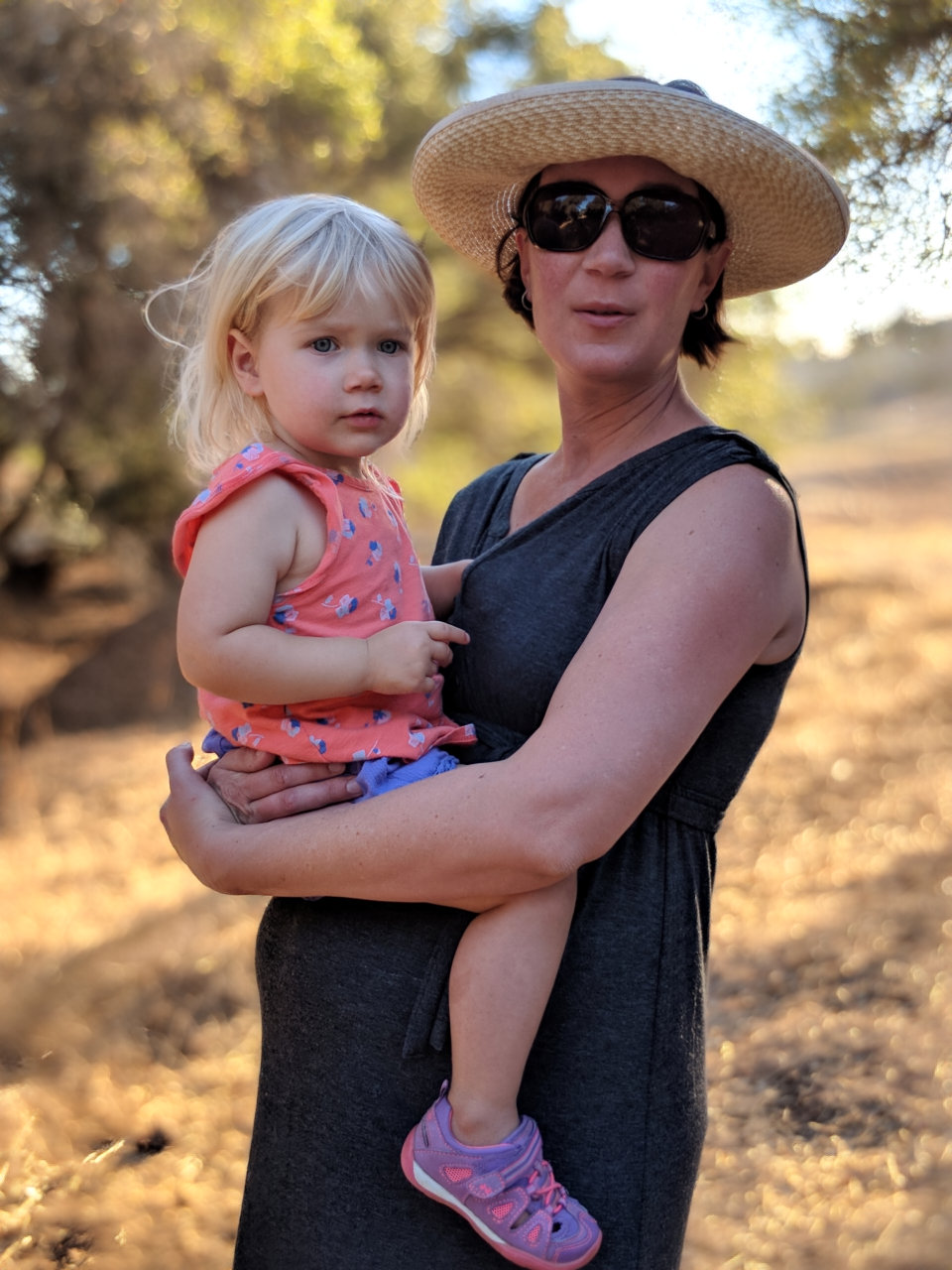} & 
        \includegraphics[width=\widthimgc\linewidth]{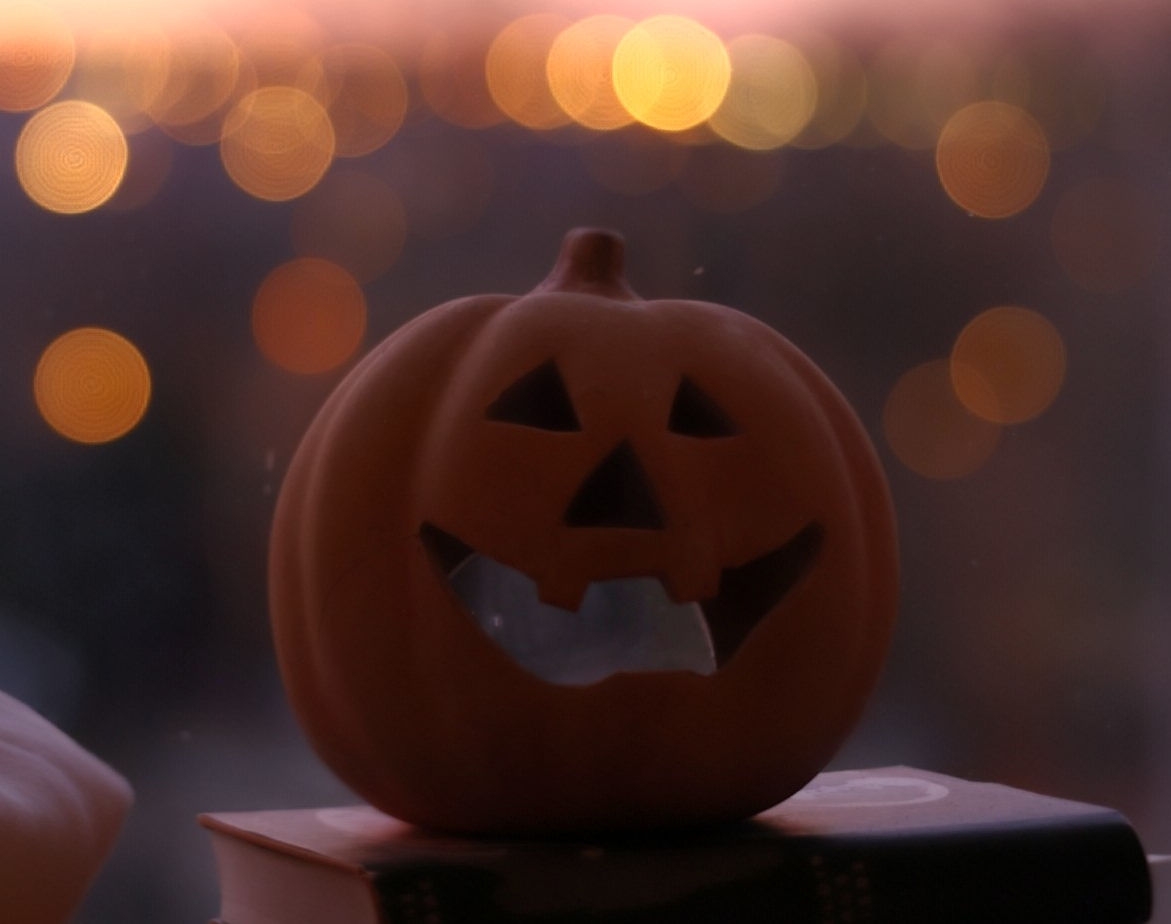} &
        \includegraphics[width=\widthimgb\linewidth, trim={0 0 0 392px},clip]{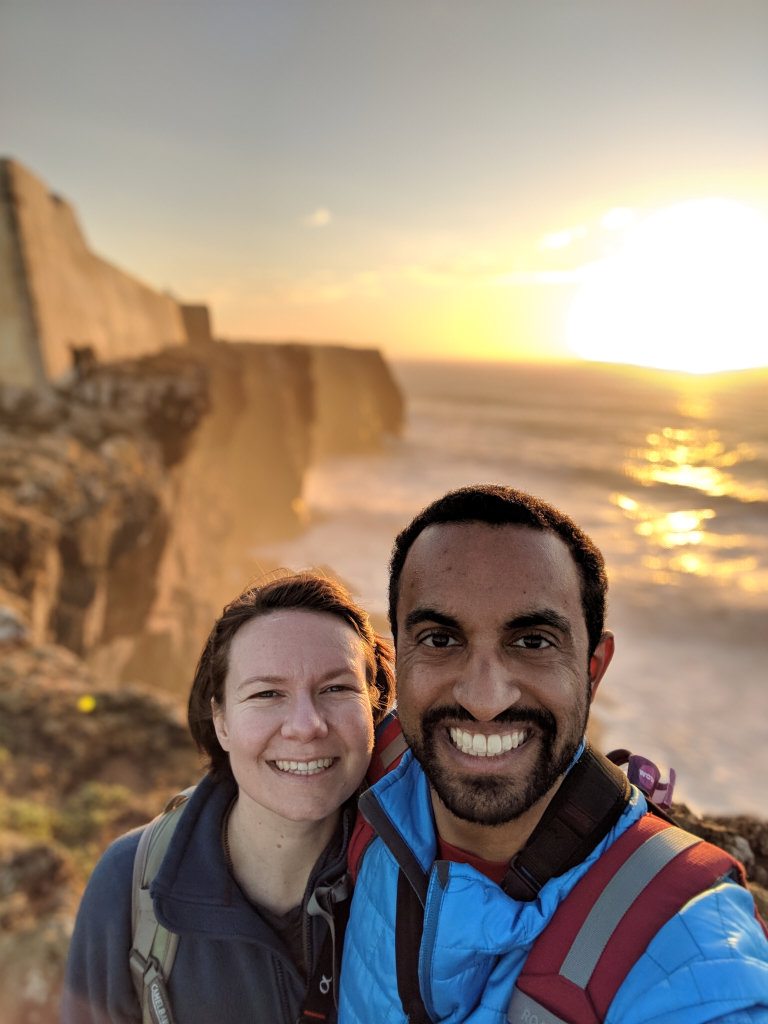} &
        \includegraphics[width=\widthimgb\linewidth, trim={158px 20px 124px 120px},clip]{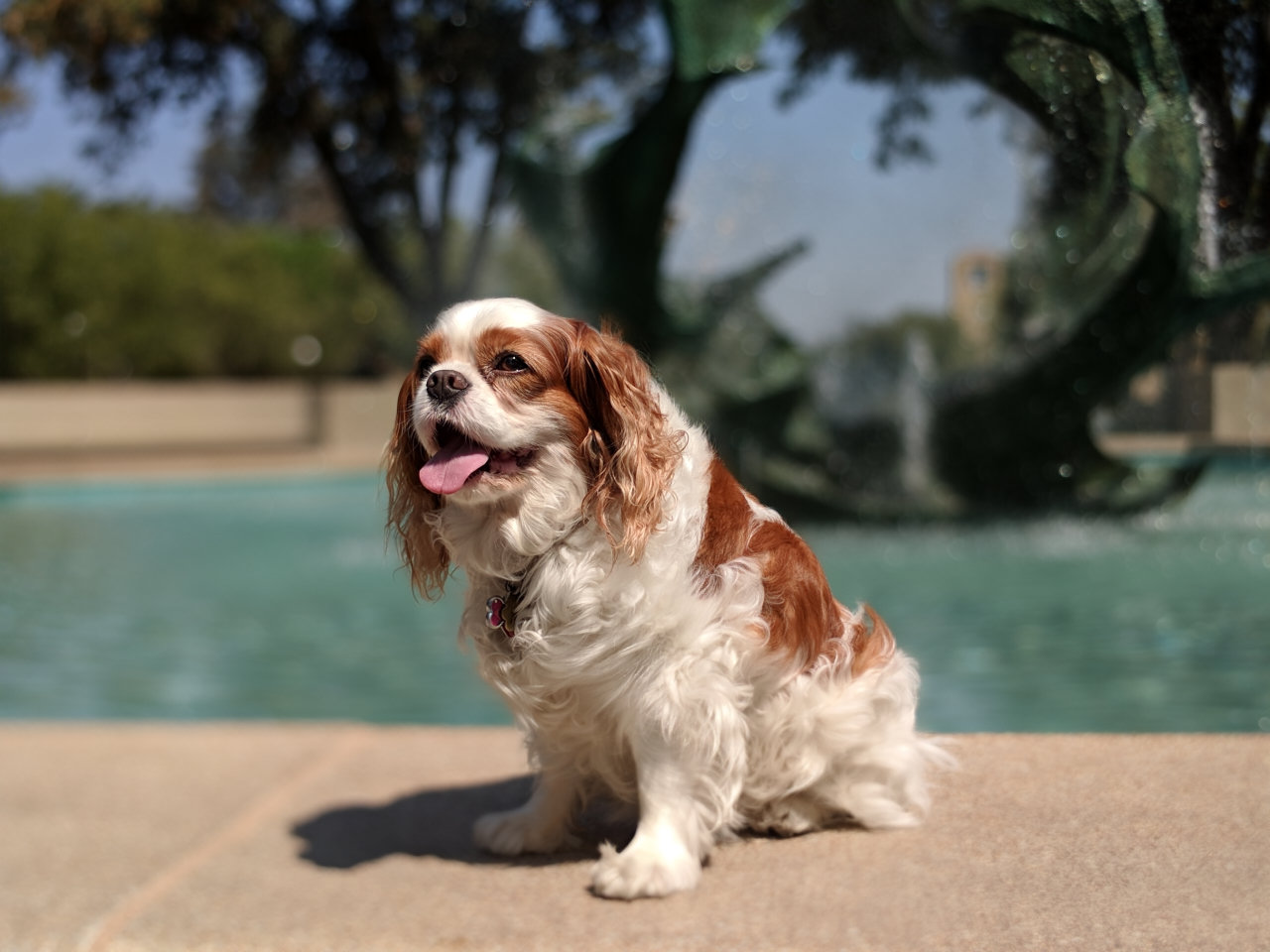} &
        \includegraphics[width=\widthimgd\linewidth, trim={600px 0 0 0},clip]{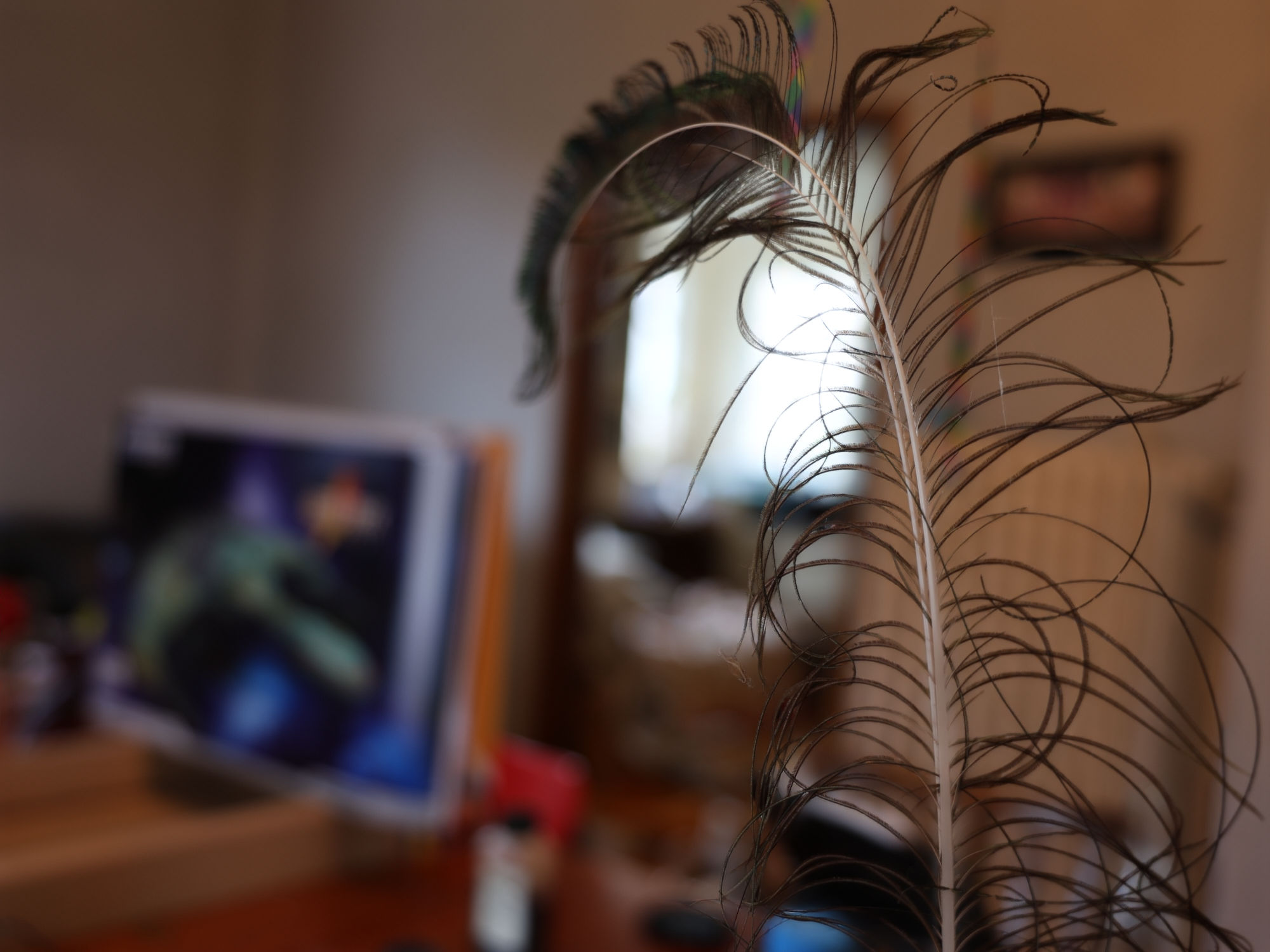}
        \\
\end{tabular}
      
\vspace{-2mm}
\captionof{figure}{Our proposed Bokehlicious architecture trained on our new RealBokeh dataset can create highly photorealistic Bokeh effects of varying intensity, without the need for depth maps or other auxiliary inputs.
Our method is able to maintain difficult foreground details like hair and excels at rendering highly complex Bokeh phenomena while maintaining low computational complexity. Some images from \cite{wadhwa2018synthetic}.}
\label{fig:main_teaser}
\vspace{-1mm}
\end{center}
}]

\begin{abstract}

\vspace{-4mm}

\noindent Bokeh rendering methods play a key role in creating the visually appealing, softly blurred backgrounds seen in professional photography. While recent learning-based approaches show promising results, generating realistic Bokeh with variable strength remains challenging. Existing methods require additional inputs and suffer from unrealistic Bokeh reproduction due to reliance on synthetic data. In this work, we propose Bokehlicious, a highly efficient network that provides intuitive control over Bokeh strength through an Aperture-Aware Attention mechanism, mimicking the physical lens aperture. To further address the lack of high-quality real-world data, we present RealBokeh, a novel dataset featuring \textbf{23,000} high-resolution (\textbf{24-MP}) images captured by professional photographers, covering diverse scenes with varied aperture and focal length settings. Evaluations on both our new RealBokeh and established Bokeh rendering benchmarks show that Bokehlicious consistently outperforms SOTA methods while significantly reducing computational cost and exhibiting strong \textbf{zero-shot} generalization. Our method and dataset further extend to defocus deblurring, achieving competitive results on the RealDOF benchmark. Our code and data can be found at \url{https://github.com/TimSeizinger/Bokehlicious}
\end{abstract}    
\section{Introduction}

\textit{Bokeh}, a term derived from the Japanese word \textit{boke}, is a key aesthetic element in professional photography. 
It emphasizes the subject by transforming potential distractions in the background of the image into a visually pleasing effect. 
Bokeh typically manifests itself as a smooth blur, although in high-contrast areas, a more intricate effect featuring circular shapes may appear~\cite{kennerdell1997bokeh}.

\begin{figure*}[th!]
    \centering
    \small
    \setlength{\tabcolsep}{1pt}
    \renewcommand{\arraystretch}{0.7}
    \def\widthcomp{0.19}
    \begin{tabular}{cccccc}
    & Input \fnum{22.0} & GT \fnum{9.0} & GT \fnum{4.5} & GT \fnum{3.2} & GT \fnum{2.0}  \\
    \addlinespace[0.2pt]
    \rotatebox{90}{\hspace{0.5mm}Focallength 70mm} &
    \includegraphics[width=\widthcomp\linewidth]{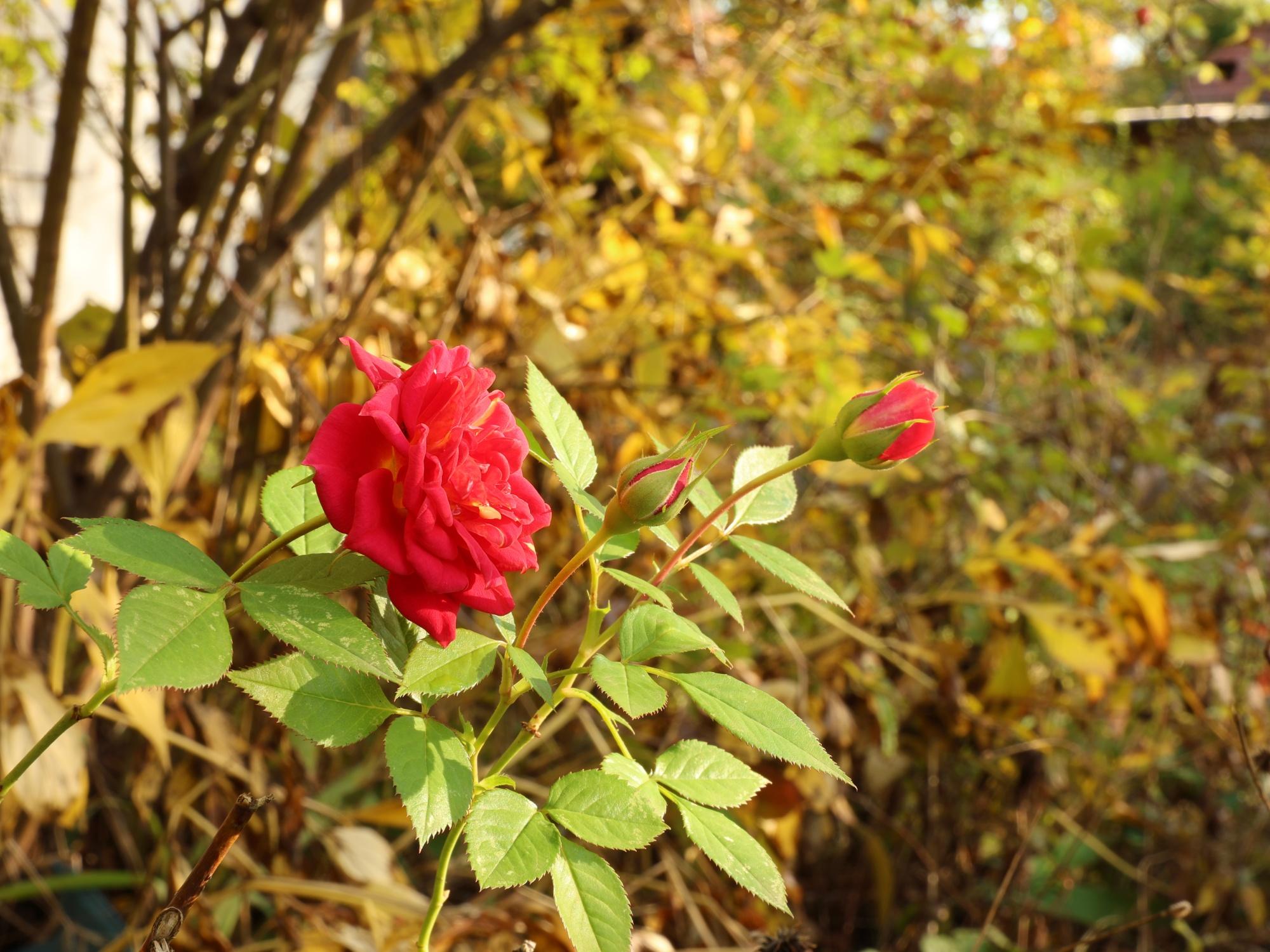} & \includegraphics[width=\widthcomp\linewidth]{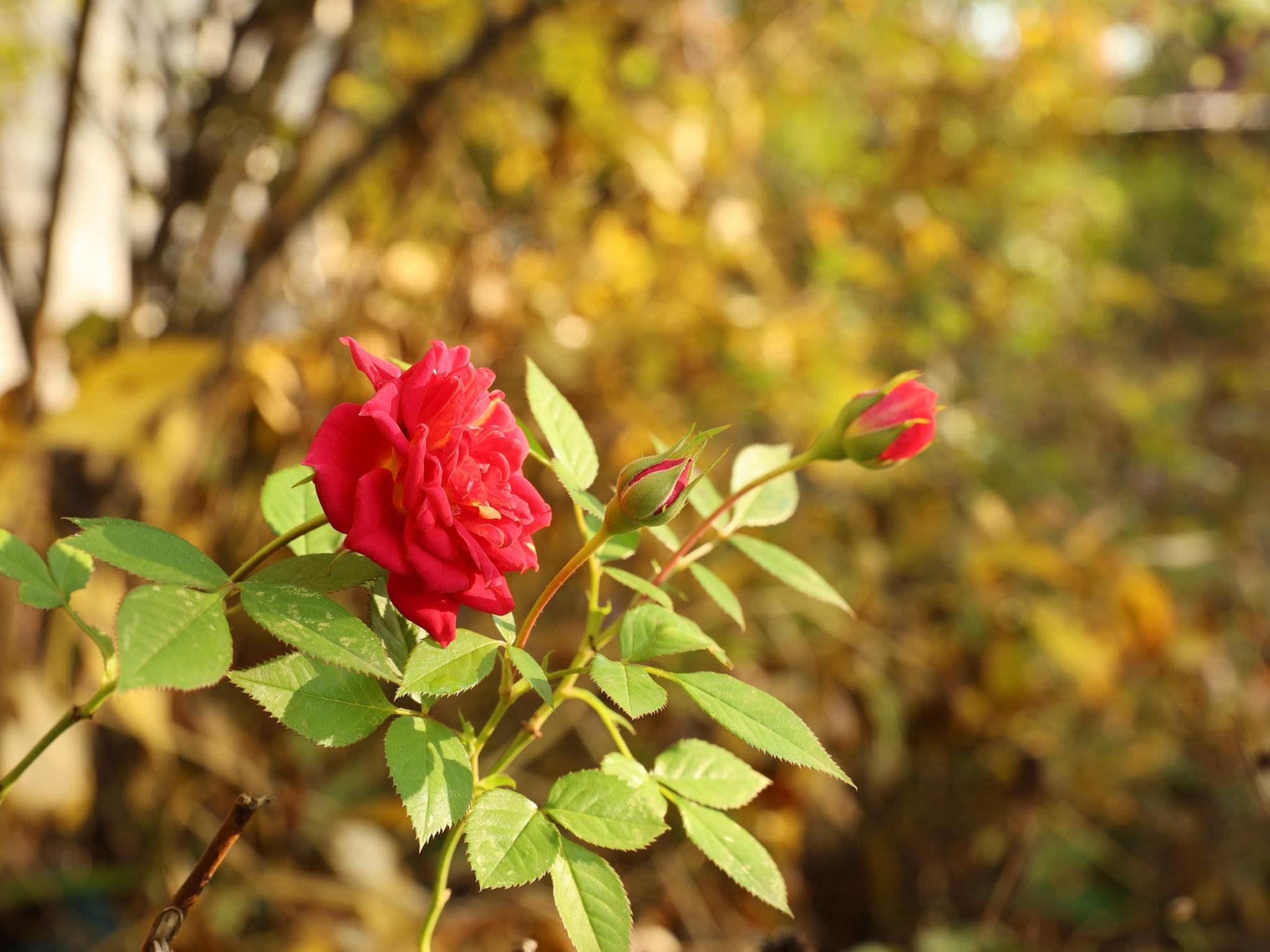} & \includegraphics[width=\widthcomp\linewidth]{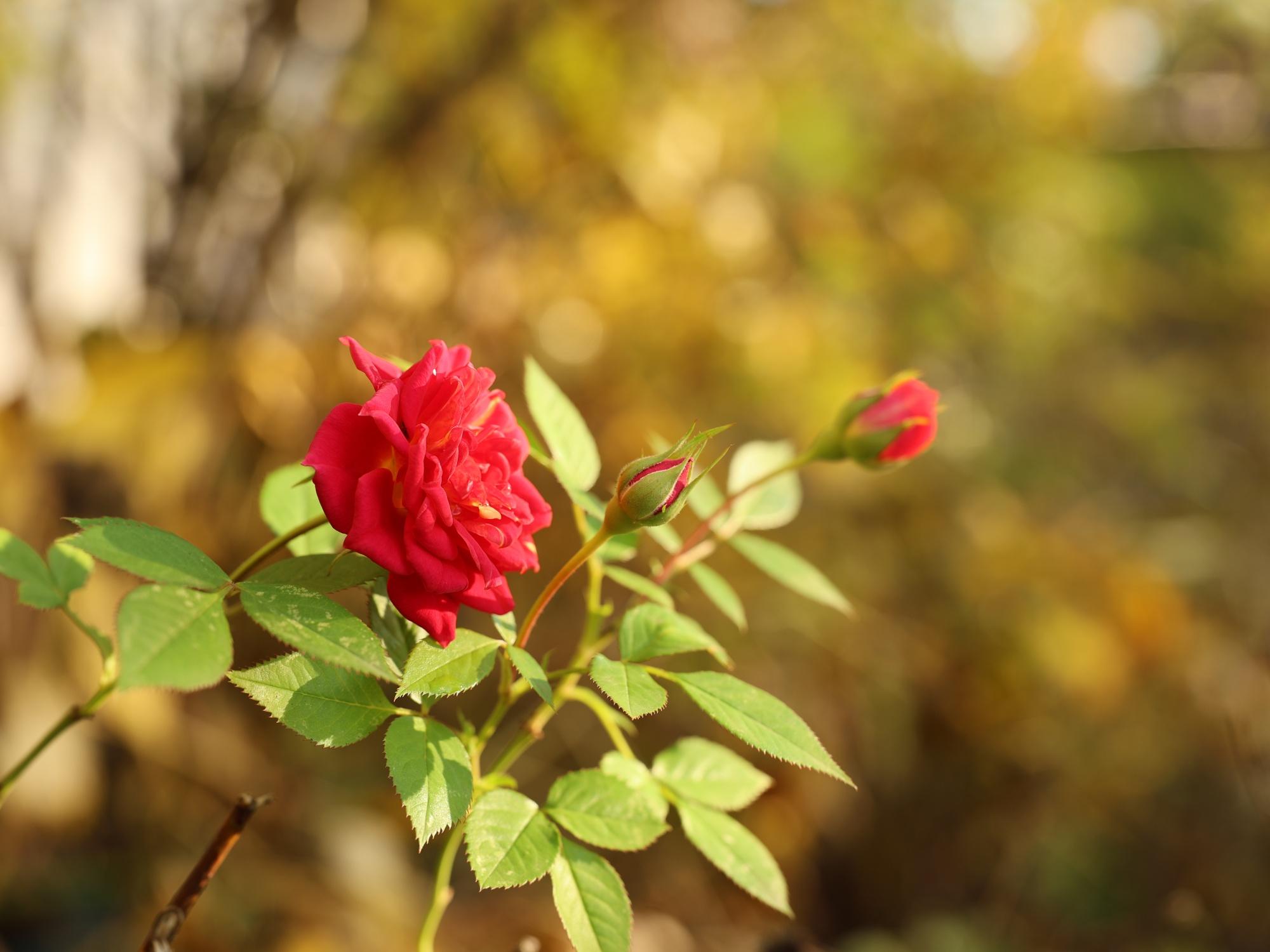} & 
    \includegraphics[width=\widthcomp\linewidth]{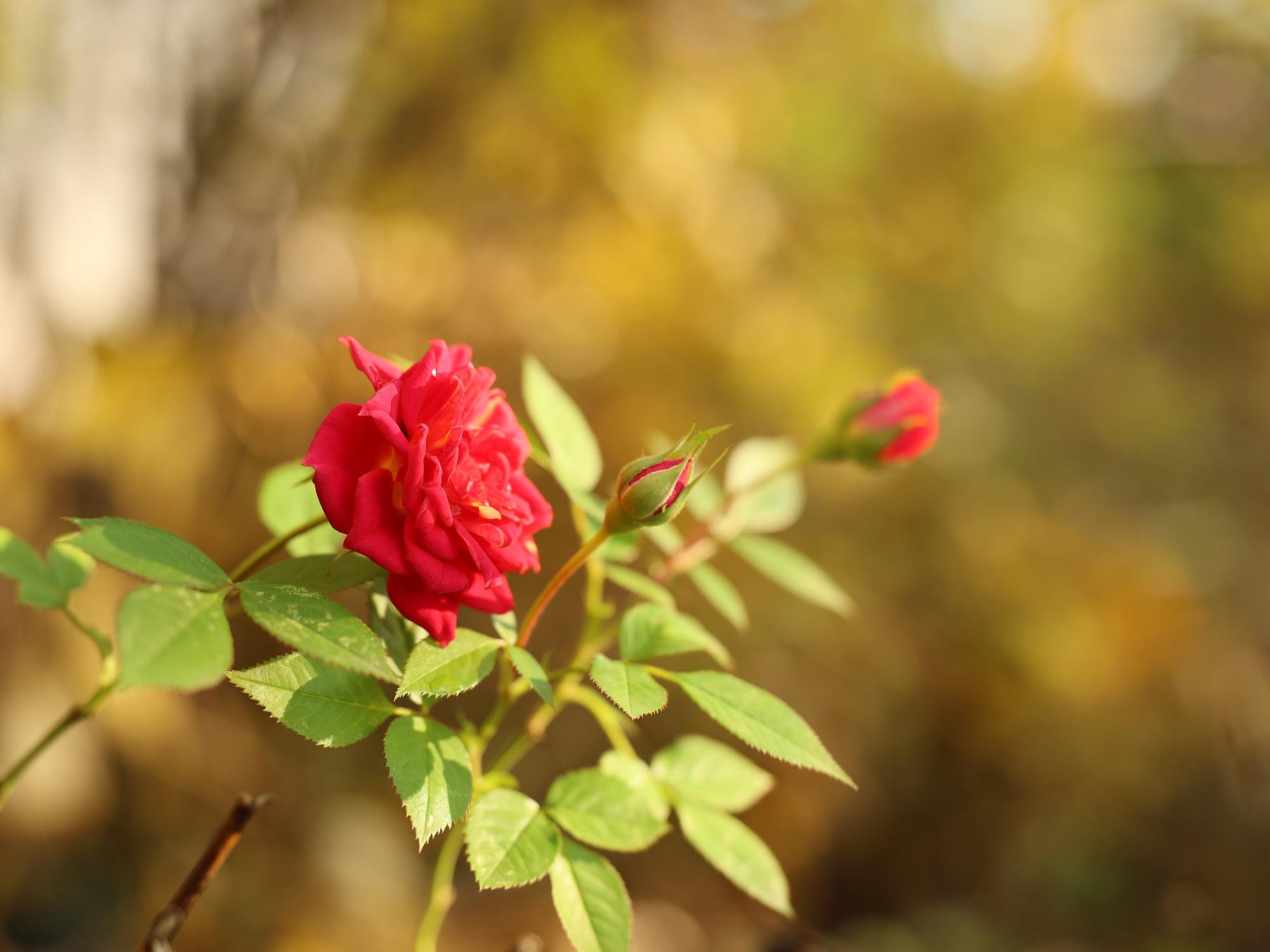} &
    \includegraphics[width=\widthcomp\linewidth]{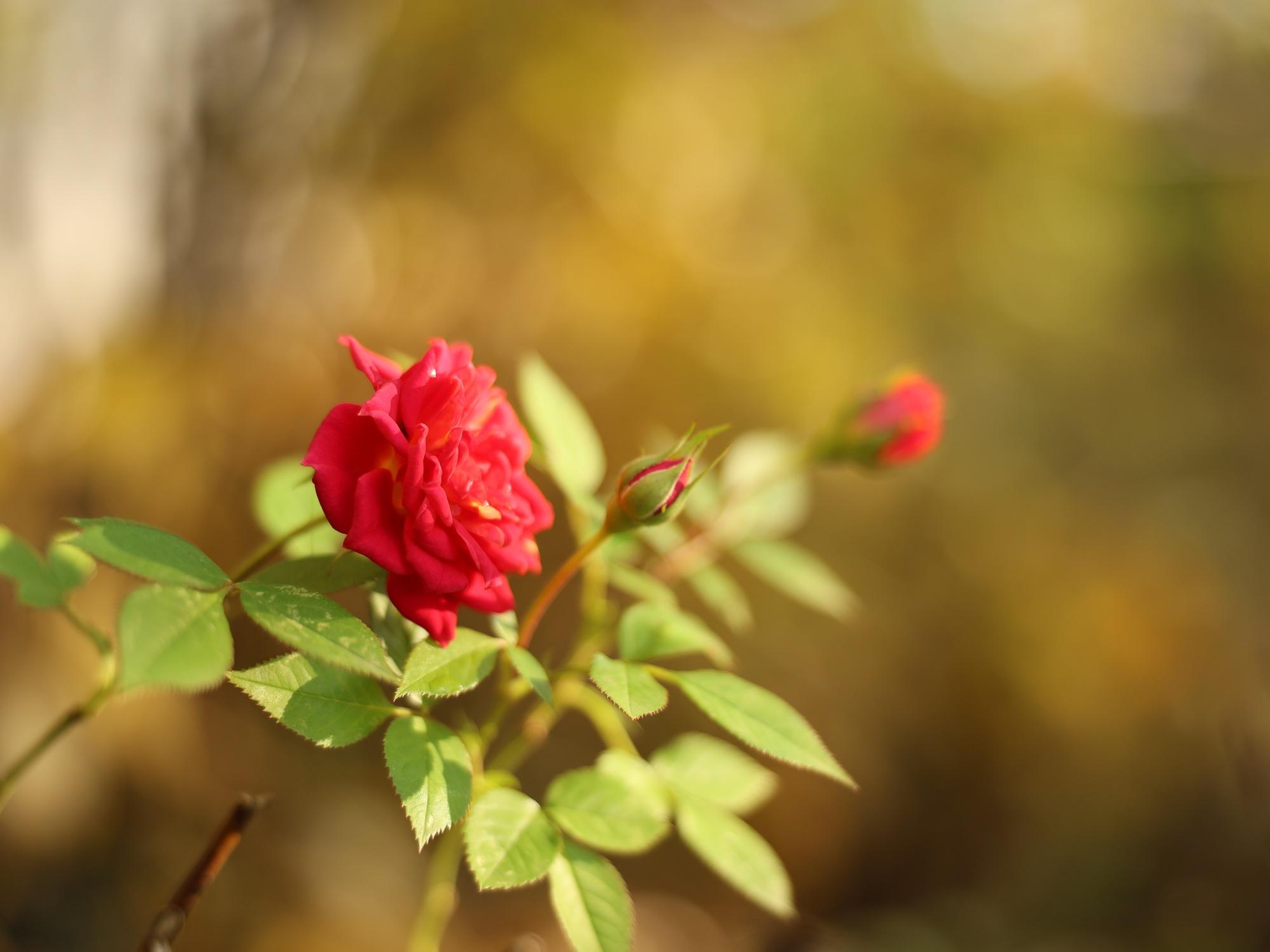} \\
    \addlinespace[0.5pt]
    & Input \fnum{22.0} & GT \fnum{10.0} & GT \fnum{4.0} & GT \fnum{2.8} & GT \fnum{2.0}  \\
    \addlinespace[0.2pt]
    \rotatebox{90}{\hspace{0.5mm}Focallength 28mm} &
    \includegraphics[width=\widthcomp\linewidth]{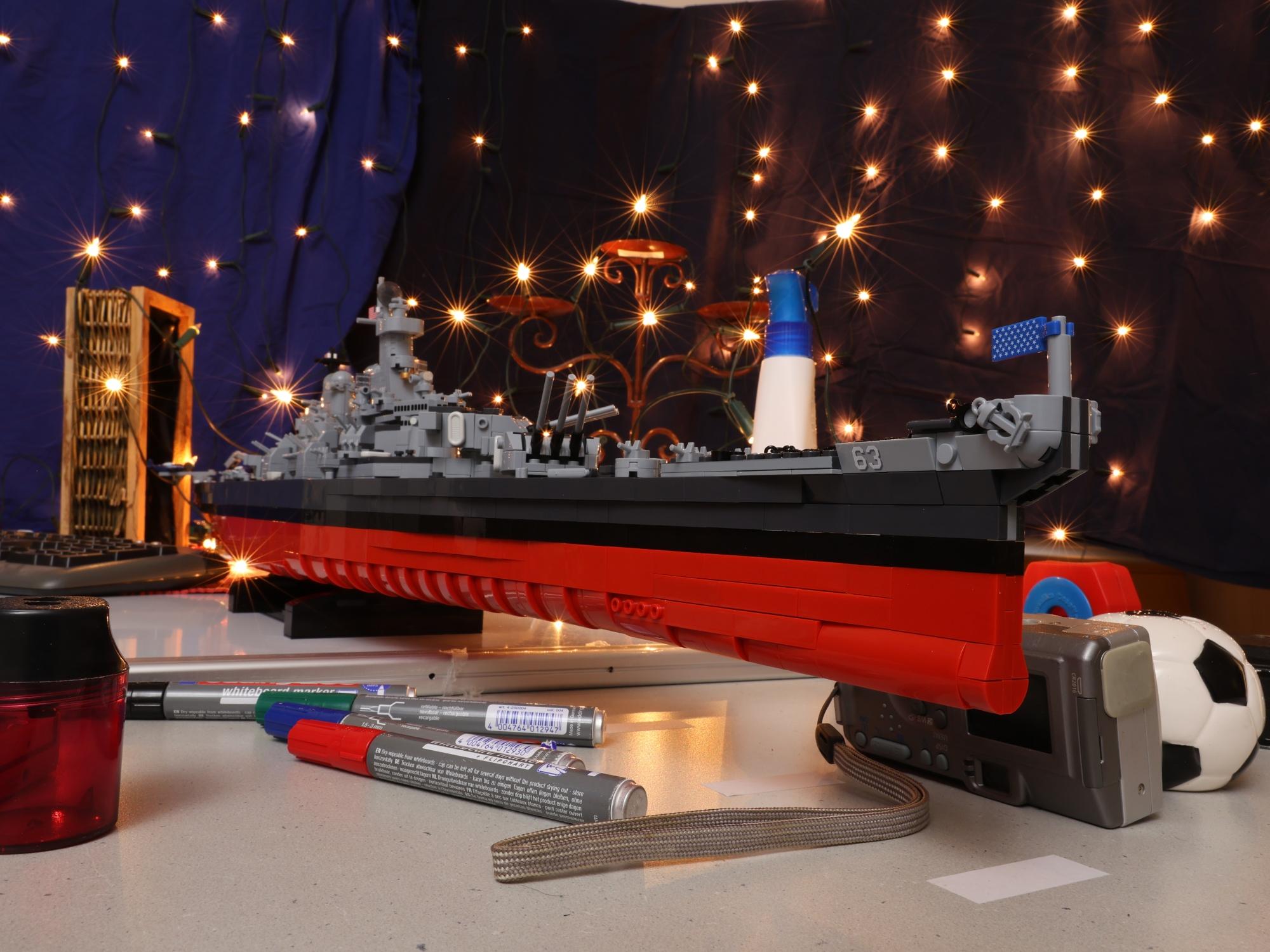} & \includegraphics[width=\widthcomp\linewidth]{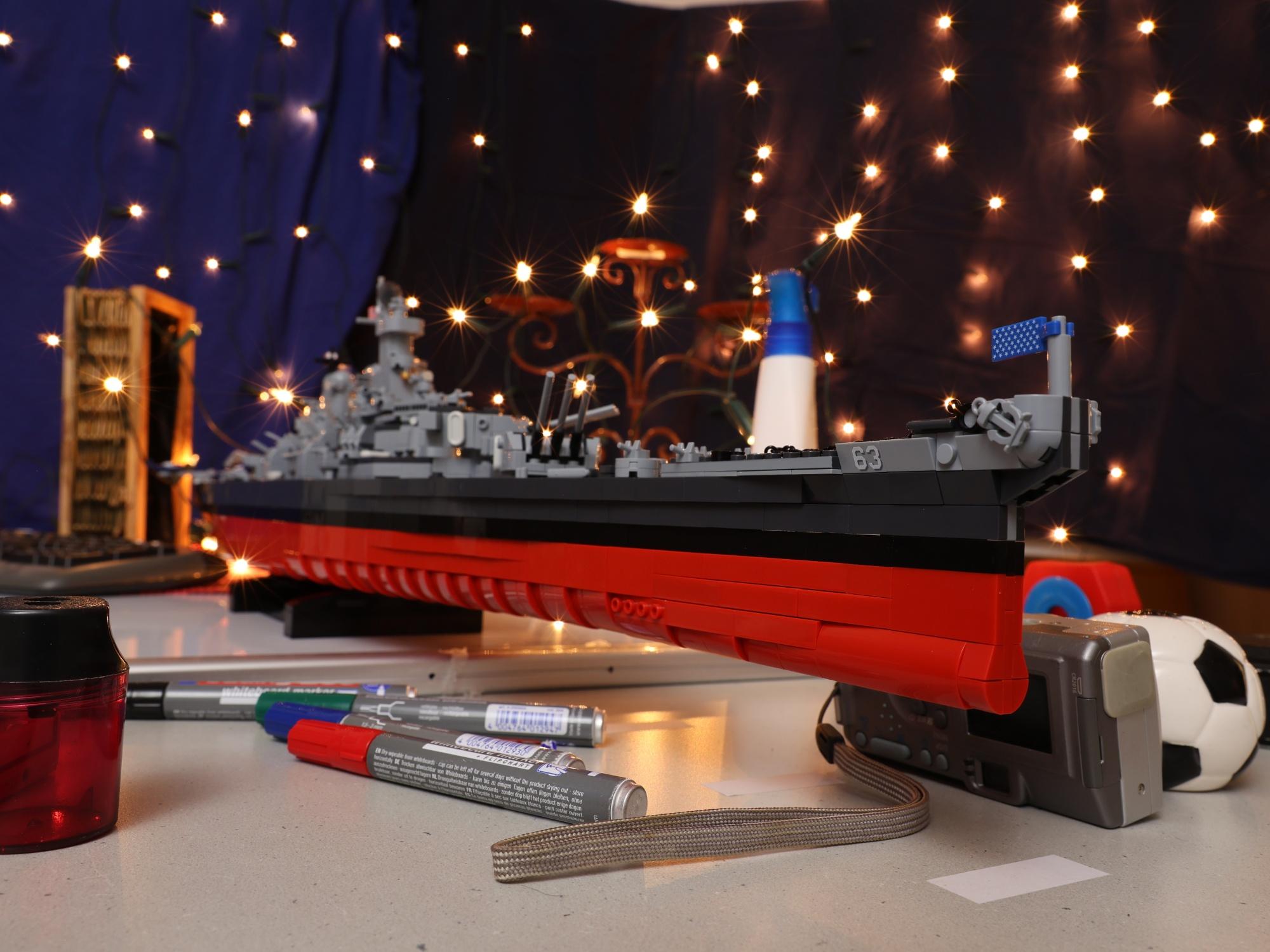} & \includegraphics[width=\widthcomp\linewidth]{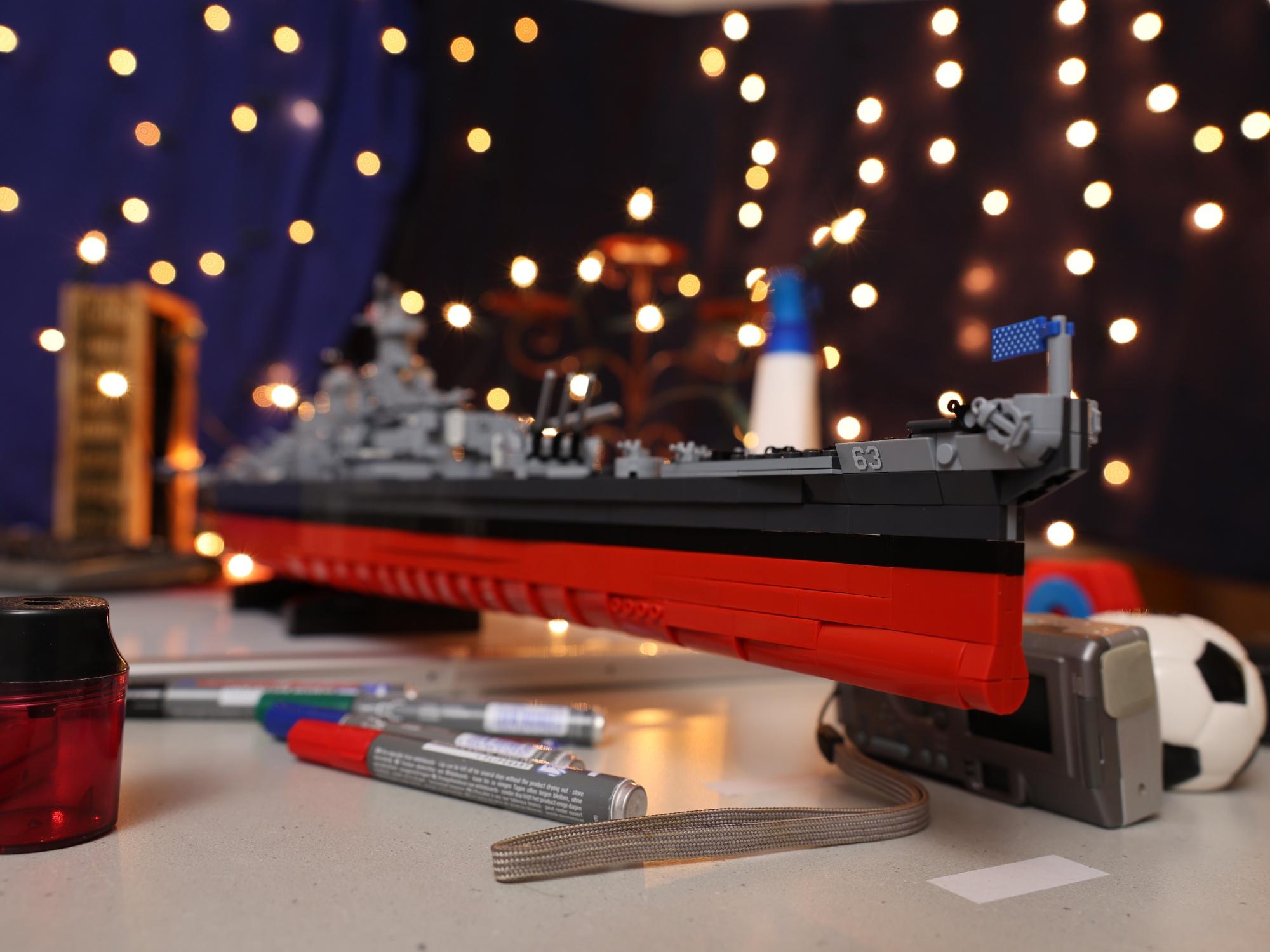} & \includegraphics[width=\widthcomp\linewidth]{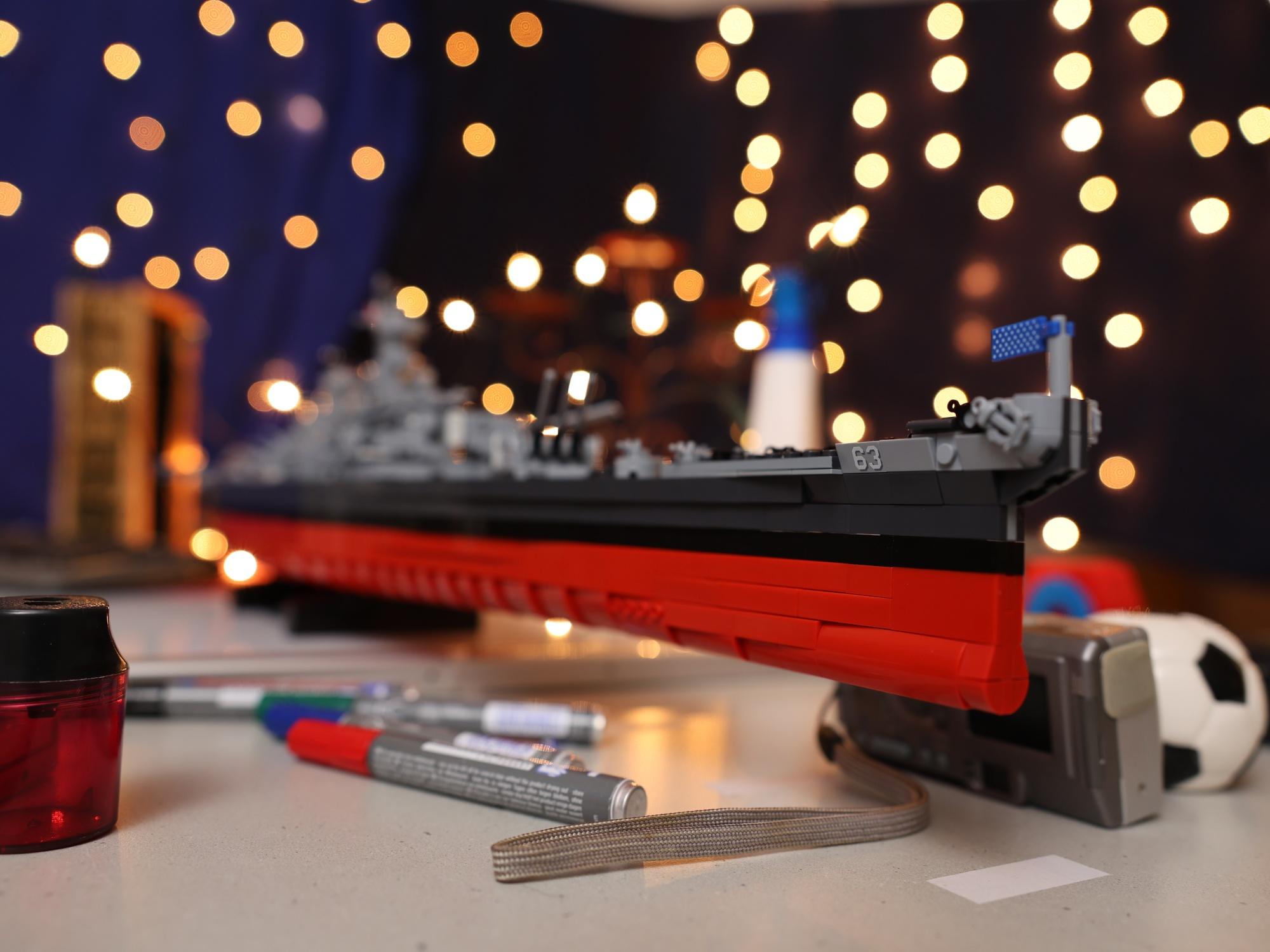} & \includegraphics[width=\widthcomp\linewidth]{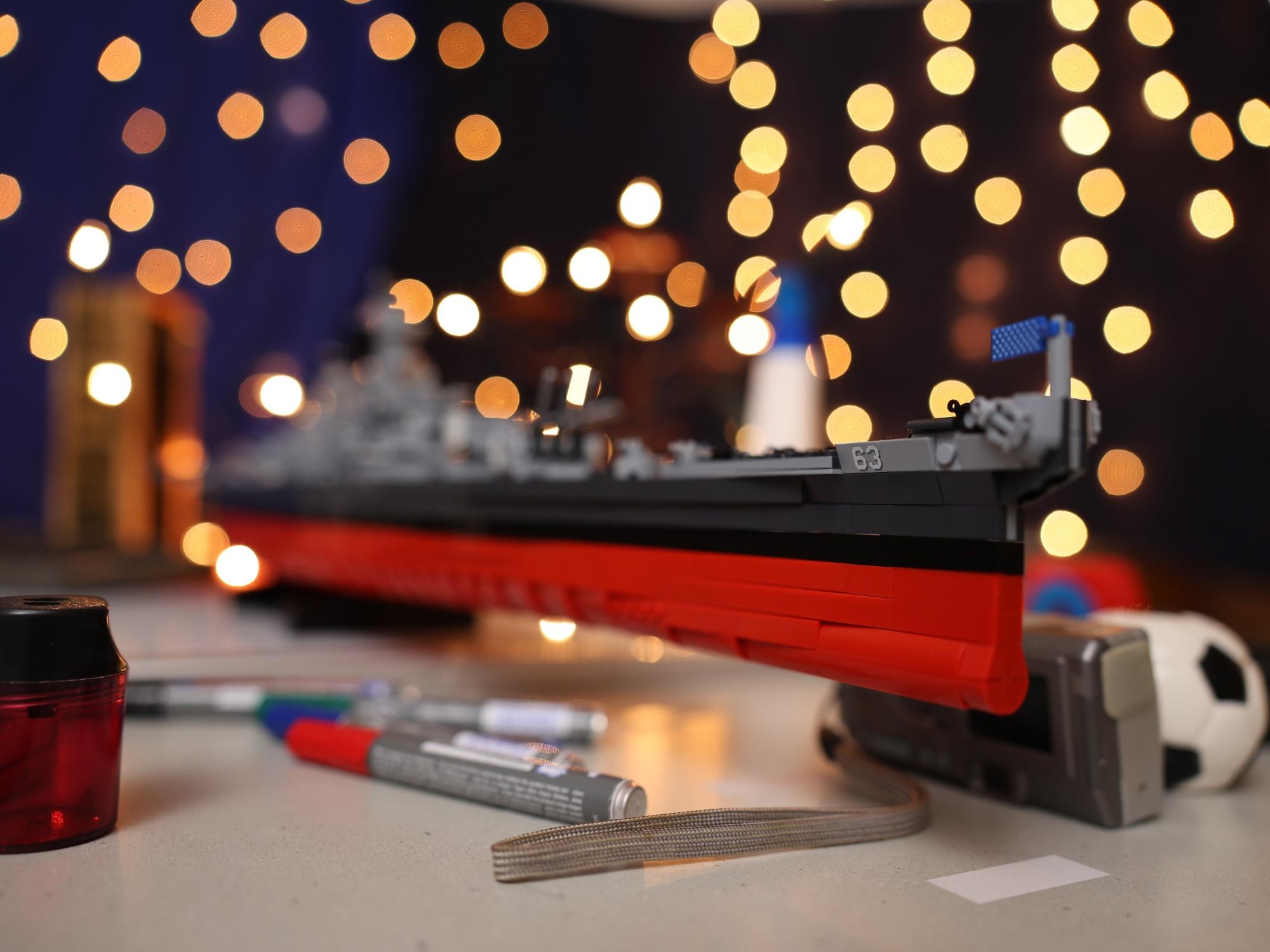} \\
    \end{tabular}
    \vspace{-0.25cm}
    \caption{Sample scenes from \textbf{\dataset}.
    Note the \textbf{perfect alignment} and pronounced Bokeh effect varying with the aperture setting.
    }
    \vspace{-3mm}
    \label{fig:RealBokeh}
\end{figure*}

Controlling the aperture to adjust the size of Bokeh is a key aspect in the photographic process~\cite{mercado2019filmmaker}.
A \textit{large} \fstop number such as \fnum{2.0} indicates an open aperture and results in a strong Bokeh effect.
For a weaker Bokeh effect, the mechanism is gradually closed down, indicated by increasingly \textit{smaller} \fstop numbers like \fnum{4.0} or \fnum{8.0} as can be observed in \cref{fig:RealBokeh}~\cite{balcerzak2004actor, Merklinger1997technicalbokeh}.

\begin{figure}[h!]
    \centering
    \setlength{\tabcolsep}{0.4pt}
    \footnotesize
    \renewcommand{\arraystretch}{0.4}
    \def\widthcomp{0.245}
    \def\widthcompp{0.17}
    \begin{tabular}{ccccc}
        & \fnum{2.0} & \fnum{2.8} & \fnum{4.0} & \fnum{8.0} \\
        \addlinespace[0.45pt]
        \rotatebox{90}{\hspace{0.5mm} BoMe~\cite{BokehMeHybrid}} &
        \includegraphics[width=\widthcompp\linewidth]{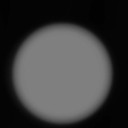} &
        \includegraphics[width=\widthcompp\linewidth]{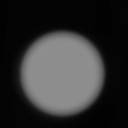} &
        \includegraphics[width=\widthcompp\linewidth]{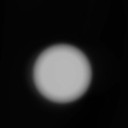} &
        \includegraphics[width=\widthcompp\linewidth]{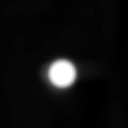} \\
        \rotatebox{90}{\hspace{3.2mm} Ours} &
        \includegraphics[width=\widthcompp\linewidth]{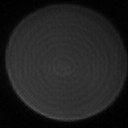} &
        \includegraphics[width=\widthcompp\linewidth]{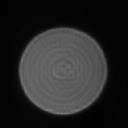} &
        \includegraphics[width=\widthcompp\linewidth]{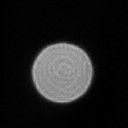} &
        \includegraphics[width=\widthcompp\linewidth]{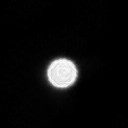} \\
        \rotatebox{90}{\hspace{3.2mm} Real} &
        \includegraphics[width=\widthcompp\linewidth]{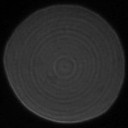} &
        \includegraphics[width=\widthcompp\linewidth]{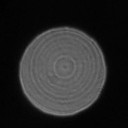} &
        \includegraphics[width=\widthcompp\linewidth]{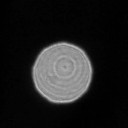} &
        \includegraphics[width=\widthcompp\linewidth]{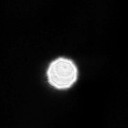} \\
    \end{tabular}
\vspace{-1mm}
\caption{Visualization of PSFs. Ours mimics the optical abbreviations visible in the real PSF while the handcrafted PSF of BoMe~\cite{BokehMeHybrid} lacks detail and is increasingly inaccurate for large \fstops.}
\label{fig:PSF}
\vspace{-5mm}
\end{figure}

Although Bokeh is a natural phenomenon, it is barely noticeable in smartphone cameras as a result of their small aperture lenses.
This led to an increasing interest in Bokeh rendering methods~\cite{GoogleCamera, barron2015fast, busam2019sterefo, yang2016virtual, ignatov2020rendering}. 
Many approaches implement multistep frameworks that require depth maps~\cite{GoogleCamera, BokehMeHybrid}.
Therefore, these methods require specialized camera hardware or additional computation, increasing their complexity.
Other works~\cite{ignatov2020rendering, nagasubramaniam2023BEViT, dutta2021stackedbokeh} attempt to model Bokeh rendering as a simple end-to-end problem, with a generator network trained on pairs of small and large aperture images.
Unfortunately, the current and most popular Bokeh rendering dataset, \textit{EBB!}~\cite{ignatov2020rendering}, lacks variability in \fstops, therefore, these solutions do not offer a way to adjust the strength of the generated Bokeh.
Furthermore, \textit{EBB!} suffers from low image detail and exhibits poor sample alignment, making it a poor choice for training and evaluation.

Although synthetic training data can correct some of these issues, the resulting approaches~\cite{wang2018deeplens, seizinger2023bokeh} have shown poor generalization in real-world settings.
Furthermore, handcrafted synthetic rendering systems often fail to match the complexity of real Bokeh as shown in \cref{fig:PSF}.
Hence, there is a strong need for a proper real-world dataset and a method that achieves \textit{aperture control}, while being \textit{efficient}.

In this paper, we aim to address these challenges simultaneously. 
To overcome the issues of \textit{EBB!}, we first introduce \dataset in \cref{sec:dataset}, a new real-world dataset for training and benchmarking. 
It contains 23,000 high-quality 24 megapixel images captured by professional photographers using studio-grade equipment.
\dataset is five times larger than EBB!~\cite{ignatov2020rendering}, covering a diverse range of environments and lighting conditions.
Most importantly, our dataset is the first to capture Bokeh effects of a real lens with variations in both \textit{aperture} and \textit{focal length}. Based on our dataset, we conduct a thorough benchmark of leading end-to-end learning methods in \cref{subsec:benchmark}. Despite the superior quality of our dataset, we find that these models still lack sharpness within the in-focus regions and struggle to accurately replicate a real PSF, as shown in \cref{fig:benchmark}. We attribute this limitation to their implicit learning strategy, which fails to effectively incorporate the underlying physical principles of the Bokeh effect.

Inspired by this observation, we introduce \method in \cref{sec:method}, a simple, \textit{efficient} yet scalable model for generating \textit{realistic} Bokeh effects with \textit{controllable apertures}. 
Drawing inspiration from physical lens mechanics, we enhance the commonly used transformer attention~\cite{fan2024rmt} to be aperture-aware. 
Specifically, we mimic the aperture mechanism by adapting the width of the attention mask according to the desired \fstop. 
Unlike existing deep architectures~\cite{ignatov2020rendering, nagasubramaniam2023BEViT} which learn the Bokeh effect implicitly, our approach is more intuitive, interpretable, and explicit.
This physical prior enables us to achieve SOTA performance with minimal computational complexity, setting new records on both the established benchmarks~\cite{dutta2021depth,BokehMeHybrid} and our newly-introduced \dataset. Thanks to our diverse dataset and explicit Bokeh modeling, our method demonstrates strong generalization to unseen scenarios in \cref{subsec:realworld} and performs well in closely related tasks, such as defocus deblurring in \cref{subsec:beyond}.

\noindent To conclude, our contributions are \textbf{threefold:}

\begin{itemize}
    \item We introduce \dataset, a novel and comprehensive dataset for Bokeh rendering. Unlike existing publicly available datasets, our proposal offers high-quality and diverse samples captured at varying f-stops.

    \item We propose \method, the first end-to-end Bokeh rendering framework with intuitive aperture control, capable of producing realistic outputs and setting new SOTA records in performance and efficiency.

    \item Leveraging our dataset, we establish an extensive benchmark that thoroughly assesses model performance in Bokeh Rendering. We hope our dataset and model can inspire future works on advancing the field.
\end{itemize}
\section{Related Work}
\label{sec:related}

Early Bokeh rendering methods segmented the subject in a photo and applied the effect to the background~\cite{shen2016automatic, Shen2016DeepAP, zhu2017fast}.
More recent proposals increase realism by using techniques such as the application of a depth-dependent kernel~\cite{BokehMeHybrid} or the decomposition of an image into depth-wise layers to individually blur each~\cite{kraus2007depth, peng2021interactive, luo2023defocus, wadhwa2018synthetic, busam2019sterefo}.
Such techniques often suffer from artifacts in depth discontinuities, causing Bokeh to bleed into focus areas~\cite{BokehMeHybrid, sheng2024dr, zhang2019synthetic}.
To mitigate this issue, the hybrid approach BoMe proposes the use of an additional neural rendering module in these difficult depth discontinuous image areas~\cite{BokehMeHybrid}.

In practice, these rendering approaches usually require a multitude of additional input channels. 
These can be for depth~\cite{wadhwa2018synthetic, zhang2019synthetic, peng2021interactive, sheng2024dr, luo2023defocus}, subject segmentation~\cite{chen2017deeplab, Shen2016DeepAP, zhu2017fast, wadhwa2018synthetic, peng2021interactive, luo2023defocus}, high dynamic range recovery~\cite{zhang2019synthetic}, or background inpainting~\cite{peng2022mpib, sheng2024dr}.
In practice, this manifests itself in specialized camera requirements or additional processing modules, increasing the overall complexity.

Moreover, the Bokeh rendering process itself is usually based on a simplified, hand-crafted virtual lens model~\cite{wadhwa2018synthetic, BokehMeHybrid, sheng2024dr}. As can be observed in \cref{fig:PSF}, such a lens model does not account for the complexity of the point spread function (PSF) associated with real lenses~\cite{ignatov2020rendering, abuolaim2021learning}.
The shape and size of the PSF depend not only on depth, but also on aperture setting, focal length, focus distance, optical aberration, and radial distortion~\cite{alzayer2023dc2, levin2011understanding, tang2012utilizing}.

To address these shortcomings, some recent work has modeled Bokeh as a single-step neural rendering problem~\cite{ignatov2020rendering, dutta2021stackedbokeh, BokehGlassGAN, nagasubramaniam2023BEViT}.
Here, a model learns to generate Bokeh from a dataset of paired small- and large-aperture images.
This process is highly dependent on the qualities of the training data set. 
For example, if trained on data collected in the real-world with a photographic lens, the model can learn to implicitly replicate the complexity of its PSF.
Most neural rendering solutions~\cite{ignatov2023realistic, BokehGlassGAN, nagasubramaniam2023BEViT, dutta2021stackedbokeh, purohit2019depth, wang2022self} are trained on the pioneering real-world \textit{EBB!} dataset~\cite{ignatov2020rendering}.
However, due to inherent limitations of \textit{EBB!}, these methods are not yet satisfactory. 
For instance, misalignment between image pairs makes it difficult to train a method that can correctly render complicated object boundaries, such as hair.
Furthermore, since \textit{EBB!} is limited to a single aperture setting, it is impossible to directly train a controllable method.

Other approaches~\cite{kong2023bokeh, seizinger2023bokeh, wang2018deeplens} use artificial training data~\cite{conde2023lens, wang2018deeplens}, but these methods show poor generalization to real-world images, limiting their wider use.
\section{\dataset}

\begin{table}[t]
\centering
\footnotesize
\setlength{\tabcolsep}{1pt}
\renewcommand{\arraystretch}{0.9}
\begin{tabular}{@{}lccccc@{}}
\toprule
                  & \textbf{\sdataset (\textit{ours})} & EBB!~\cite{ignatov2020rendering}& Aperture~\cite{zhang2019synthetic}& BEDT\cite{conde2023lens}\\ \midrule
\# Samples        & \textbf{23,000}  & 4694              & 2942        & 20,000    \\
\# Train         & \textbf{20,500}         & 4400             & 2942        & 20,000     \\
\# Validation        & \textbf{1,250}  & 294              & -        & -    \\
\# Test        & \textbf{1,250}  & -              & -        & -    \\
Apertures        & \textbf{\fnum{20.0} - \fnum{2.0}}               & \fnum{1.8}                 & \fnum{8.0} \& \fnum{2.0}           & \fnum{2.0} \& \fnum{1.8}        \\
Focal Length & \textbf{28mm - 70mm}            & 85mm                & unknown     & unknown  \\
Resolution & \textbf{6000$\times$4000} & 1536$\times$1024 & 1200$\times$750 & 1920$\times$1080 & \\
Real              & \textbf{Yes}                       & Yes               & Yes         & No       \\
Aligned           & \textbf{Yes}                       &  No  & No   & Yes      \\
Public           & \textbf{Yes}                       & Val294~\cite{dutta2021depth} & partial & Train-only       \\ \bottomrule
\end{tabular}
\vspace{-1mm}
\caption{Statistics about currently proposed Bokeh Rendering datasets compared to our \dataset.}
\label{tab:datasets}
\vspace{-4mm}
\end{table}

\label{sec:dataset}

To promote the research of Bokeh Rendering in the deep learning era, a large number of realistic, diverse, high-quality sample pairs with good alignment are critical~\cite{vasluianu2024ntire}. 
Unfortunately, these qualities are not well represented in the current real-world Bokeh rendering datasets~\cite{zhang2019synthetic,ignatov2020rendering}.

To address this gap, we propose \dataset. As shown in \cref{tab:datasets}, it contains 23.000 wide and small aperture image pairs showing \datasetsizescene different scenes in total. Our proposal excels in scale and variety.

Expert photographers collected our dataset in the wild using a high-end Canon Eos R6 II DSLM camera system with a Canon 28-70mm \fnum{2.0} \textit{L} zoom lens.
Pixel-level alignment of image pairs is ensured by best practices such as the use of tripod, electronic shutter, and remotely triggered automated capture.
In addition, each pair was manually checked and discarded if misalignment was found.
It is the first real-world study of Bokeh with varying focal lengths and aperture \fstops. Some examples can be seen in \cref{fig:RealBokeh}.

\noindent\textbf{$\text{\dataset}_{\text{\textit{bin}}}$}:
Most Bokeh rendering systems are designed to be trained with a fixed aperture across all samples~\cite{ignatov2020rendering, nagasubramaniam2023BEViT, dutta2021stackedbokeh}. 
Hence, for benchmarking purposes, we provide a \textit{binary} version of \dataset limited to a single aperture, similar to EBB!~\cite{ignatov2020rendering}. 
To create $\text{\dataset}_{\text{\textit{bin}}}$ we use the split of \cref{tab:datasets} but only include sample pairs for \fnum{2.0}.

\begin{table}[]
    \footnotesize
    \centering
    \setlength{\tabcolsep}{3pt}
    \renewcommand{\arraystretch}{0.8}
    \begin{tabular}{l|ccc}
	Setting & Day/Night & Sunny/Cloudy & Out-/In-door \\
	\midrule
	Distribution & 70\% / 30\% \hspace{2.4pt} & 60\% / 40\% \hspace{2.2pt} & \hspace{0.7pt} 70\% / 30\%  \\
    \end{tabular}
    \vspace{-1mm}
    \caption{Environment Conditions represented in RealBokeh.}
    \vspace{-4mm}
    \label{tab:RB_dist}
\end{table}

\noindent \textbf{Capture Process:}
\dataset is collected with the help of camera automation, capturing a scene in less than two seconds.
This short capture time, combined with remote triggering, significantly reduces alignment issues caused by camera operations or dynamic environments.

For each scene, the photographer selects an appropriate focal length and focus subject. The camera then captures five images at different \fstops.
First, the \textit{input} image is captured at \fnum{22.0}.
Then three \textit{ground truth} images are taken at random \fstops between \fnum{20.0} and \fnum{2.2} at 1/3-stop increments.
Finally a \textit{ground truth} image at \fstop{2.0} is shot.
\dataset thus contains the \fstop variability of a real camera lens~\cite{allen2012manual}.
Examples of this protocol can be found in \cref{fig:RealBokeh}.

\noindent \textbf{Variety:}
The Bokeh effect is most pronounced in high-contrast lighting conditions. 
However, we found that previous datasets contain only a small number of outdoor locations under neutral daylight conditions.
To address this shortcoming, our \dataset features more than 120 distinct indoor and outdoor locations under a variety of environmental lighting conditions.
We outline the distributions of scenes depending on their environment in \cref{tab:RB_dist}.

Moreover, we designed a studio setup with numerous small background lights to foster the creation of complex Bokeh effects. In each scene, the positions of the lights, camera, and subjects were varied. Overall, \sdataset contains 300 studio scenes featuring more than 150 different objects. An example scene is shown in the last row of \cref{fig:RealBokeh}.

Previous datasets mostly lack portraits, which are crucial for real-world applications and challenging due to intricate details like hair.
As we found that natural human movement causes severe misalignment, we used a realistic full-sized human puppet with different outfits and hairstyles to include this important setting in \sdataset.
\section{\method}
\label{sec:method}

Leveraging physical priors within our Aperture Aware Attention (AAA), we propose \method as a simple yet efficient and effective baseline architecture for the task of Controllable Bokeh Rendering.

\begin{figure}
  \centering
  \includegraphics[width=\linewidth]{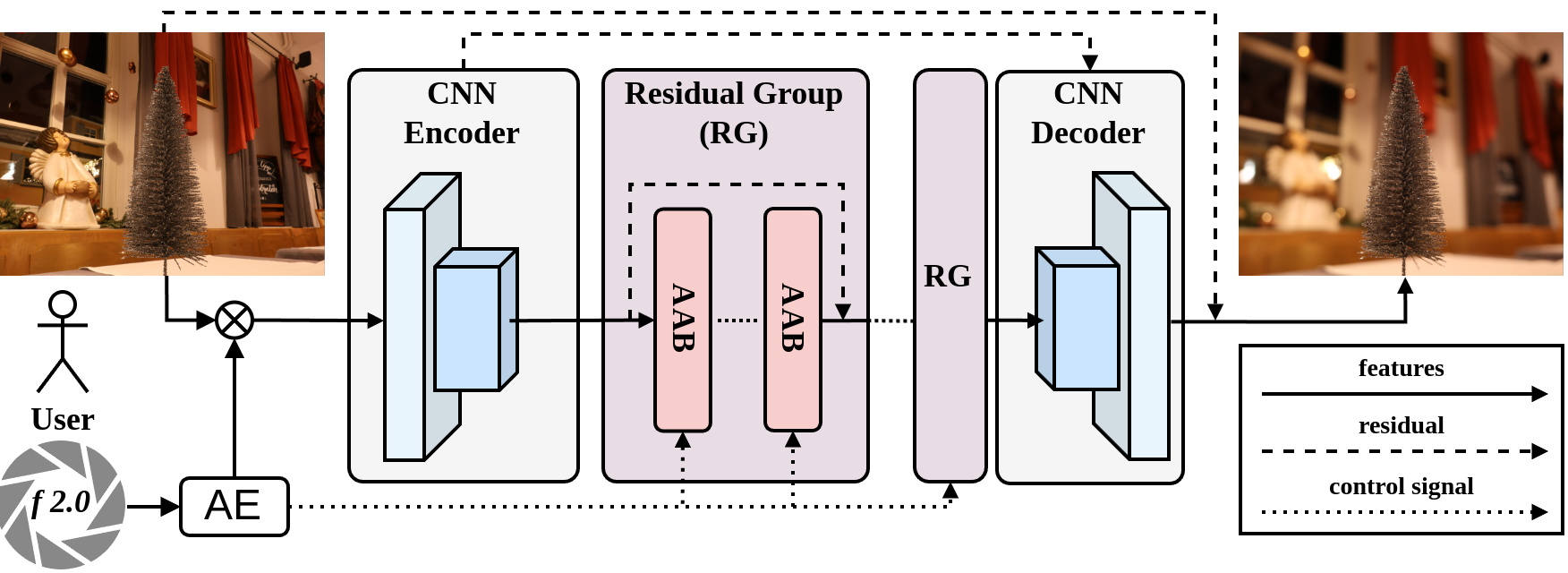}
  \vspace{-0.6cm}
  \caption{Overview of our propoesd \method architecture.}
  \vspace{-0.5cm}
  \label{fig:Arch}
\end{figure}

\noindent Key insights into design requirements are as follows.
\begin{itemize}
    \item Low processing demands are critical for handling large images. As shown in \cref{fig:Arch} we utilize an efficient en-/de-coder with residuals to maintain image detail.
    \item Controlling the Bokeh should be intuitive. For this we describe an Aperture Encoding scheme in \cref{eq:AVEnc}.
    \item Processing needs to be locally biased, while the receptive field has to be large enough to render strong Bokeh effects. Therefore, we employ the idea of Manhattan Self-Attention~\cite{fan2024rmt} as defined in \cref{eq:ManhattanDecayMask}.
    \item Bokeh strength varies throughout the image and depends on the aperture, therefore, we propose a novel adaptive multiscale transformer as shown in \cref{fig:AAA}.
\end{itemize}

\subsection*{Architecture Overview}

As depicted in \cref{fig:Arch} both CNN Encoder and Decoder utilize a lightweight NAFNet~\cite{chen2022simple} block design for feature compression and reconstruction.
Here, residual connections are crucial for maintaining details in the output image.

The proposed architecture uses a number of Residual Groups (RG) for deep feature processing.
Each RG aggregates the features of its multiple Aperture Attention Blocks (AABs) via residual connections.
This allows each RG to focus on a certain task, such as the rendition of Bokeh or the refining of earlier features.
For the design of AAB we adopt the block template of~\cite{fan2024rmt} to embed our AAA mechanism.

To further improve the rendition of Bokeh~\cite{seizinger2023bokeh}, we implement a CoordConv scheme~\cite{liu2018intriguing} in the first convolution operation of every block.

\subsection*{Aperture Control}
\label{subsec:AE}

To enable intuitive control of the Bokeh generation strength, we use a single scalar input.
To compute this parameter $f$,  we take the desired \fstop \textit{av} and apply a simple yet representative Aperture Encoding (AE) that represents the fractional diameter of a physical aperture through \cref{eq:AVEnc}.

\vspace{-3mm}

\begin{align}
    AE(av) &= \frac{av}{av_{max}}
    \label{eq:AVEnc}
\end{align}

\vspace{-1mm}

\begin{figure}
  \centering
  \includegraphics[width=\linewidth]{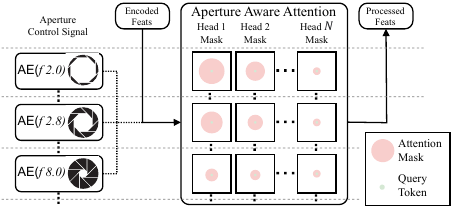}
  \vspace{-4mm}
  \caption{
  Each of the $N$ parallel heads within our AAA has a individual decay mask size, allowing it to attend to particular blur kernels. 
  The masks are further tuned towards rendering specific \fstops via the signal from our AE~\cref{eq:AVEnc}.}
  \vspace{-5mm}
  \label{fig:AAA}
\end{figure}

In \cref{eq:AVEnc}, $av_{max}$ is the maximum \fstop the network can reproduce, which corresponds to \fnum{2.0} in our dataset.
In \cref{fig:Arch}, the resulting parameter $f$ is used both for preconditioning and for controlling our AAA mechanism in \cref{fig:AAA}.

\subsection*{Aperture Aware Attention}
\label{subsec:AAA}

Our AAA mechanism in \cref{fig:AAA} is fundamentally inspired by distance-decay masks for attention in language modeling, introduced by Sun et al.~\cite{sun2023retentive}.
For vision applications, this distance is determined via the Manhattan-Distance between tokens within their 2D embedding matrix~\cite{fan2024rmt}.

Following~\cite{fan2024rmt} we can define a simple Manhattan decay mask $D$ for all pairs of tokens $(x_n, y_m)$ via \cref{eq:ManhattanDecayMask}.

\begin{equation}
    \label{eq:ManhattanDecayMask}
    \begin{aligned}
        D_{nm}^{2d}&=\gamma^{|x_n-x_m|+|y_n-y_m|} \\
        \mathrm{MaSA}(X) &= (\mathrm{Softmax}(Q K^\intercal )\odot D^{2d})V \\
    \end{aligned}
\end{equation}

Manhattan Self-Attention (MaSA) is well-suited for Bokeh Rendering, as the local spatial awareness introduced by the decay mask \textit{D} intuitively supports the modeling of circular blur kernels.
Furthermore, by introducing a factor $\lambda$ to control the extent of the activated attention map, the receptive field can encompass extensive areas of the image. 

When combining MaSA heads in a multi-head attention scheme, the decay rate $\lambda_i$ of each head can be adjusted to induce a varying local bias~\cite{fan2024rmt}.
As Bokeh can gradually increase in strength throughout image regions, this multi-head decay tailors each head toward processing blur kernels of a certain size, as seen in the first row of \cref{fig:AAA}. 
This eventually increases the smoothness of the rendered effect.

Ultimately, to improve the performance in the Controllable Bokeh Rendering task, we additionally tune the attention mechanism towards rendering a certain \fstop by using our aperture encoding scheme \cref{eq:AVEnc}.
The decay rate $\lambda_i$ of the $i$-th head depends on the control parameter $f$ via \cref{eq:aperture_retention}.

\begin{equation}
    \lambda_i(f) = 1-2^{-a-\frac{(f b-a)i}{N}}
    \label{eq:aperture_retention}
\end{equation}

In \cref{eq:aperture_retention} the smallest decay mask size is determined by the parameter $a$, while the largest is determined by $b$.

Given a Manhattan Decay mask $D_M$, our AAA mechanism is defined in \cref{eq:AAA}.

\begin{equation}
    \begin{aligned}
        D(f) &= \lambda_i(f) D_M \\
        AAA(X, f) &= (Softmax(QK^\intercal) \odot D(f)
    \end{aligned}
    \label{eq:AAA}
\end{equation}

The procedure is illustrated in \cref{fig:AAA}, where the decay masks in each row are tailored to render a specific \fstop.
\subsection{Implementation}
\label{subsec:implementation}
We extensively studied and determined suitable hyperparameters for our proposed \method architecture. 
Detailed results of our hyperparameter study are in the supplementary material.

\method-M is implemented with a CNN en-/de-coder width of 16 channels, three RGs, each containing three AABs with three attention heads on a 96-dimensional embedding.
To explore scalability, we additionally propose a \textit{large} version.
To obtain \smethod-L every above listed parameter of \method-M is doubled.

\paragraph{\textbf{Loss Function:}}

We explored losses from various SOTA Bokeh Rendering methods. 
A $L_1$ pixel loss term is widely adopted, while using a VGG-feature loss term~\cite{johnson2016perceptual} is also common~\cite{ignatov2020rendering, ignatov2020aim, BokehGlassGAN}. 
In our tests, the introduction of $\text{LPIPS}_{VGG}$ as an additional loss term to $L_1$ as suggested by~\cite{georgiadis2023adaptivebokeh} produced notably pleasing visuals, when balanced correctly. 
Therefore, our loss function is defined as \cref{eq:loss}.

\vspace{-2mm}

\begin{align}
    \mathcal{L} &= L_1 + \lambda L_{\text{LPIPS}_{VGG}}
    \label{eq:loss}
\end{align}

In \cref{eq:loss} the parameter $\lambda$ is used for balancing purposes. 
We find that $\lambda~=~0.6$ is a good choice to maximize visual fidelity within the learned Bokeh effect.

\paragraph{\textbf{Training:}}
\label{par:training}

All variants of \smethod utilized a batch size of $4$ and a training patch resolution of $512 \times 512 px$ to effectively capture the extent of strong Bokeh effects. 
We use Adam with a learning rate of $5e-4$. 
All experiments were performed on a single NVIDIA L40 GPU.
\begin{figure*}[h!]
    \centering
    \setlength{\tabcolsep}{0.25pt}
    \small
    \renewcommand{\arraystretch}{0.2}
    \def\widthcomp{0.195}
    \def\widthcompp{0.0969}
    \begin{tabular}{cccccccccccccc}
        \multicolumn{2}{c}{Input} & \multicolumn{2}{c}{Restormer~\cite{zamir2022restormer}} & \multicolumn{2}{c}{DMSHN~\cite{dutta2021stackedbokeh}} & \multicolumn{2}{c}{Ours} & \multicolumn{2}{c}{GT} \\
        \addlinespace[0.5pt]
        \multicolumn{2}{c}{\includegraphics[width=\widthcomp\linewidth, trim={0 80px 0 0px},clip]{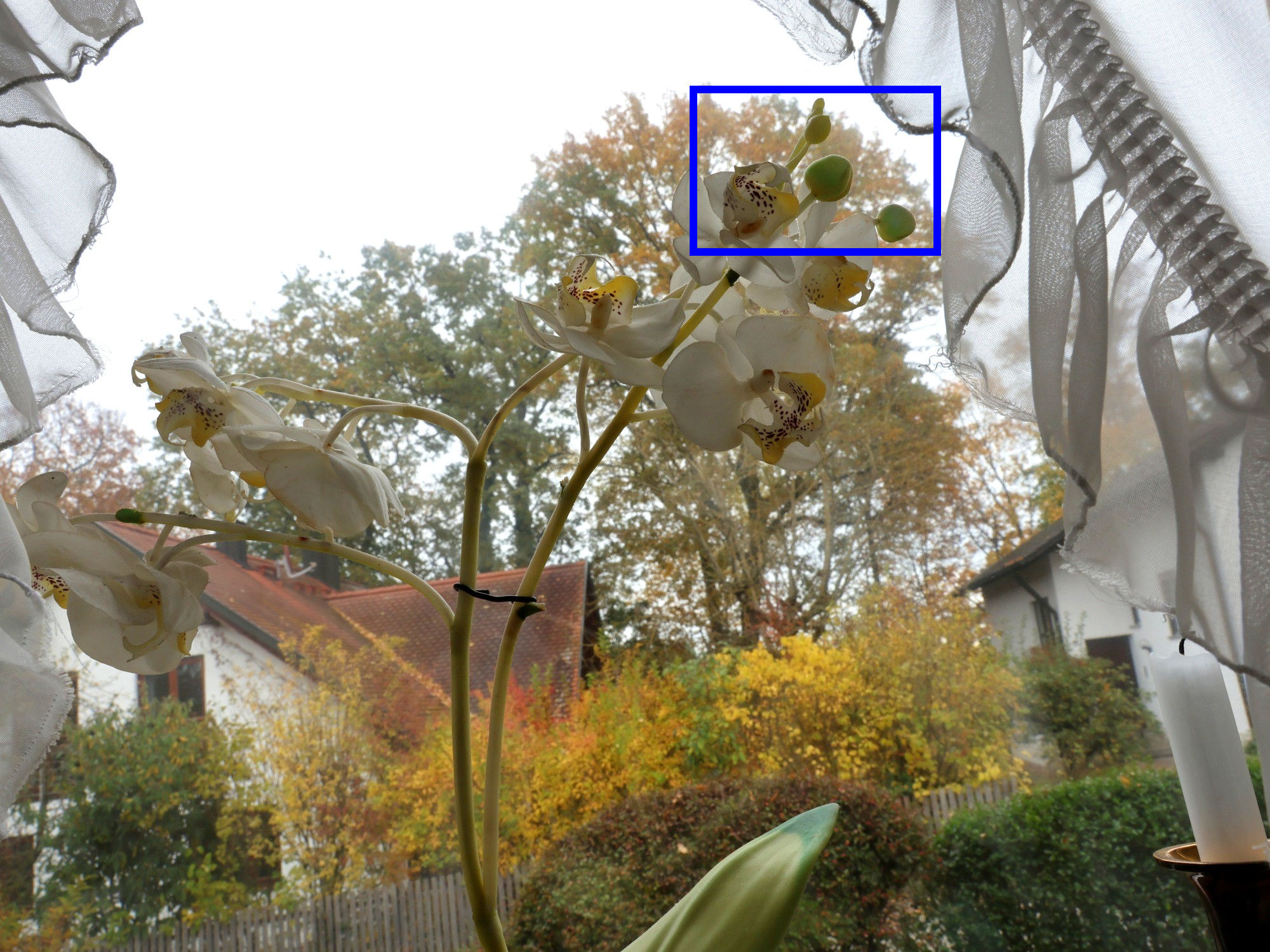}} &
        \multicolumn{2}{c}{\includegraphics[width=\widthcomp\linewidth, trim={0 80px 0 0px},clip]{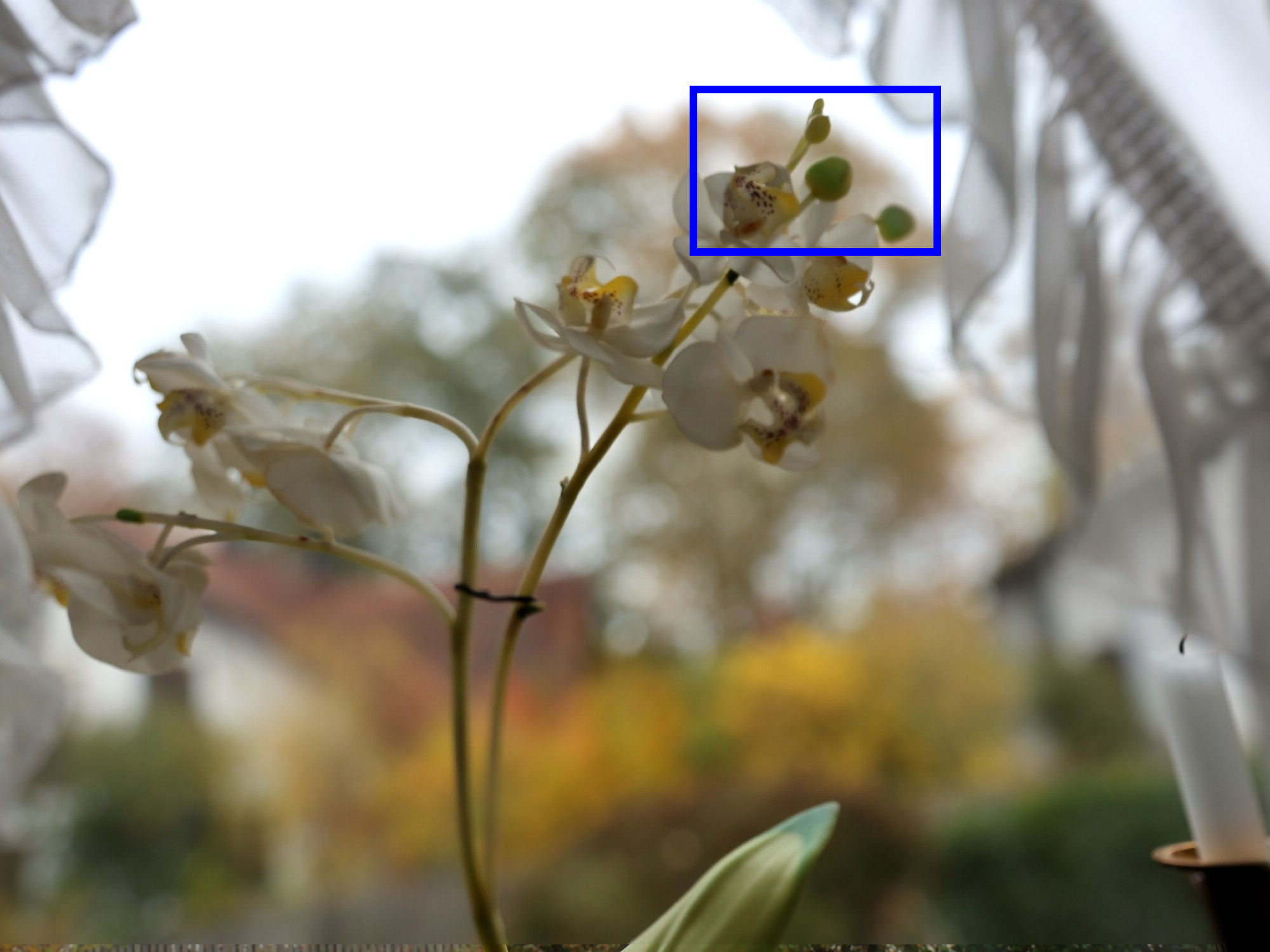}} &
        \multicolumn{2}{c}{\includegraphics[width=\widthcomp\linewidth, trim={0 80px 0 0px},clip]{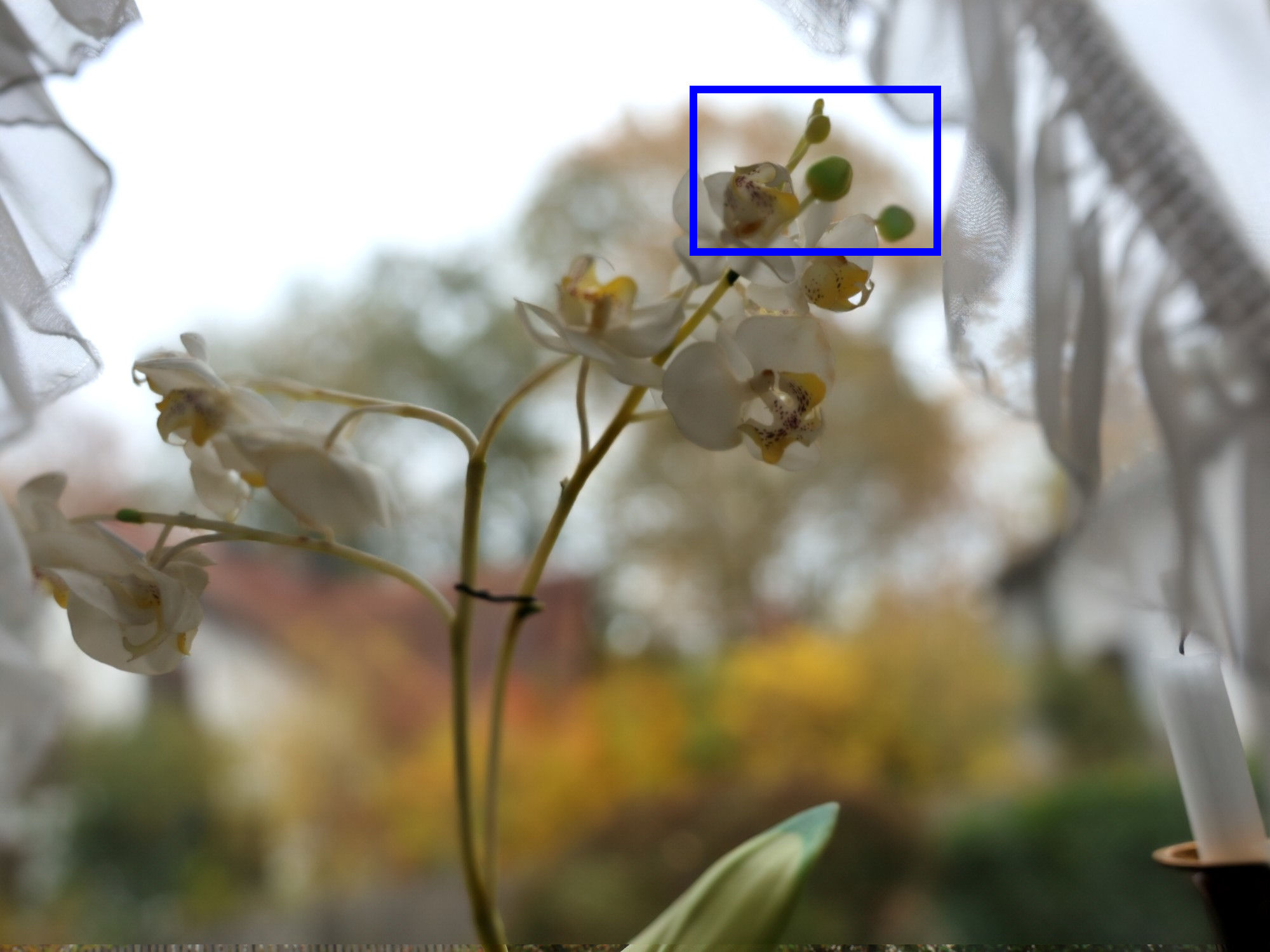}} &
        \multicolumn{2}{c}{\includegraphics[width=\widthcomp\linewidth, trim={0 80px 0 0px},clip]{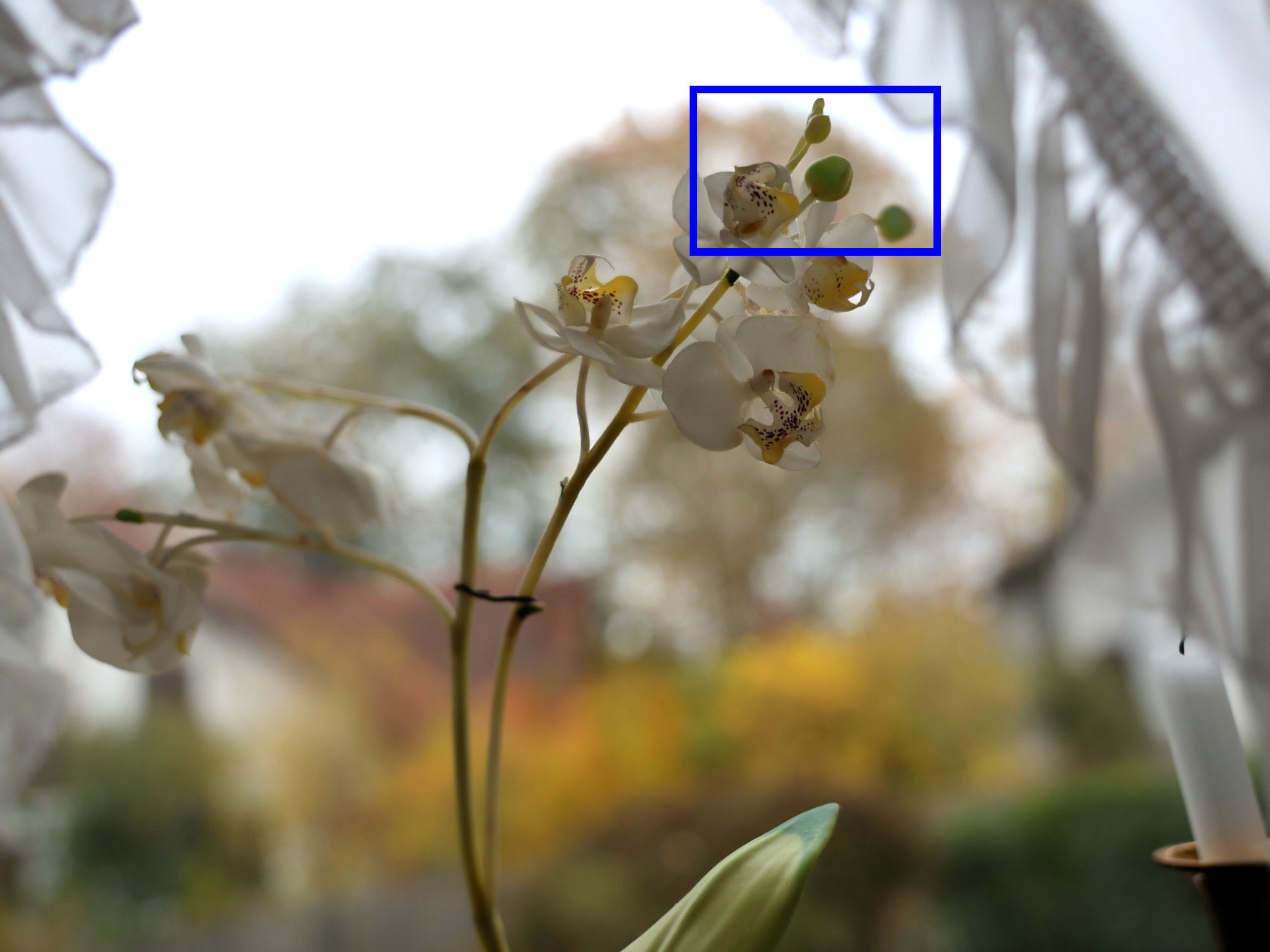}} &
        \multicolumn{2}{c}{\includegraphics[width=\widthcomp\linewidth, trim={0 80px 0 0px},clip]{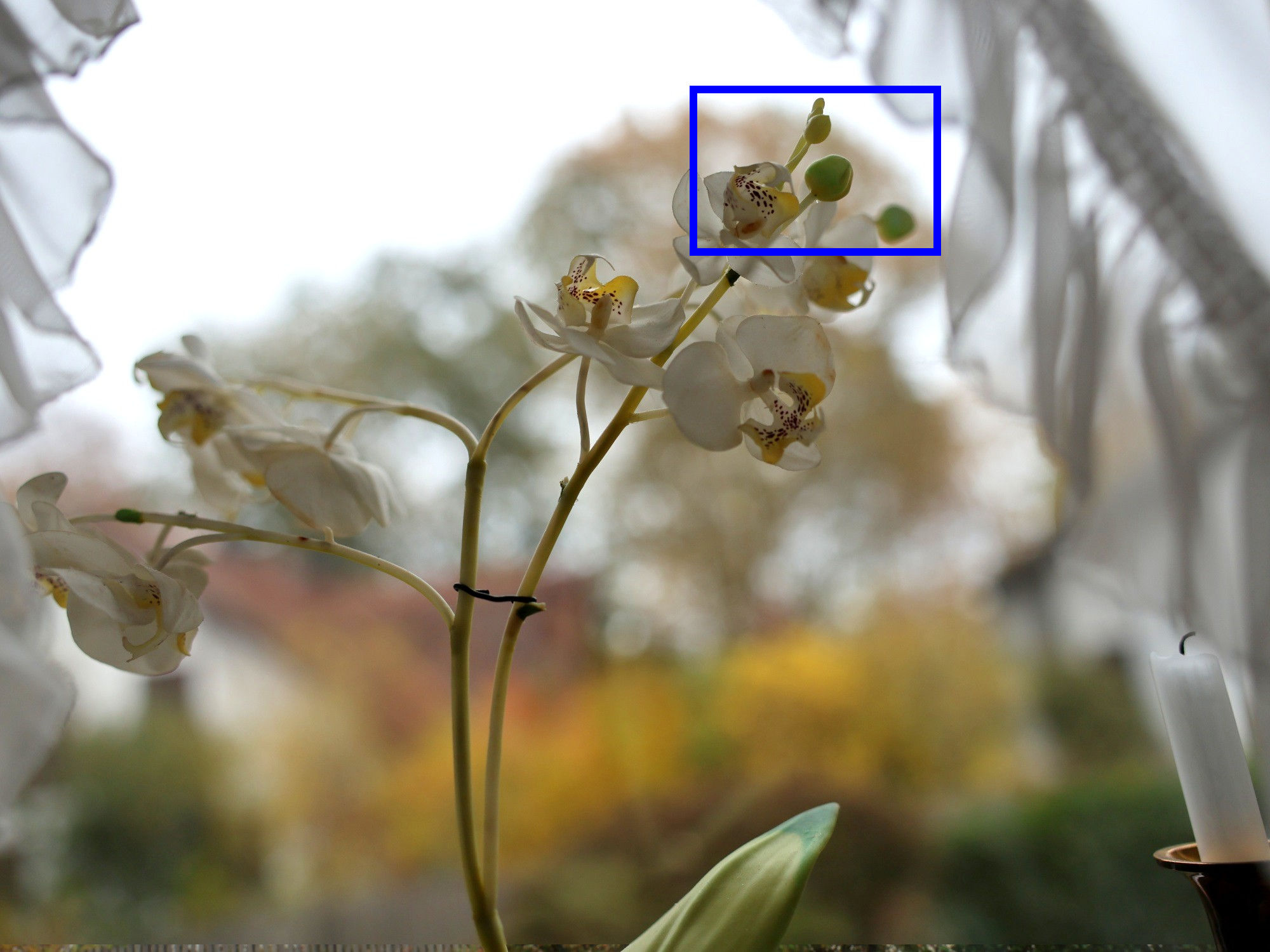}} \\
        \includegraphics[width=\widthcompp\linewidth]{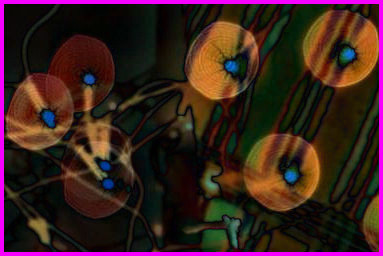} &
        \includegraphics[width=\widthcompp\linewidth]{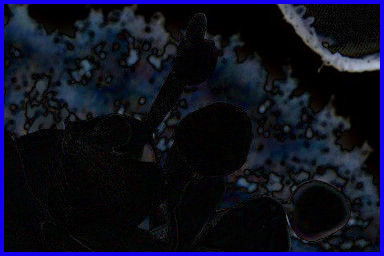} &        
        \includegraphics[width=\widthcompp\linewidth]{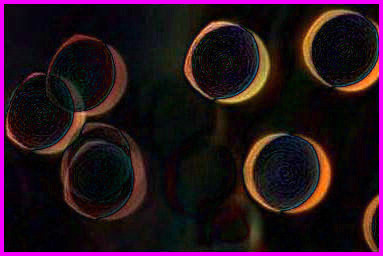} &
        \includegraphics[width=\widthcompp\linewidth]{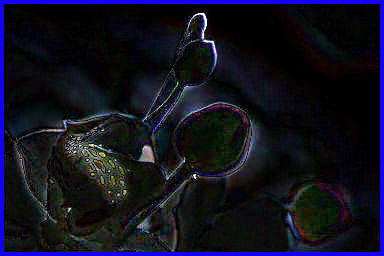} &
        \includegraphics[width=\widthcompp\linewidth]{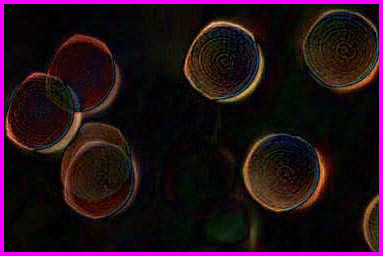} &
        \includegraphics[width=\widthcompp\linewidth]{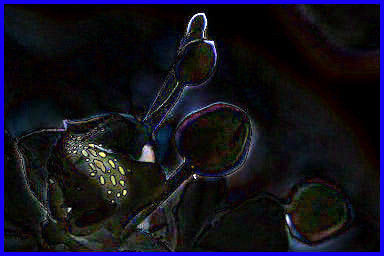} &
        \includegraphics[width=\widthcompp\linewidth]{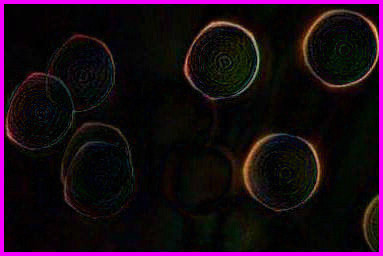} &
        \includegraphics[width=\widthcompp\linewidth]{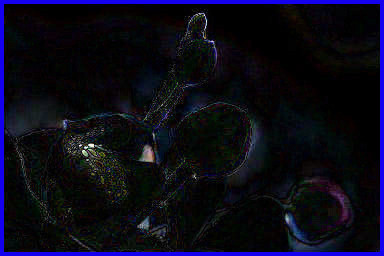} &
        \includegraphics[width=\widthcompp\linewidth]{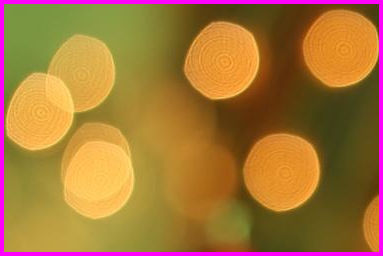} &
        \includegraphics[width=\widthcompp\linewidth]{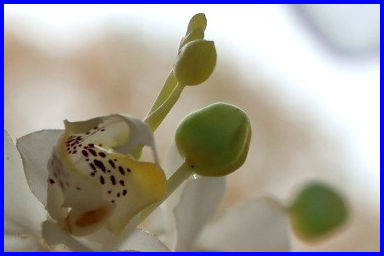} \\
        \addlinespace[0.5pt]
        \multicolumn{2}{c}{\includegraphics[width=\widthcomp\linewidth, trim={0 280px 0 0},clip]{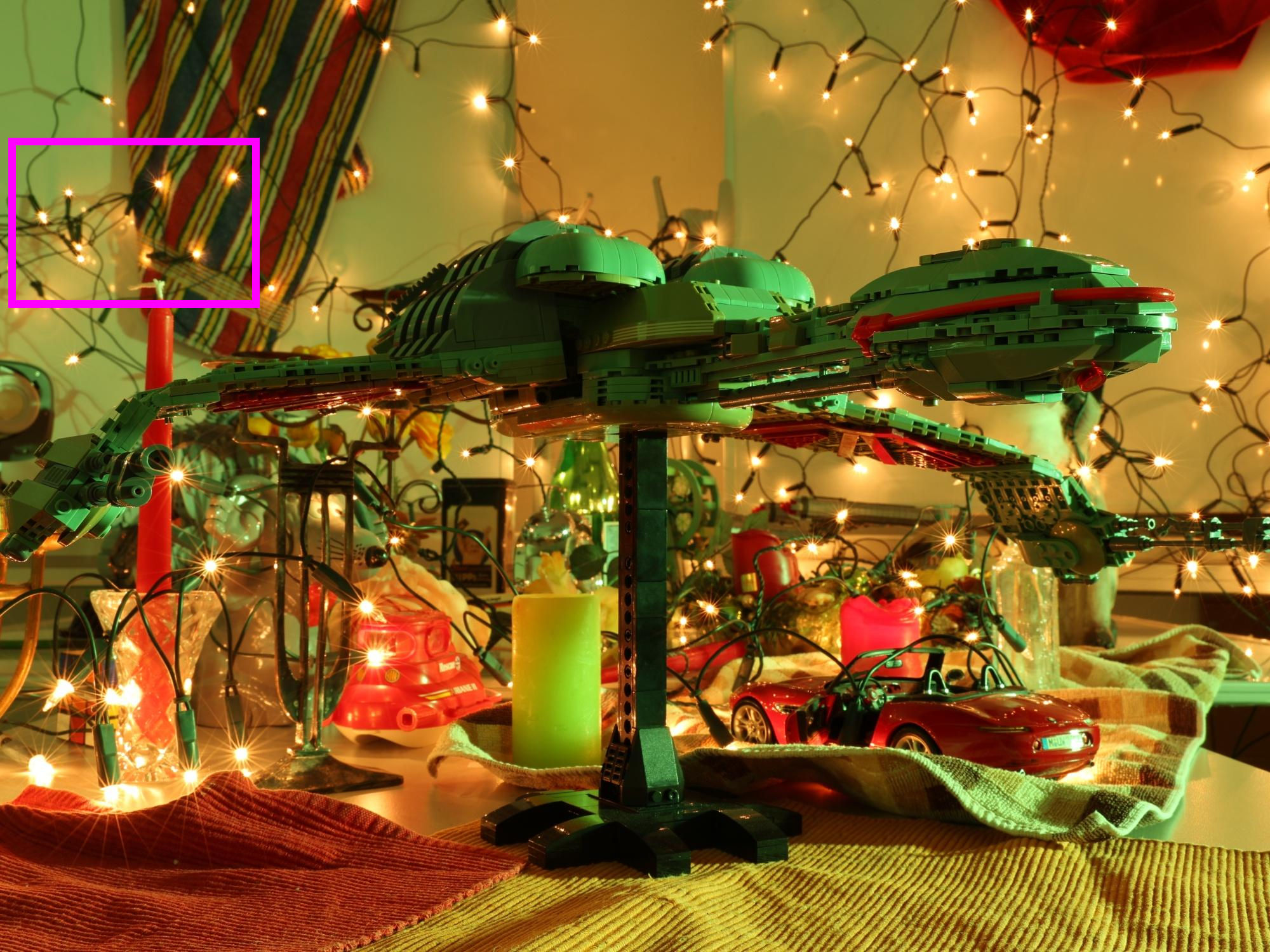}} &
        \multicolumn{2}{c}{\includegraphics[width=\widthcomp\linewidth, trim={0 280px 0 0},clip]{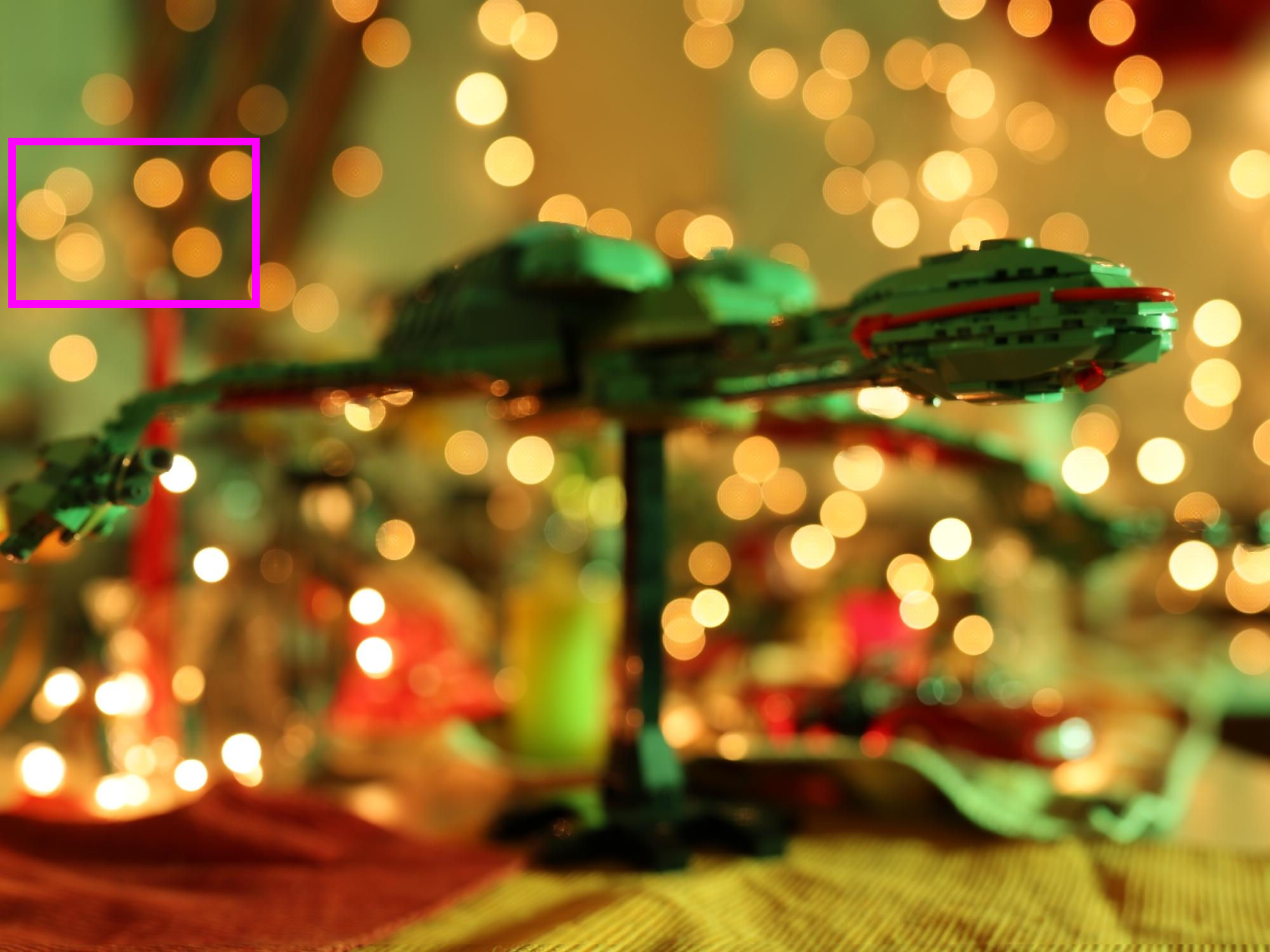}} &
        \multicolumn{2}{c}{\includegraphics[width=\widthcomp\linewidth, trim={0 280px 0 0},clip]{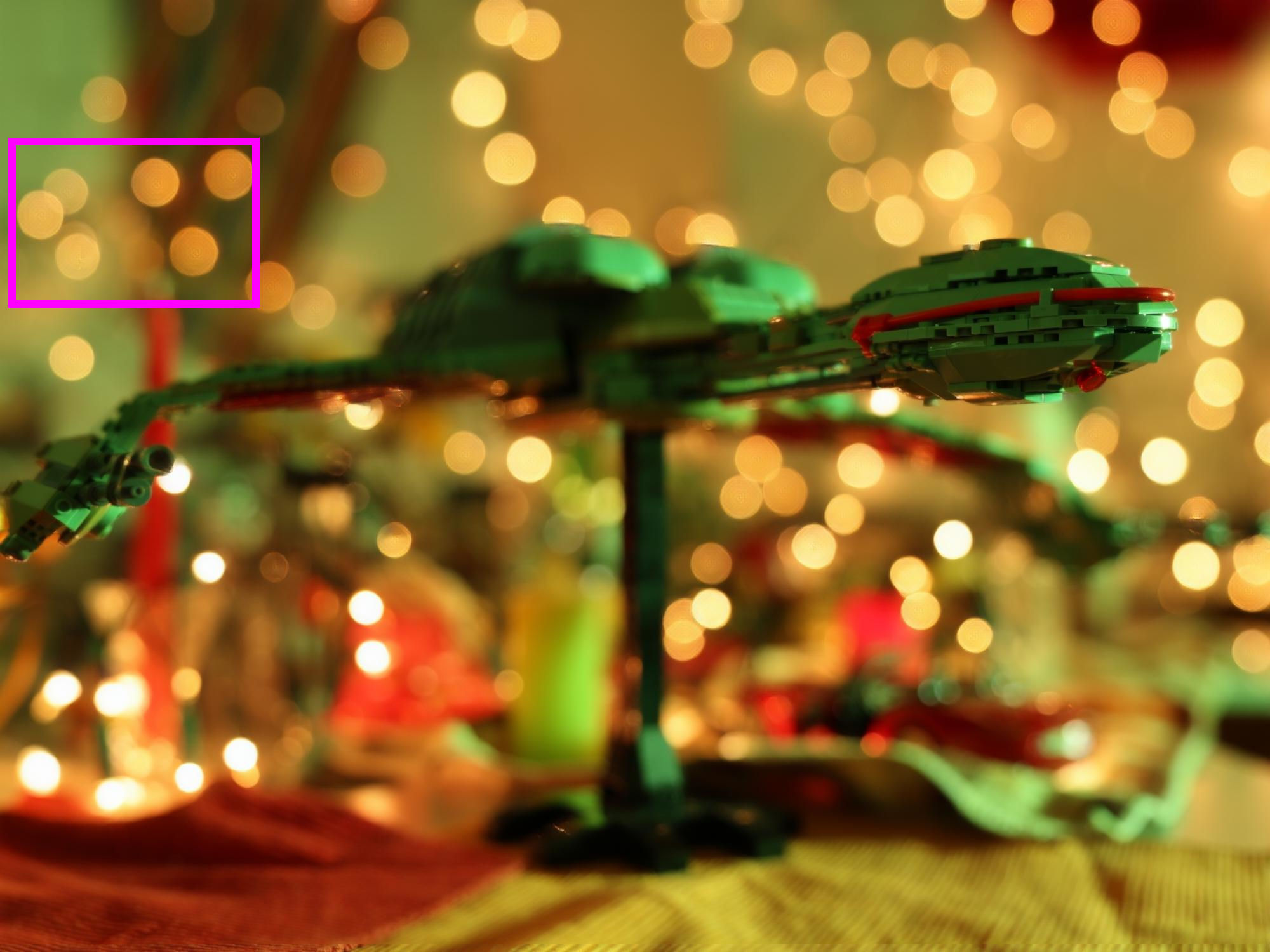}} &
        \multicolumn{2}{c}{\includegraphics[width=\widthcomp\linewidth, trim={0 280px 0 0},clip]{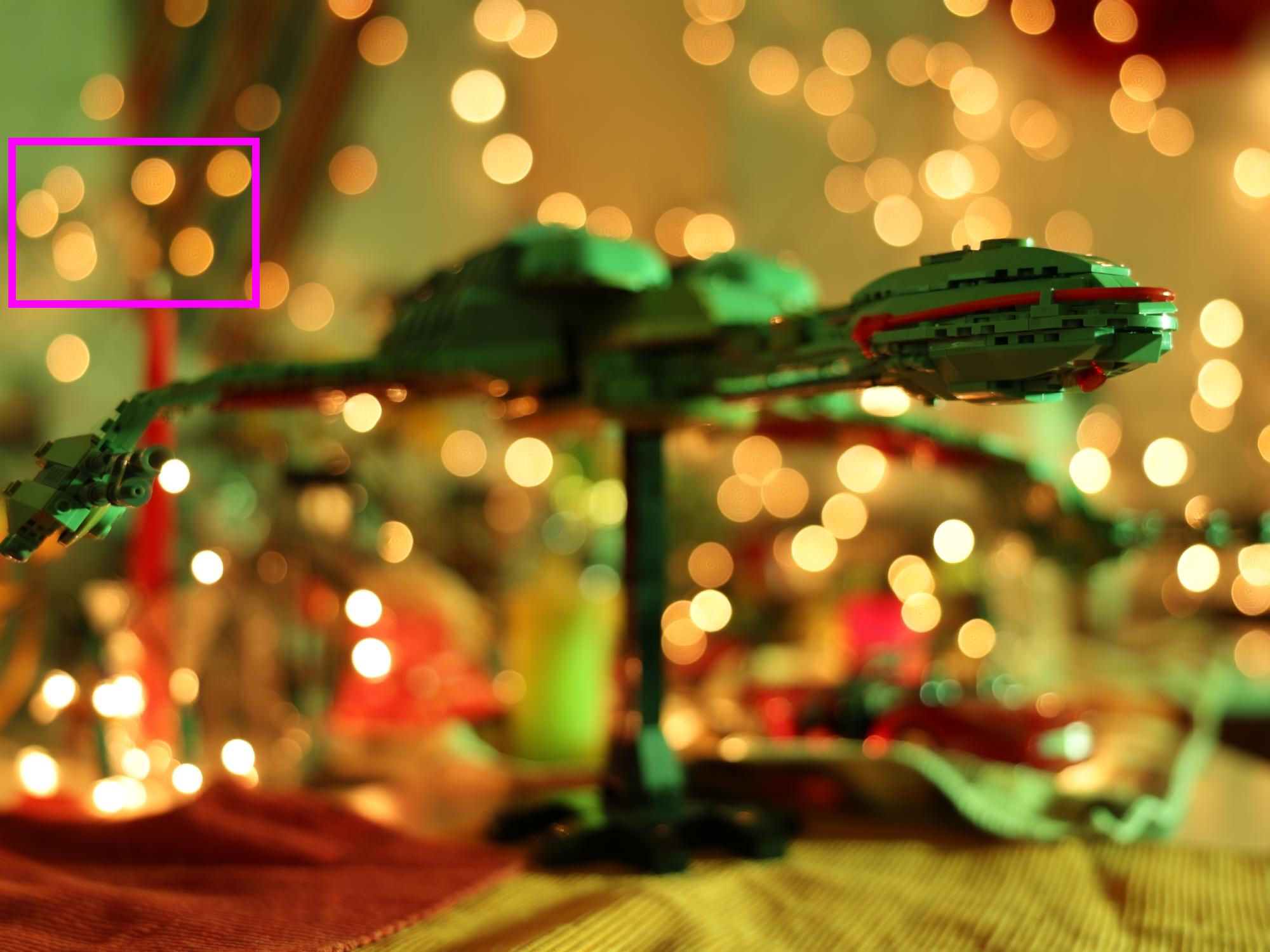}} &
        \multicolumn{2}{c}{\includegraphics[width=\widthcomp\linewidth, trim={0 280px 0 0},clip]{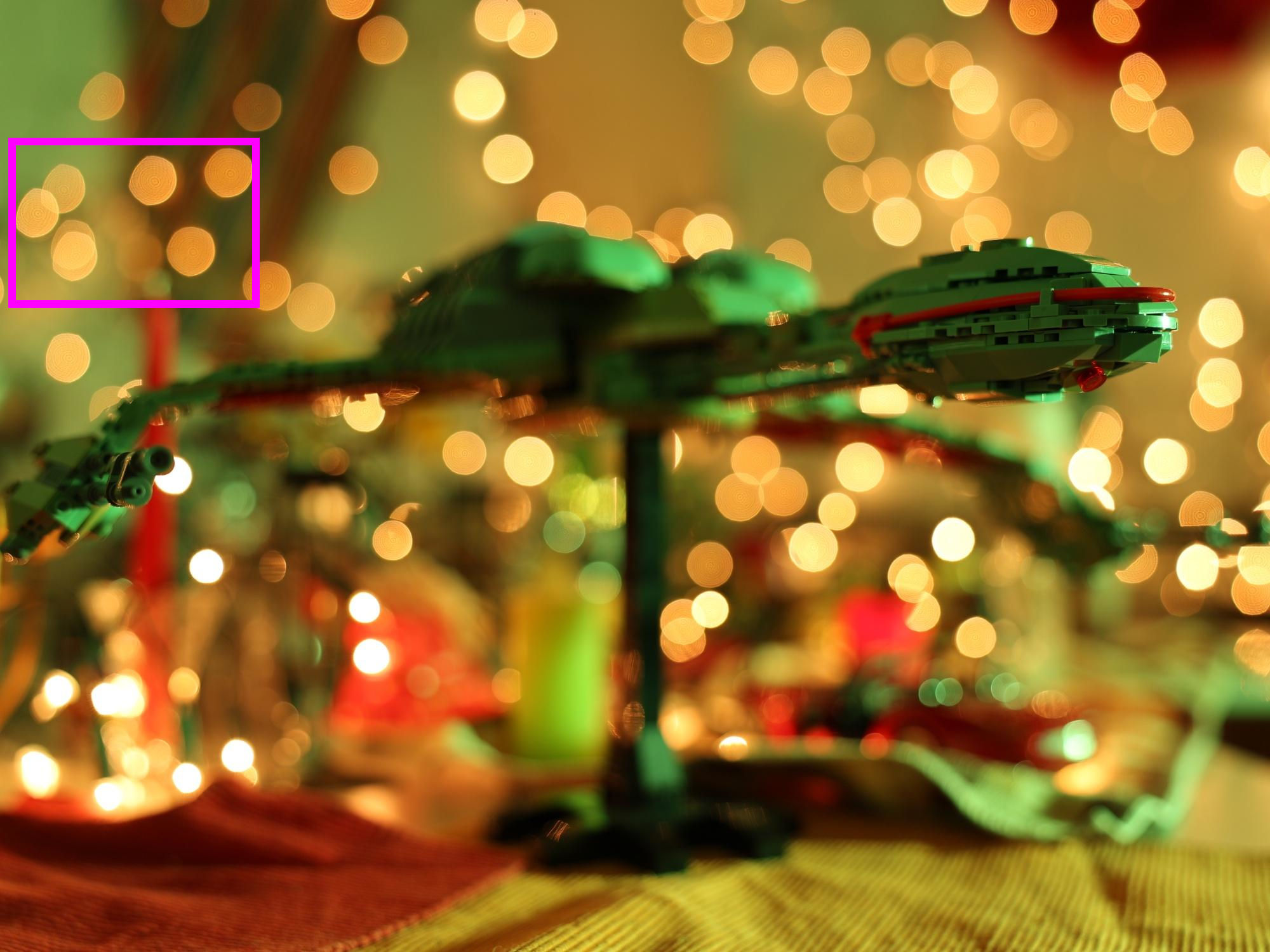}} \\
    \end{tabular}
\vspace{-2mm}
\caption{Comparison between the top methods of our Benchmark in \cref{tab:benchmark_bin}. Note how our method accuratly replicates the spatially varying nature of the Bokeh shapes, as illustrated by the \textit{pink} error map, while rendering sharper in-focus objects, indicated by the \textit{blue} error map.}
\label{fig:benchmark}
\vspace{-3mm}
\end{figure*}

\begin{table}[]
    \footnotesize
    \centering
    \small
    \setlength{\tabcolsep}{2pt}
    \renewcommand{\arraystretch}{0.8}
    \begin{tabular}{l|cc|ccc}
        \toprule
        Method                                & M.Para$\downarrow$ & GMACs$\downarrow$ & PSNR$\uparrow$ & SSIM$\uparrow$ & LPIPS$\downarrow$ \\
        \midrule
        Input                                 & -             & -                          & 19.689          & 0.6434           & 0.5193             \\
        \midrule
        \multicolumn{4}{l}{\hspace{1mm}\textbf{\textit{General Image Restoration Architectures}}} \\
        GRL\cite{li2023efficient}             & 3.19          & 195.41                  & 27.982          & 0.9126           & 0.1463             \\
        SwinIR~\cite{liang2021swinir}         & \textbf{0.91}          & 57.99                   & 28.851          & 0.9215           & 0.1272             \\
        MambaIR~\cite{guo2025mambair}       & 8.27          & 32.68                   & 28.952          & 0.9231           & 0.1253             \\
        NAFNet~\cite{chen2022simple}          & 17.11         & 15.96                   & 29.099          & 0.9193           & 0.1280             \\
        Restormer~\cite{zamir2022restormer}        & 26.13         & 141.24                 & 29.198          & 0.9231           & 0.1215             \\
        \midrule
        \multicolumn{4}{l}{\hspace{1mm}\textbf{\textit{Bokeh Rendering Architectures}}} \\
        D2F~\cite{luo2023defocus}         & 41.6         & \textit{7.52}                  & 25.977          & 0.8808           & 0.2255             \\
        BRViT~\cite{nagasubramaniam2023BEViT} & 123.15        & 54.04                   & 27.932          & 0.9159           & 0.1730             \\
        PyNET~\cite{ignatov2020rendering}     & 47.55         & 447.37                  & 28.891          & 0.9218           & 0.1279             \\
        DMSHN~\cite{dutta2021stackedbokeh}    & 10.85         & 45.57                   & 29.311          & 0.9246           & 0.1270             \\
        \midrule           \\
        \textbf{Ours}-M                       & \textit{1.21} & \textbf{5.93}           & \textit{30.358} & \textit{0.9263}  & \textit{0.1079}    \\
        \textbf{Ours}-L                       & 13.96         & 60.55                   & \textbf{31.250} & \textbf{0.9333}  & \textbf{0.1014}    \\ 
        \bottomrule
    \end{tabular}
    \caption{SOTA Bokeh Rendering and image restoration methods on \bindataset.
    MACs are calculated for $256 \times 256px$.}
    \label{tab:benchmark_bin}
    \vspace{-3mm}
\end{table}

\section{Experiments}
\label{sec:experiments}

We perform an extensive benchmark on \bindataset~by retraining nine models from their officially released code in \cref{subsec:benchmark}. 
These methods are SOTA Bokeh rendering works~\cite{luo2023defocus, nagasubramaniam2023BEViT, ignatov2020rendering, dutta2021stackedbokeh}, or from the closely-related Image Restoration (IR) field~\cite{li2023efficient, guo2025mambair, liang2021swinir, chen2022simple, zamir2022restormer}. 
As these restoration methods do not come with a loss tailored to Bokeh Rendering, we adopt ours~\cref{eq:loss} for a fair comparison.

Only a limited number of multi-stage Bokeh Rendering methods offer aperture control, all requiring depth maps. We compare with three SOTA approaches~\cite{sheng2024dr, dutta2021DBSIbokeh, BokehMeHybrid} on the full \dataset dataset and EBB400~\cite{BokehMeHybrid} in \cref{subsec:fstopbenchmark}.

For evaluation, we set the resolution to $2000 \times 1500px$. We report metrics for both image fidelity (PSNR, SSIM and LPIPS) and computational cost (M. Param., GMAC). Training details can be found in the supplementary material.

\subsection{Benchmark on Conventional Bokeh Rendering}
\label{subsec:benchmark}

As Bokeh rendering is underexplored, with few open-source methods, we are limited to retraining only a small number of competitors~\cite{ignatov2020rendering, dutta2021stackedbokeh, nagasubramaniam2023BEViT, luo2023defocus}.
As these proposals are relatively outdated, we additionally evaluate recent SOTA Image Restoration (IR) methods in our benchmark.
Recent works in IR explore Transformers \cite{liu2021swin, zamir2022restormer}, Self-Attention \cite{chen2022simple, li2023efficient}, and State-Space models \cite{guo2025mambair}.

In \cref{tab:benchmark_bin}, the performance of Ours-M is on par with SOTA methods, but notably reducing computational complexity.
Our scaled-up L model achieves higher fidelity with competitive computational complexity.

We find that good performance on IR datasets does not imply effectiveness in our challenging Bokeh rendering task. 
This may be due to IR methods often relying on spatial attention mechanisms designed to prioritize localized detail recovery.
However, Bokeh effects require a more broad approach to rendering, particularly in managing complex large blur kernels.
In contrast, our model incorporates tailored and aperture-aware attention. 
This allows us to achieve high fidelity with significantly lower computational cost.

In Figure~\ref{fig:benchmark}, we provide detailed comparisons on a challenging outdoor scene and a highly complex indoor scene. 
We can appreciate how our method preserves in-focus areas, while the other methods smooth and harm them while struggling to accurately reproduce the Bokeh effect.

\begin{table}[]
        \centering
        \renewcommand{\arraystretch}{0.8}
        \footnotesize
            \begin{tabular}{l|ccc}
            \toprule
            Method                                  & PSNR$\uparrow$ & SSIM$\uparrow$ & LPIPS$\downarrow$ \\
            \midrule
            Input                                   & 20.79          & 0.7097         & 0.4557 \\
            \midrule
            DDDF~\cite{purohit2019depth}            & 24.14          & 0.8713         & 0.2482 \\
            DBSI~\cite{dutta2021DBSIbokeh}          & 23.45          & 0.8675         & 0.2463 \\
            BGGAN~\cite{BokehGlassGAN}              & 24.39          & 0.8645         & 0.2467 \\
            DMSHN~\cite{dutta2021stackedbokeh}      & 24.72          & 0.8793         & 0.2271 \\
            MPFNet~\cite{wang2022self}              & 24.74          & 0.8806         & 0.2255 \\
            PyNet~\cite{ignatov2020rendering}       & \textit{24.93} & 0.8788         & 0.2219 \\
            BRViT~\cite{nagasubramaniam2023BEViT}   & 24.76          & \textit{0.8904}& 0.1924 \\
            AMPN~\cite{georgiadis2023adaptivebokeh} & 24.50          & 0.8847         & \textit{0.1718} \\
            \midrule
            \textbf{Ours}-M                                  & \textbf{25.02}          & \textbf{0.8915}& \textbf{0.1638} \\ 
            \bottomrule
            \end{tabular}
            \vspace{-1mm}
            \caption{Results on the \textit{EBB!}~\cite{ignatov2020rendering} \textit{Val294}~\cite{dutta2021depth} set.
            }
            \vspace{-3.5mm}
        \label{tab:ebb}
\end{table}

\noindent \textbf{Performance on EBB! Val294:}
We follow prior works on EBB!~\cite{ignatov2020rendering}~Val294~\cite{dutta2021stackedbokeh} by retraining our method from scratch on the corresponding training set and obtaining metrics via~\cite{nagasubramaniam2023BEViT}.
Our method outperforms the previous SOTA in \cref{tab:ebb}, while showing visible improvements over in \cref{fig:Ours_EBB}. Additional qualitative comparisons can be found in the supplementary material.

We observe that the rankings on the established EBB! Val294 benchmark differs from our new RealBokeh benchmark in \cref{tab:benchmark_bin}.
There are two factors at play here.
Due to its low alignment and image quality, the Val294 benchmark de-emphasizes retention of high-frequency detail for optimal performance. 
Conversely, as our dataset has a high image quality with very good alignment, this becomes more important. 
As the Bokeh effect should highlight detailed in-focus areas, this is naturally aligned with the aesthetic requirements of Bokeh in photography.

\begin{figure}[]
    \vspace{-1mm}
    \centering
    \small
    \setlength{\tabcolsep}{0.5pt}
    \renewcommand{\arraystretch}{0.3}
    \def\widthcomp{0.455}
    \begin{tabular}{cc}
        Input & BRViT~\cite{nagasubramaniam2023BEViT} \\
        \addlinespace[0.5pt]
        \includegraphics[width=\widthcomp\linewidth, trim={0 0px 0 0px},clip]{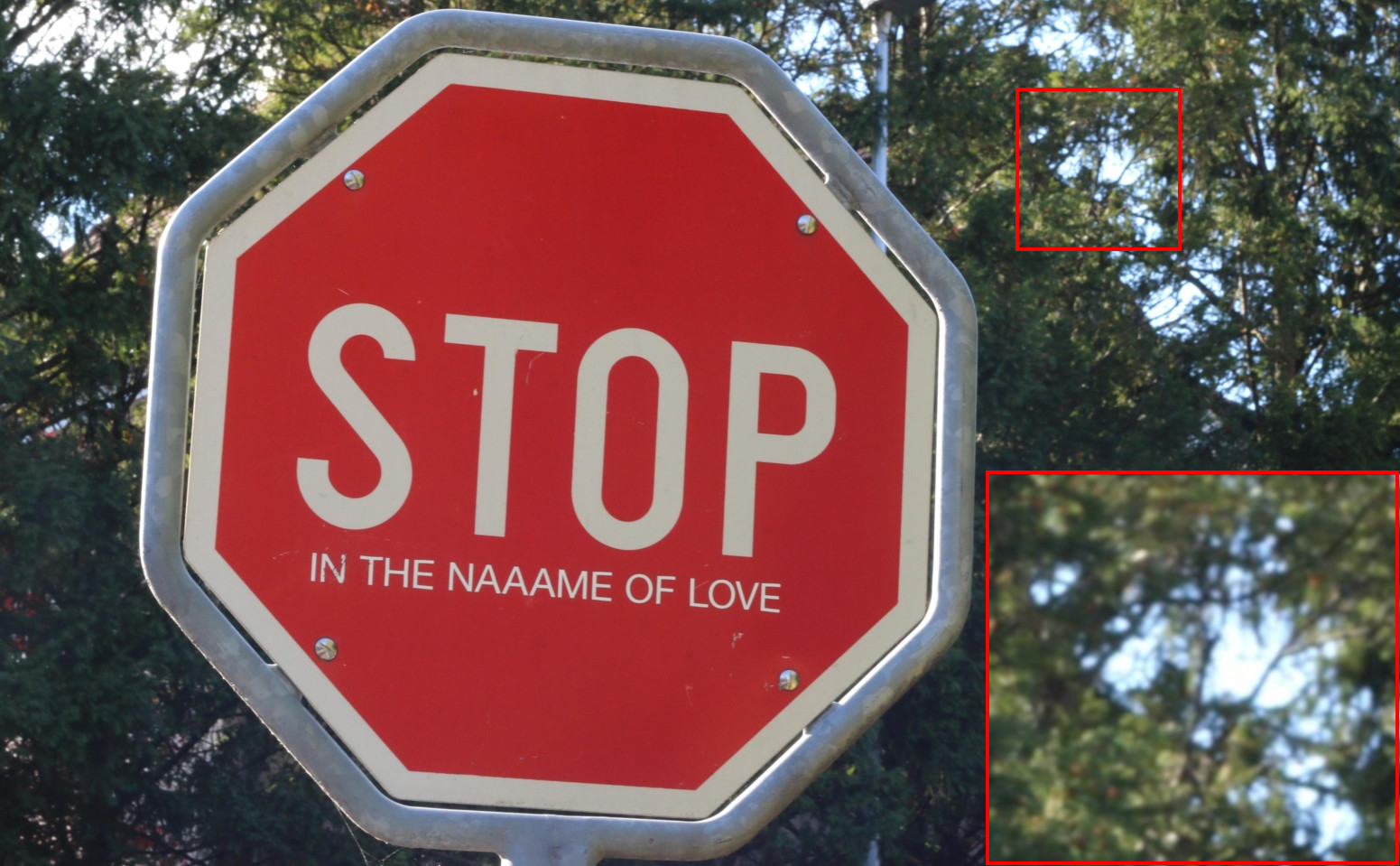} &
        \includegraphics[width=\widthcomp\linewidth, trim={0 0px 0 00px},clip]{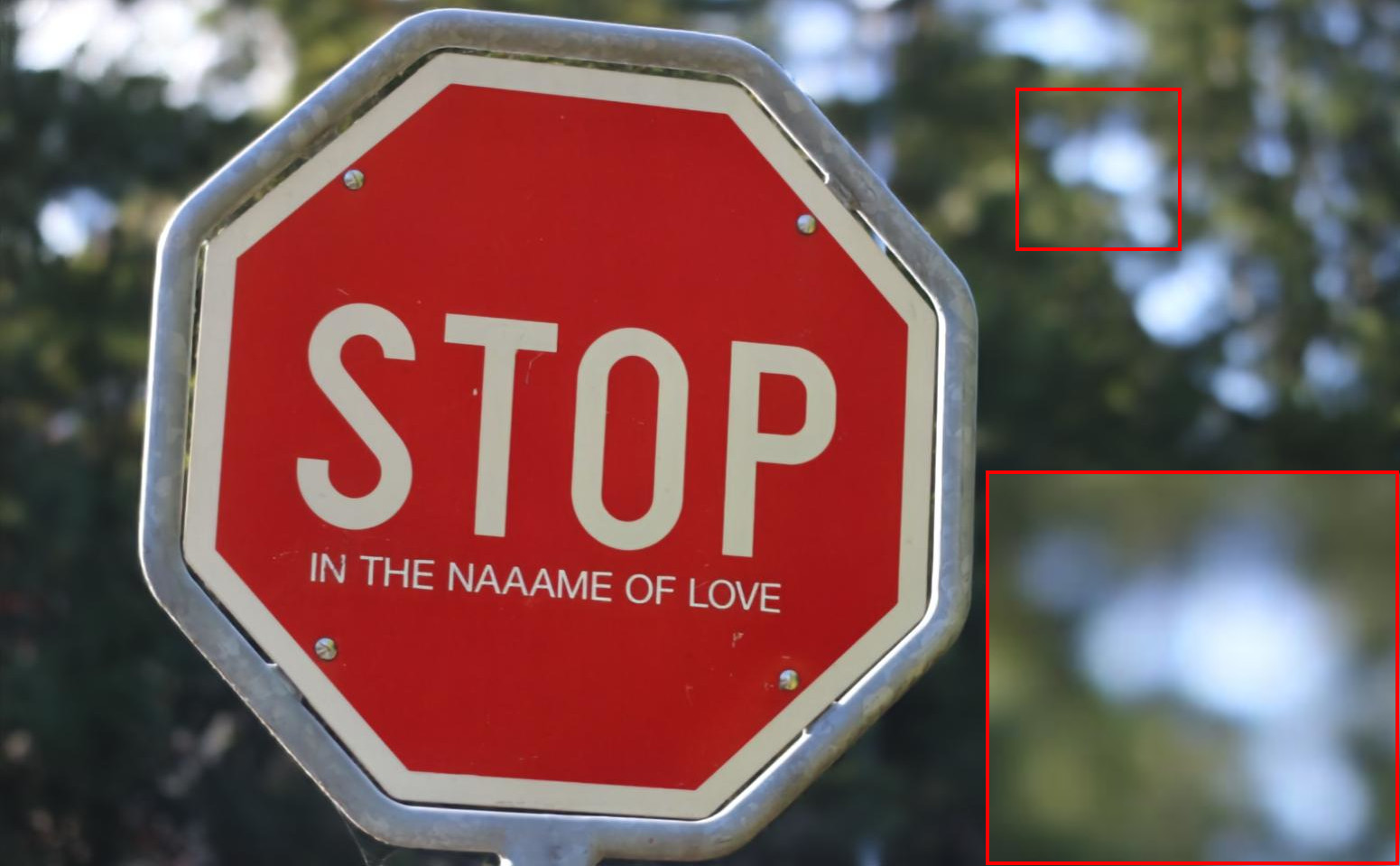} \\
        \includegraphics[width=\widthcomp\linewidth, trim={0 0px 0 00px},clip]{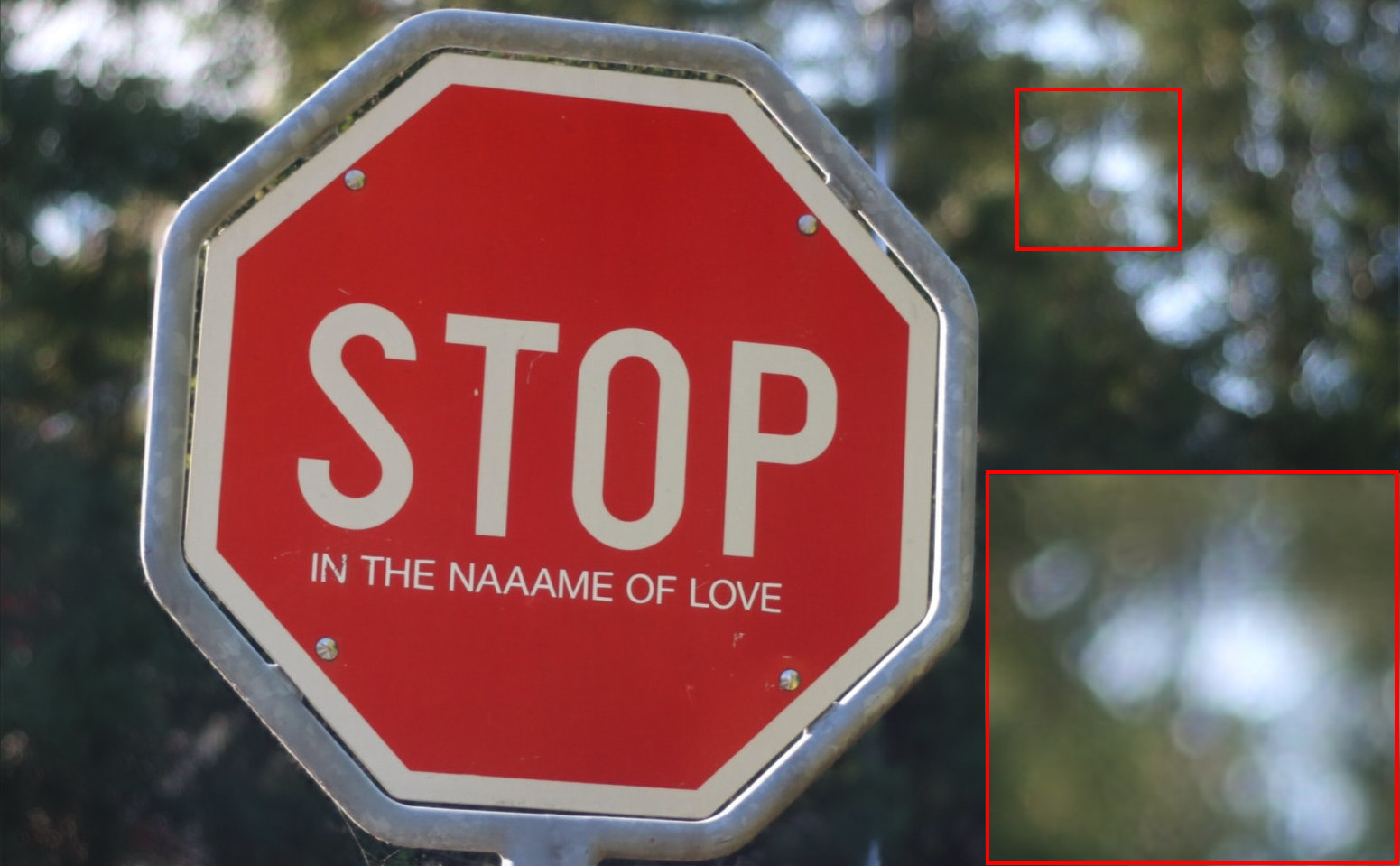} &
        \includegraphics[width=\widthcomp\linewidth, trim={0 0px 0 00px},clip]{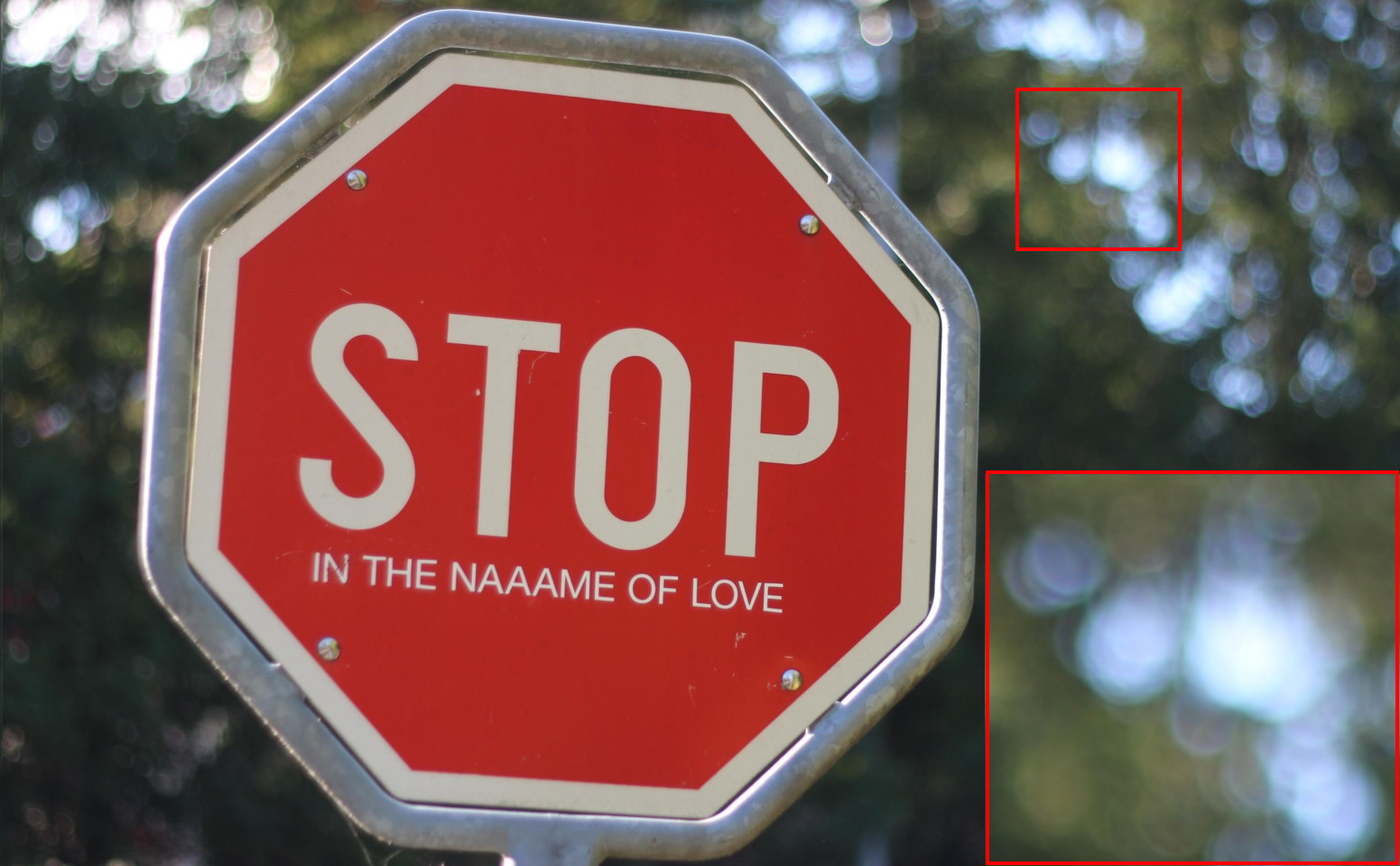} \\
        \textbf{Ours} & GT \\
    \end{tabular}
    \vspace{-1mm}
    \caption{Comparison on EBB! Val294 with the top performing open source method. Note how ours replicates the distinct Bokeh as seen in the GT while BRViT~\cite{nagasubramaniam2023BEViT} produces a uniform blur.}
    \label{fig:Ours_EBB}
    \vspace{-3mm}
\end{figure}

\subsection{Benchmark on Aperture Bokeh Rendering}
\label{subsec:fstopbenchmark}

When evaluated in the Bokeh Rendering task with aperture control, our method outperforms all competing multi-step approaches in \cref{tab:bokn_comp}.
This is done while maintaining a lower time complexity, even when excluding the potentially required calculation of image depth and etc.

\begin{figure}
    \centering
    \footnotesize
    \setlength{\tabcolsep}{0.5pt}
    \renewcommand{\arraystretch}{0.3}
    \def\widthcomp{0.312}
    \def\widthcompp{0.154}
    \def\hcomp{5.7mm}
    \begin{tabular}{ccccccc}
        &\multicolumn{2}{c}{BoMe~\cite{BokehMeHybrid}} & \multicolumn{2}{c}{\textbf{Ours}} & \multicolumn{2}{c}{GT} \\
        \addlinespace[0.5pt]
        \rotatebox{90}{\hspace{\hcomp} \fnum{2.0}} &
        \multicolumn{2}{c}{\includegraphics[width=\widthcomp\linewidth, trim={0 90px 0 0px},clip]{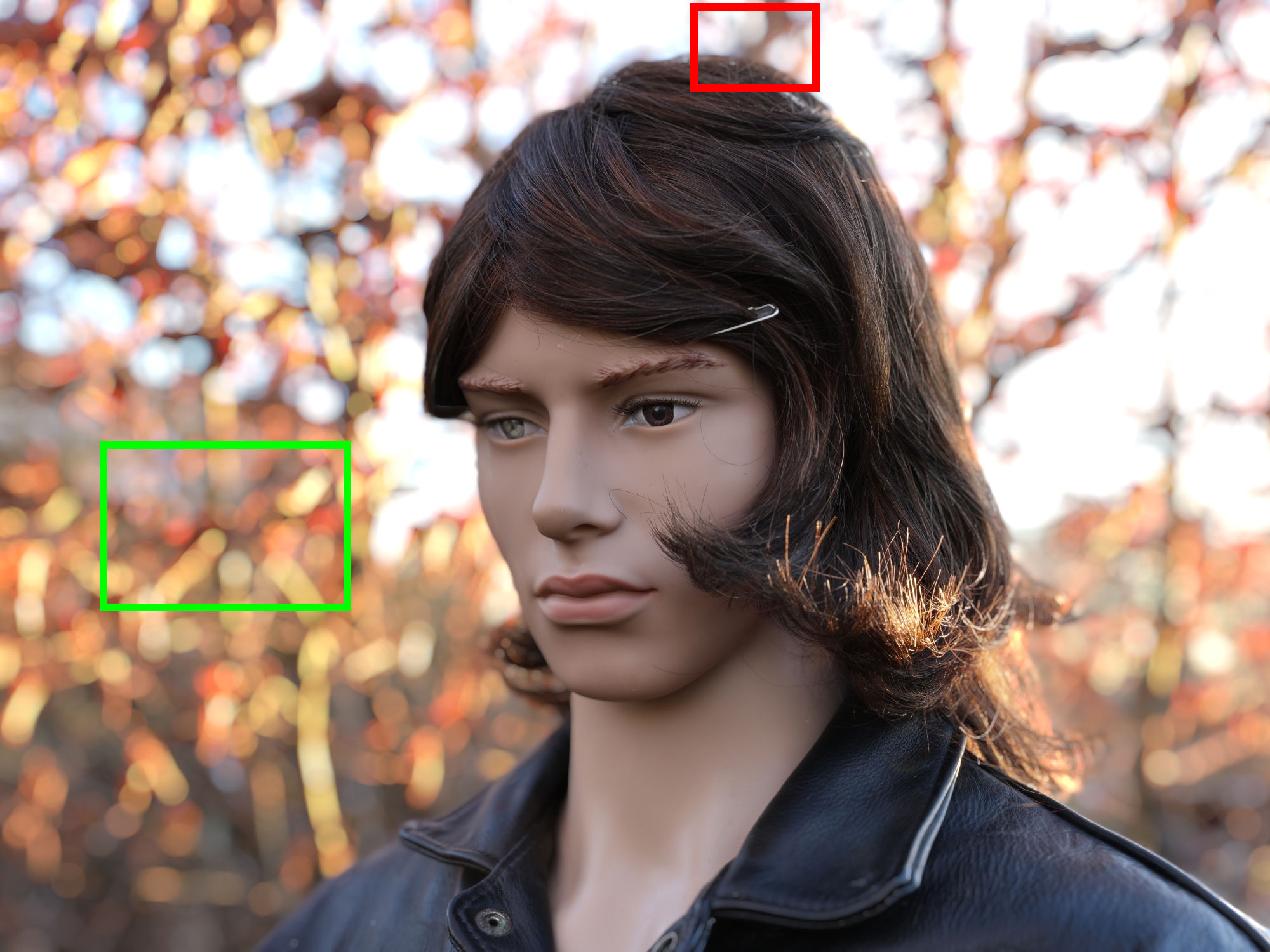}} &
        \multicolumn{2}{c}{\includegraphics[width=\widthcomp\linewidth, trim={0 90px 0 0px},clip]{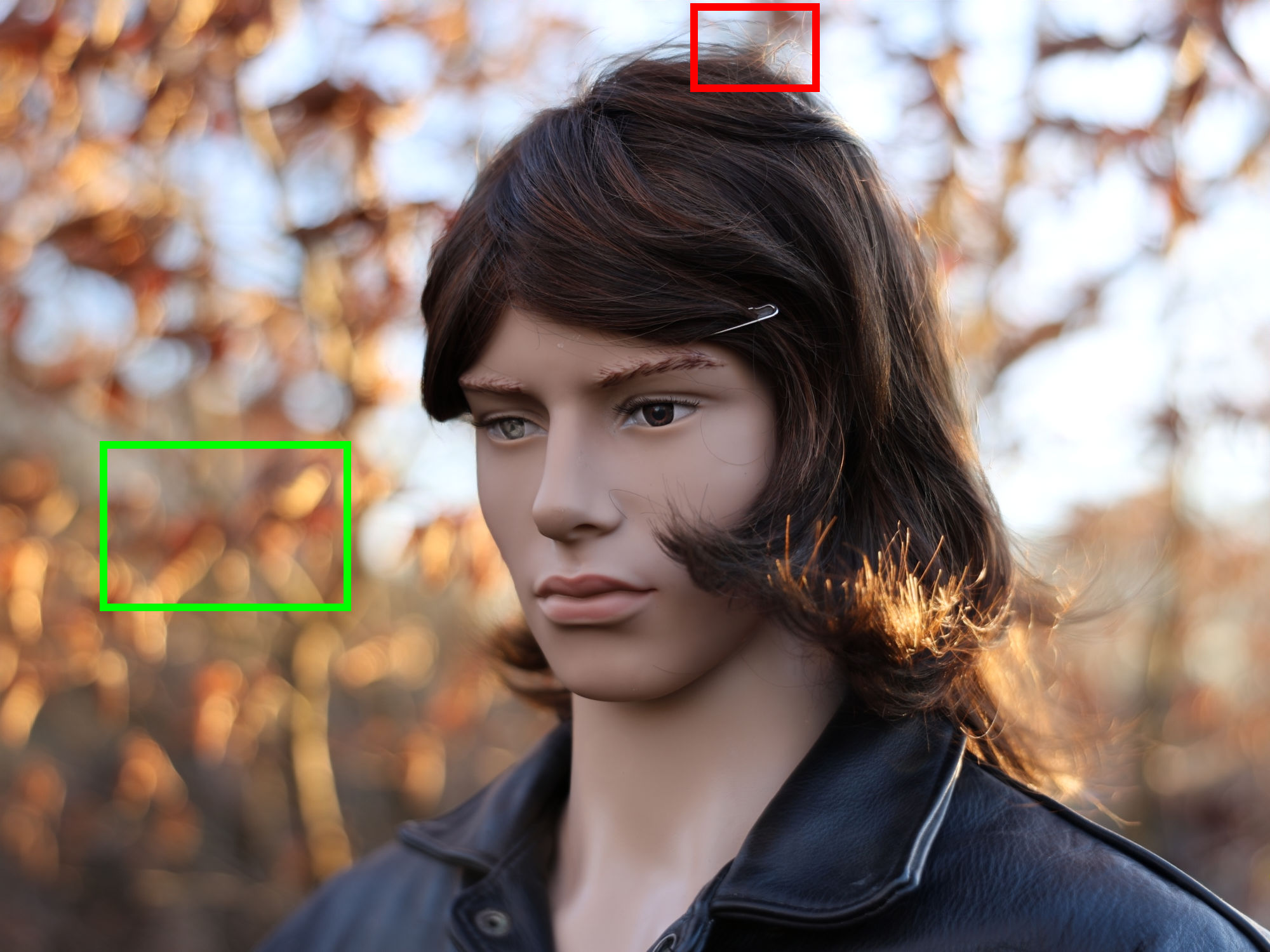}} &
        \multicolumn{2}{c}{\includegraphics[width=\widthcomp\linewidth, trim={0 90px 0 0px},clip]{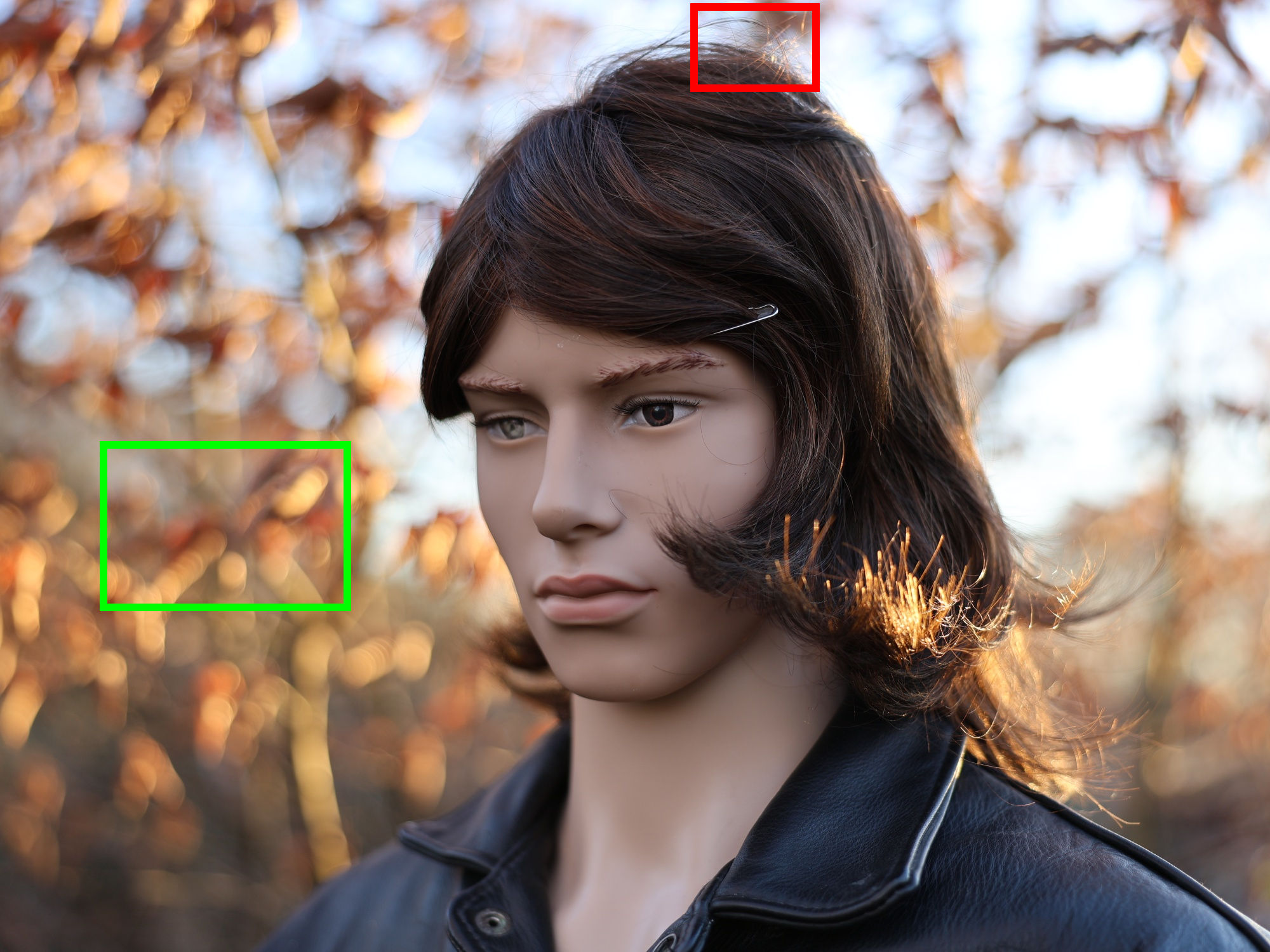}} \\
        &
        \includegraphics[width=\widthcompp\linewidth]{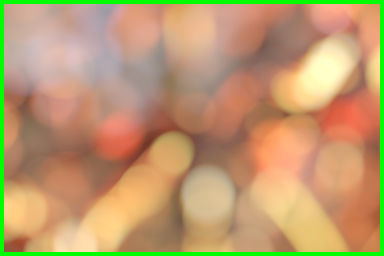} &
        \includegraphics[width=\widthcompp\linewidth]{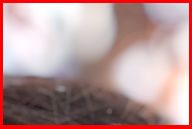} &
        \includegraphics[width=\widthcompp\linewidth]{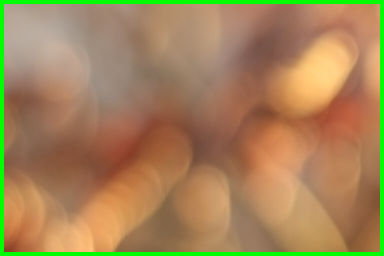} &
        \includegraphics[width=\widthcompp\linewidth]{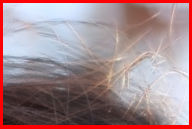} &
        \includegraphics[width=\widthcompp\linewidth]{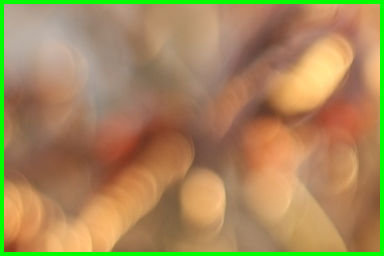} &
        \includegraphics[width=\widthcompp\linewidth]{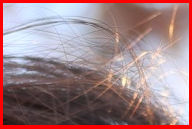} \\
        \rotatebox{90}{\hspace{\hcomp} \fnum{4.5}} &
        \multicolumn{2}{c}{\includegraphics[width=\widthcomp\linewidth, trim={0 90px 0 0px},clip]{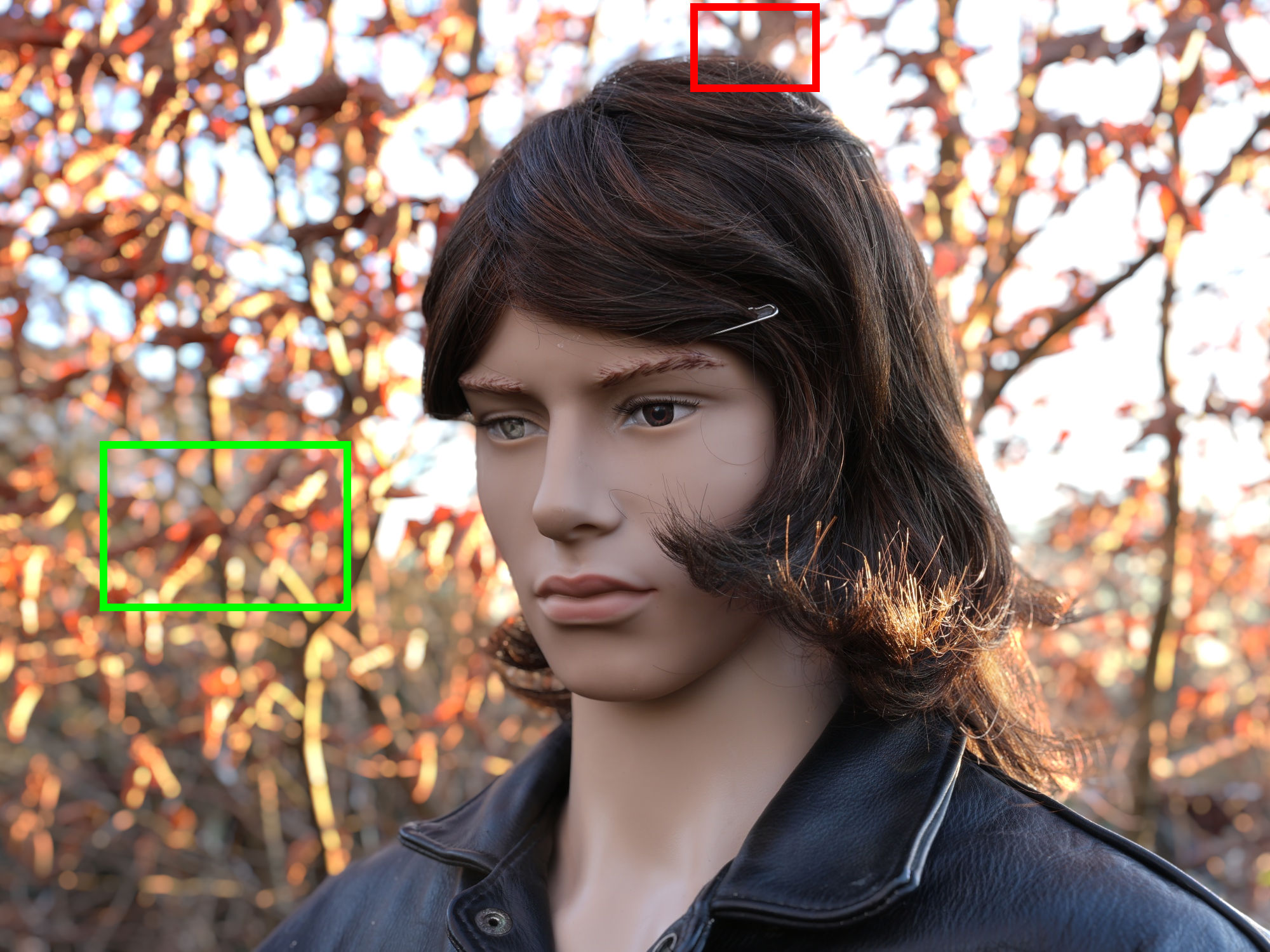}} &
        \multicolumn{2}{c}{\includegraphics[width=\widthcomp\linewidth, trim={0 90px 0 0px},clip]{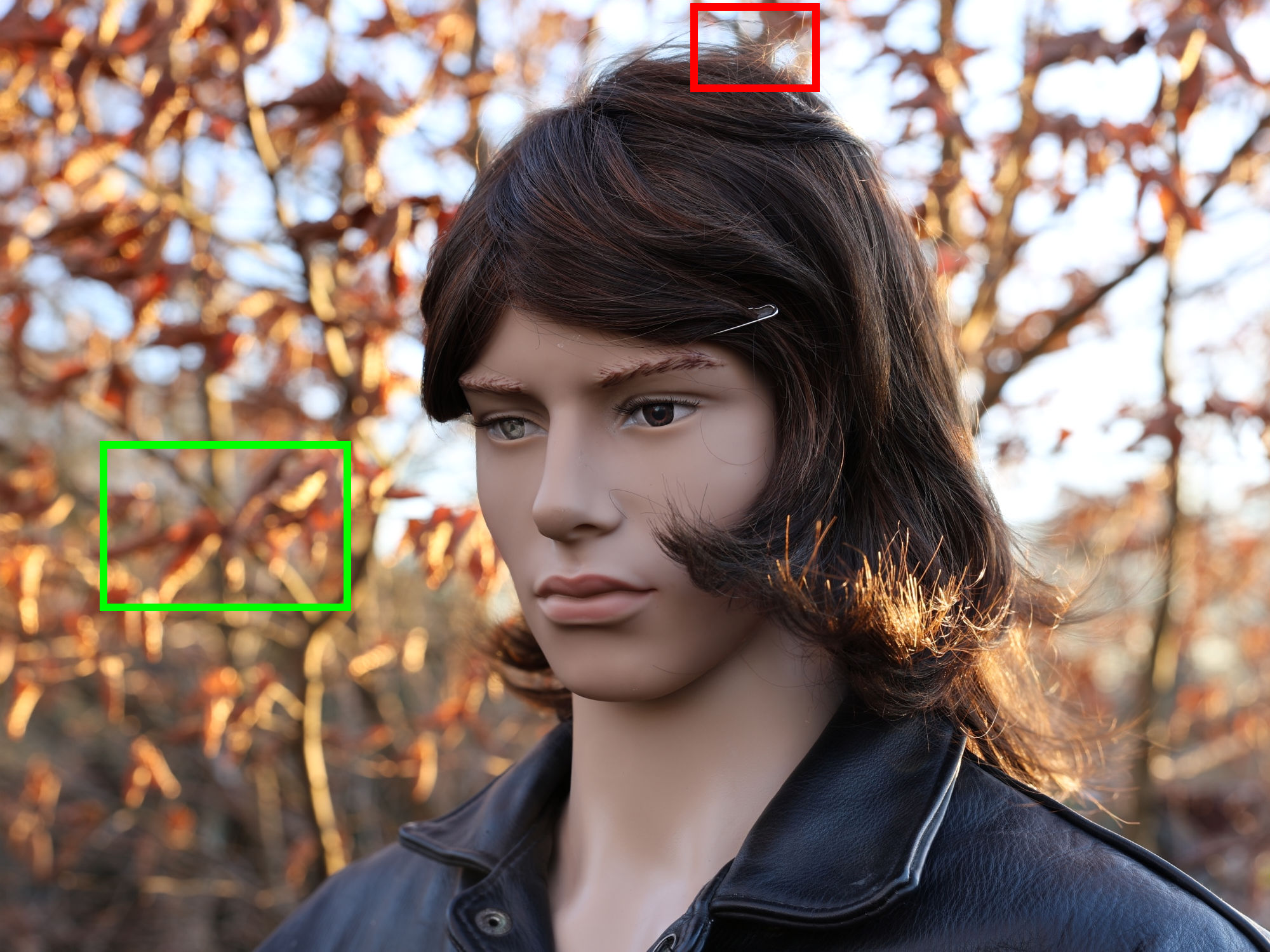}} &
        \multicolumn{2}{c}{\includegraphics[width=\widthcomp\linewidth, trim={0 90px 0 0px},clip]{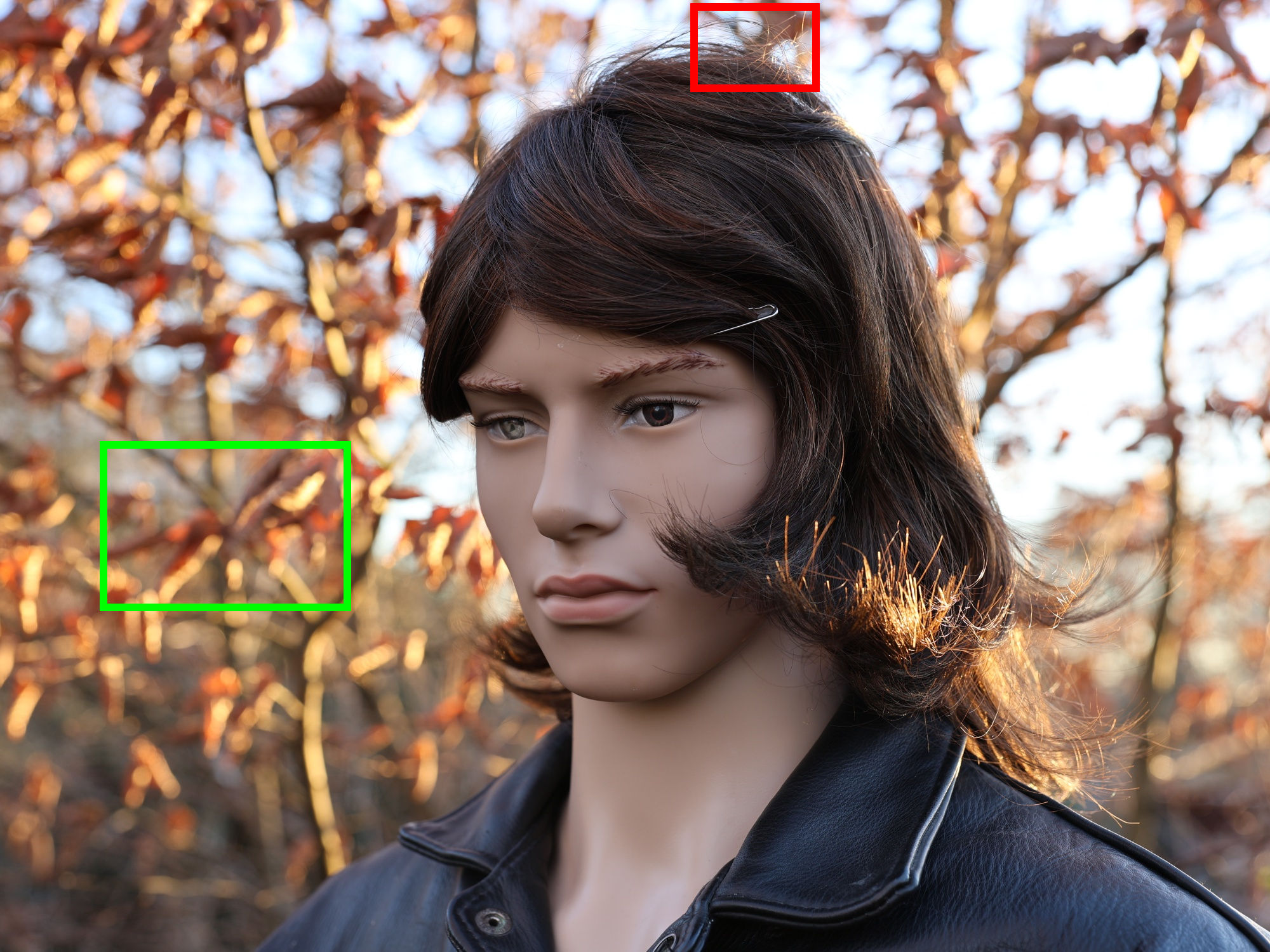}} \\
        &
        \includegraphics[width=\widthcompp\linewidth]{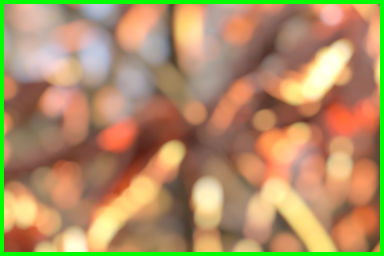} &
        \includegraphics[width=\widthcompp\linewidth]{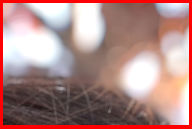} &
        \includegraphics[width=\widthcompp\linewidth]{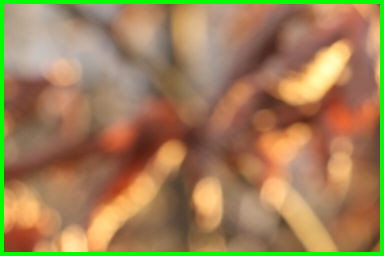} &
        \includegraphics[width=\widthcompp\linewidth]{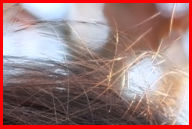} &
        \includegraphics[width=\widthcompp\linewidth]{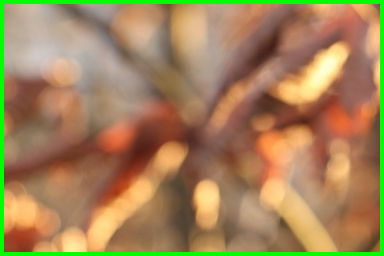} &
        \includegraphics[width=\widthcompp\linewidth]{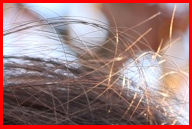} \\
        \rotatebox{90}{\hspace{\hcomp} \fnum{8.0}} &
        \multicolumn{2}{c}{\includegraphics[width=\widthcomp\linewidth, trim={0 90px 0 0px},clip]{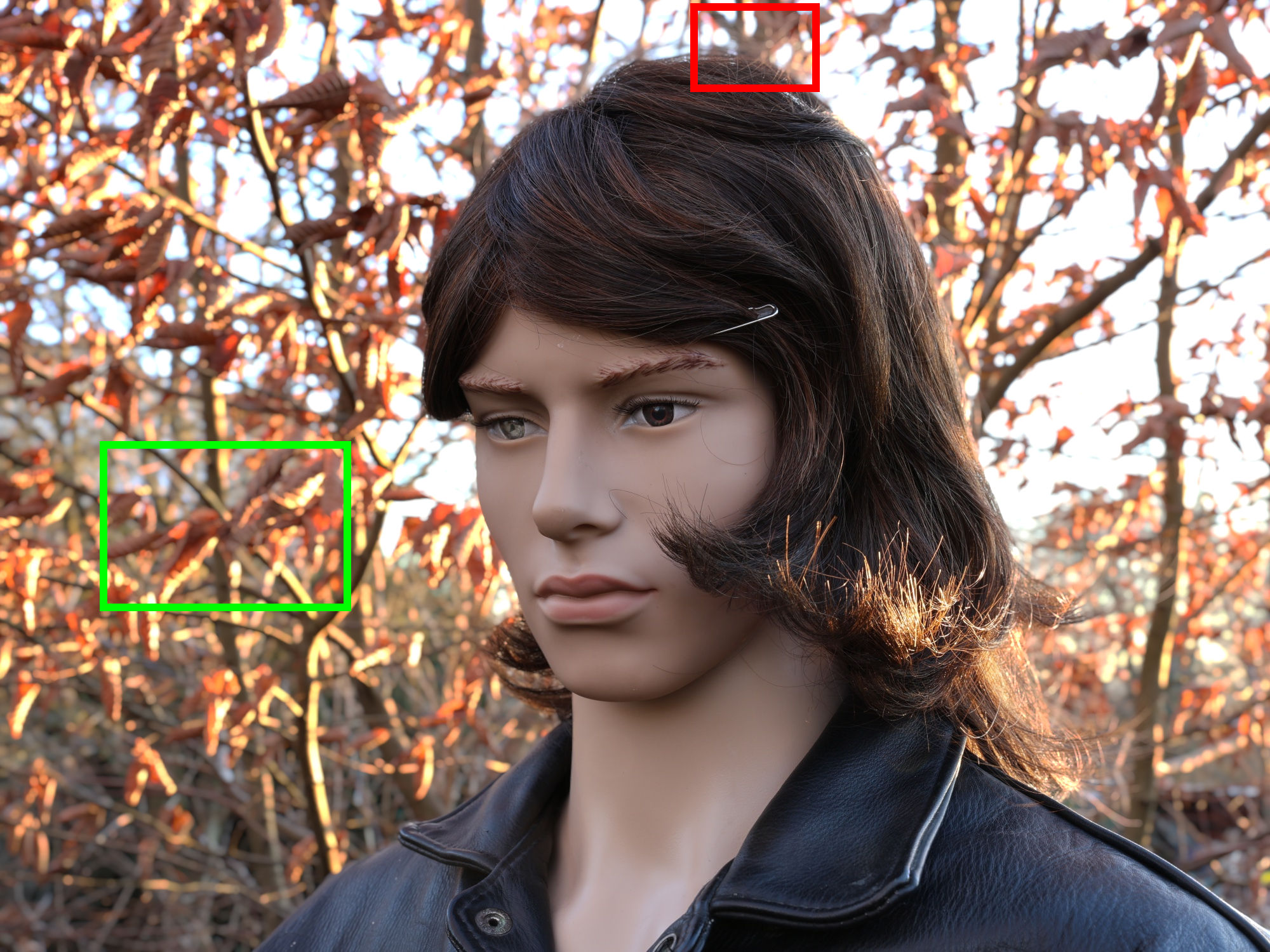}} &
        \multicolumn{2}{c}{\includegraphics[width=\widthcomp\linewidth, trim={0 90px 0 0px},clip]{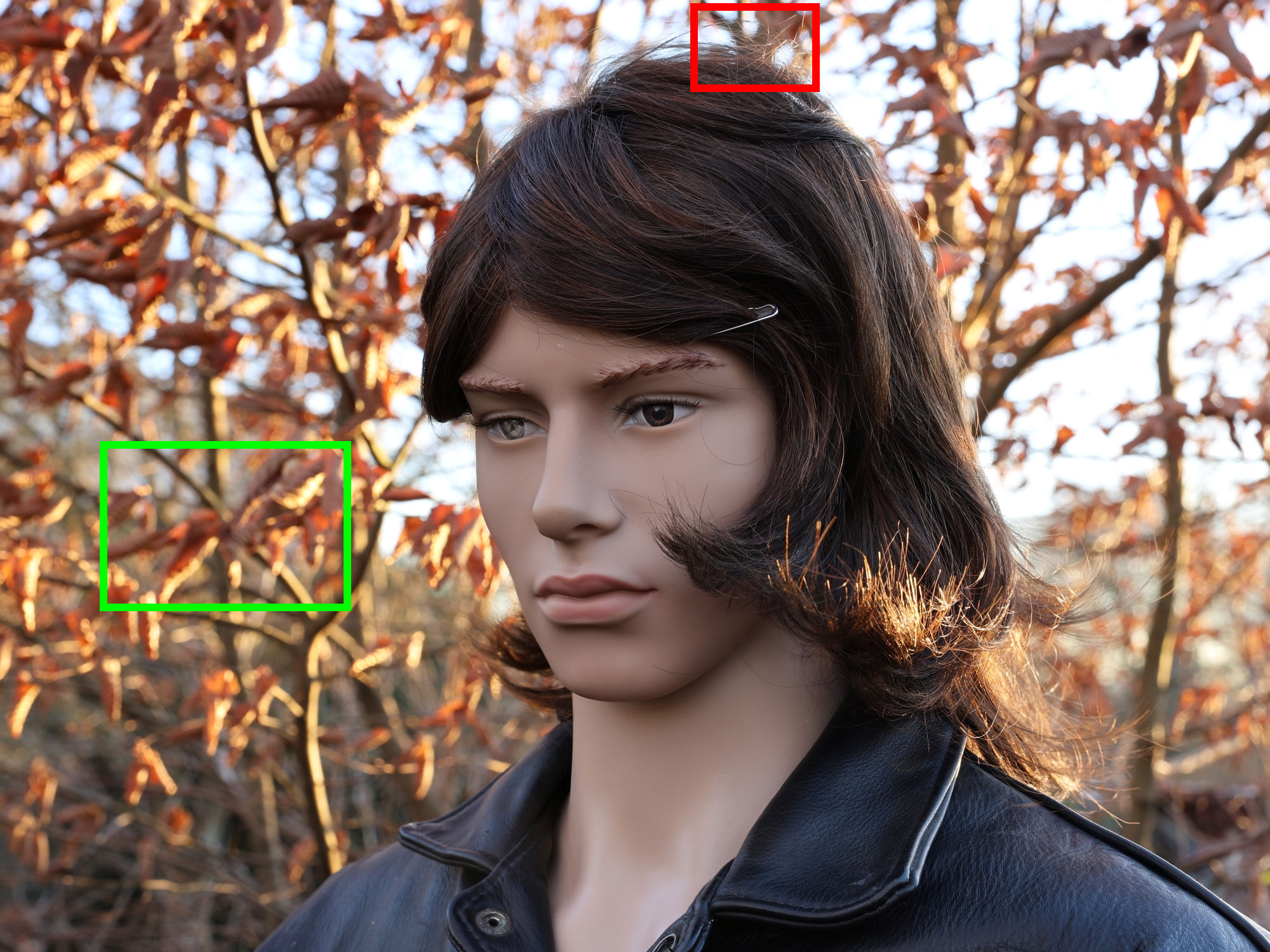}} &
        \multicolumn{2}{c}{\includegraphics[width=\widthcomp\linewidth, trim={0 90px 0 0px},clip]{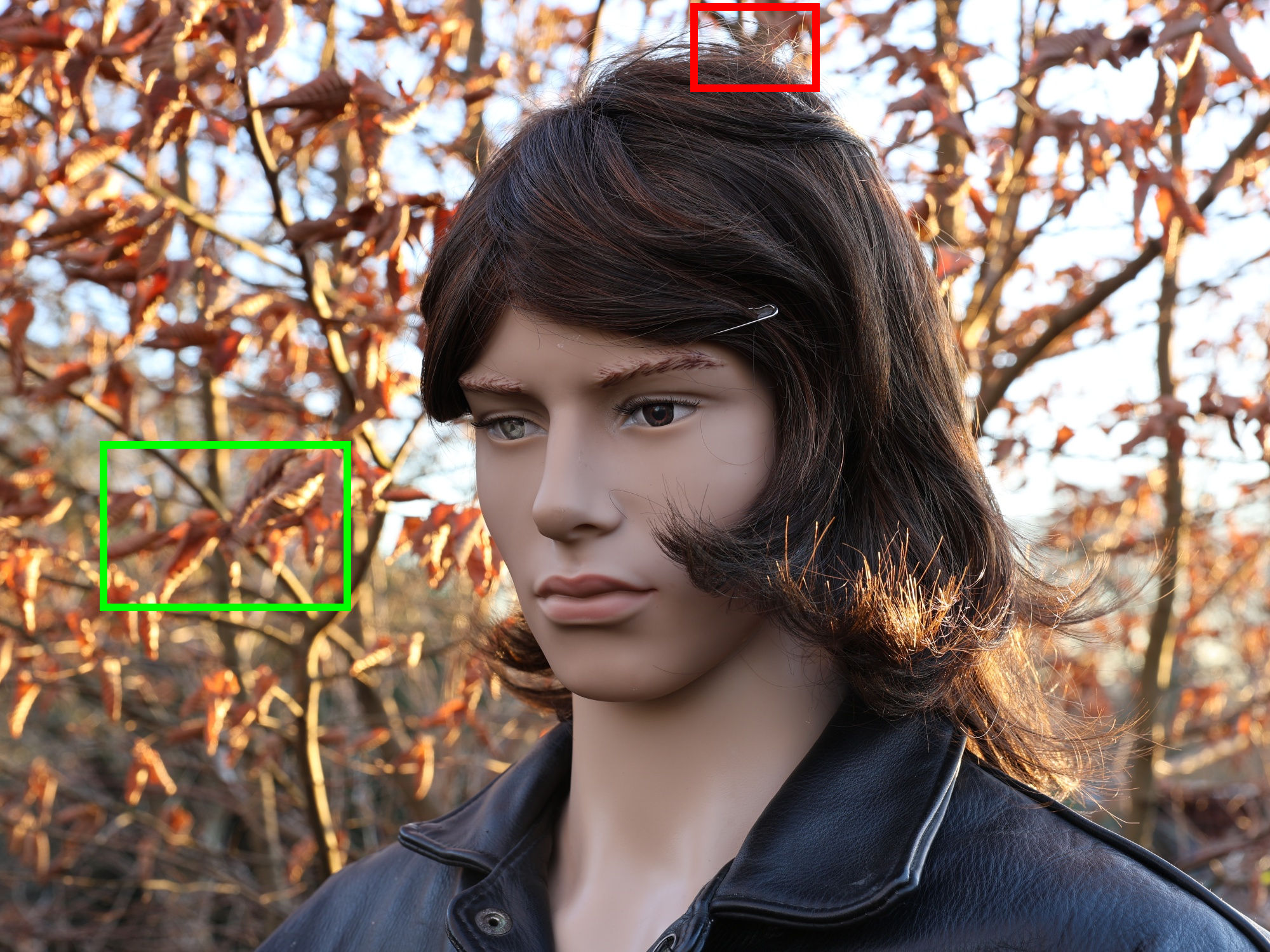}} \\
        &
        \includegraphics[width=\widthcompp\linewidth]{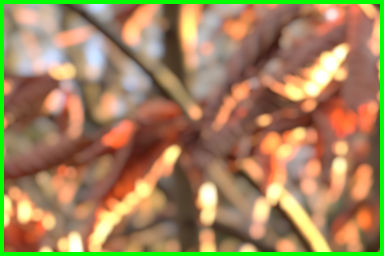} &
        \includegraphics[width=\widthcompp\linewidth]{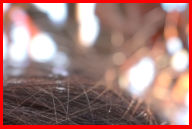} &
        \includegraphics[width=\widthcompp\linewidth]{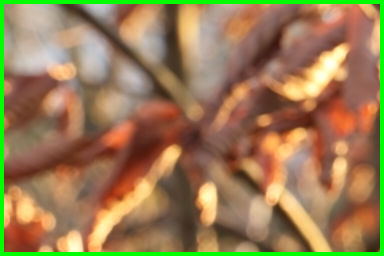} &
        \includegraphics[width=\widthcompp\linewidth]{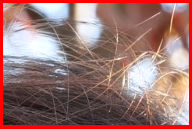} &
        \includegraphics[width=\widthcompp\linewidth]{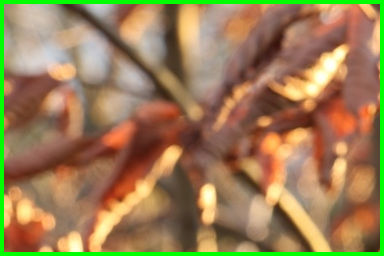} &
        \includegraphics[width=\widthcompp\linewidth]{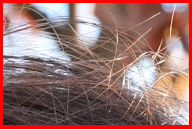} \\
    \end{tabular}
\vspace{-2mm}
\caption{Notice how our Bokeh accurately follows the reference at any \fstop while retaining more fine details than BoMe~\cite{BokehMeHybrid}.}
\label{fig:av_series}
\vspace{-2mm}
\end{figure}

In \cref{fig:av_series} we can see that in contrast to the best performing competitor BoMe~\cite{BokehMeHybrid}, our solution has a more natural color and Bokeh.
Unlike simulating the effect using hand-crafted rendering algorithms, our method implicitly learns to generate Bokeh from real exemplary images.
This enables us to produce a more accurate effect.
Additionally, our method retains fine foreground details and correctly separates the depth of multiple foliage layers in the background.

\begin{table}[]
\centering
\footnotesize
\setlength{\tabcolsep}{1.4pt}
\renewcommand{\arraystretch}{0.8}
\begin{tabular}{@{}l|c|cc|cc|cc@{}}
\toprule
  \multirow{2}{*}{Method} &
  time &
  \multicolumn{2}{c}{\fnum{2.0}} &
  \multicolumn{2}{c}{\fnum{4.0}} &
  \multicolumn{2}{|c}{all \fstops} \\
 &
  sec$\downarrow$ &
  PSNR$\uparrow$ &
  LPIPS$\downarrow$ &
  PSNR$\uparrow$ &
  LPIPS$\downarrow$ &
  PSNR$\uparrow$ &
  LPIPS$\downarrow$ \\ 
  \midrule
Input &
- &
  19.689 & 
  0.5193 & 
  21.092 & 
  0.4150 & 
  24.507 & 
  0.3278 \\ 
\midrule
Dr.B~\cite{sheng2024dr} &
  22.69 &
  25.265 & 
  0.2392 & 
  26.156 & 
  0.2053 & 
  27.716 & 
  0.1749 \\ 
MPIB~\cite{peng2022mpib} &
  1.76 &
  25.101 & 
  0.2466 & 
  25.778 & 
  0.2079 & 
  27.827 & 
  0.1751 \\ 
BoMe~\cite{BokehMeHybrid} &
  \textit{0.41} &
  26.143 & 
  0.2153 & 
  27.238 & 
  0.1787 & 
  29.125 & 
  0.1508 \\ 
\midrule
\textbf{Ours}-M &
  \textbf{0.28} &
  \textit{29.794} & 
  \textit{0.1158} & 
  \textit{31.790} & 
  \textit{0.0793} & 
  \textit{33.300} & 
  \textit{0.0694} \\ 
\textbf{Ours}-L &
  1.63 &
  \textbf{30.974} & 
  \textbf{0.1096} & 
  \textbf{32.901} & 
  \textbf{0.0770} & 
  \textbf{34.198} & 
  \textbf{0.0663} \\ 
\bottomrule
\end{tabular}
\vspace{-2mm}
\caption{\textbf{Performance on Real Bokeh} with aperture control. The \fstop columns indicate performance at the designated aperture, while the last column is the average performance on the entire dataset. The inference time was measured on RTX 4090 and excludes generation of auxiliary data for the competing methods. The full table can be found in the supplementary material.}
\vspace{-2mm}
\label{tab:bokn_comp}
\end{table}

\begin{figure}[]
    \centering
    \footnotesize
    \setlength{\tabcolsep}{0.4pt}
    \renewcommand{\arraystretch}{0.3}
    \def\widthcomp{0.32}
    \begin{tabular}{cccc}
        BoMe~\cite{BokehMeHybrid} & \textbf{Ours} & GT \\
        \addlinespace[0.7pt]
        \includegraphics[width=\widthcomp\linewidth]{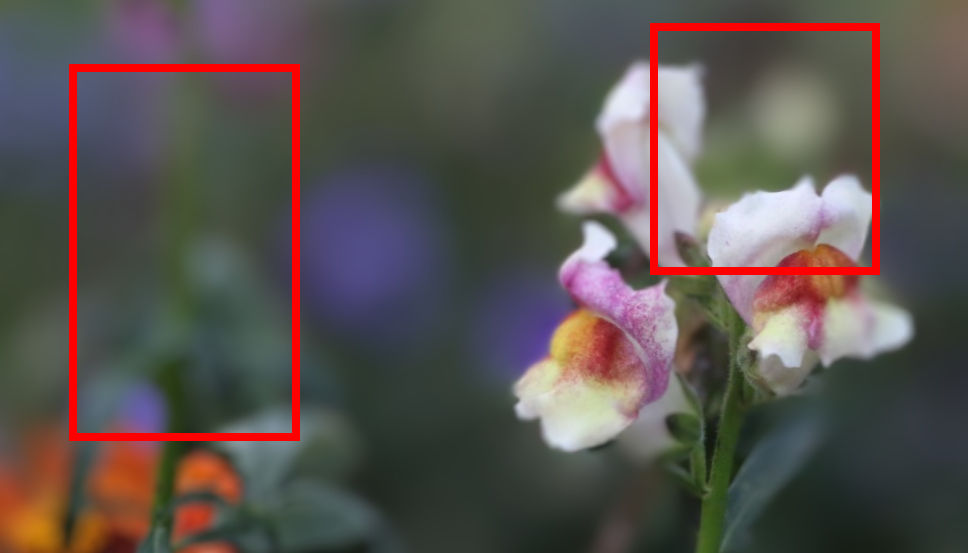} &
        \includegraphics[width=\widthcomp\linewidth]{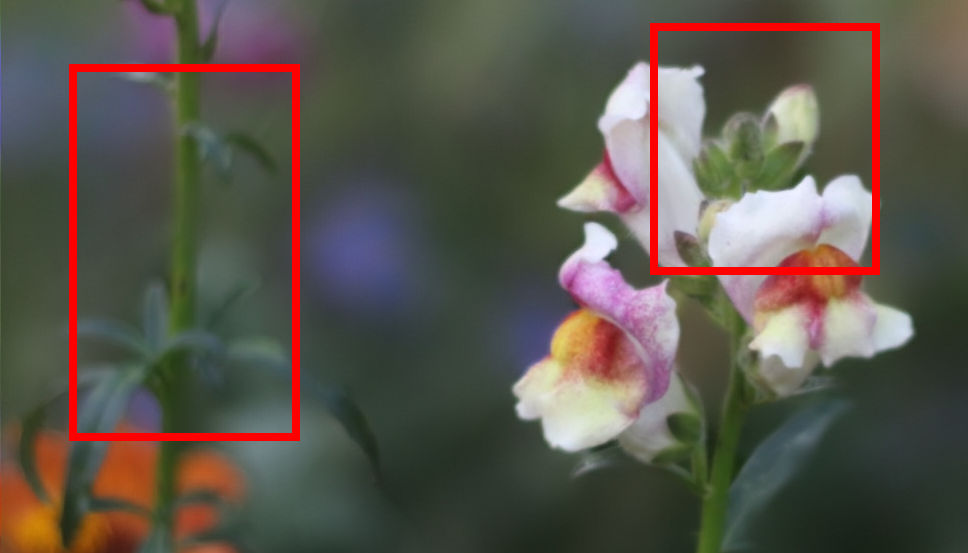} &
        \includegraphics[width=\widthcomp\linewidth]{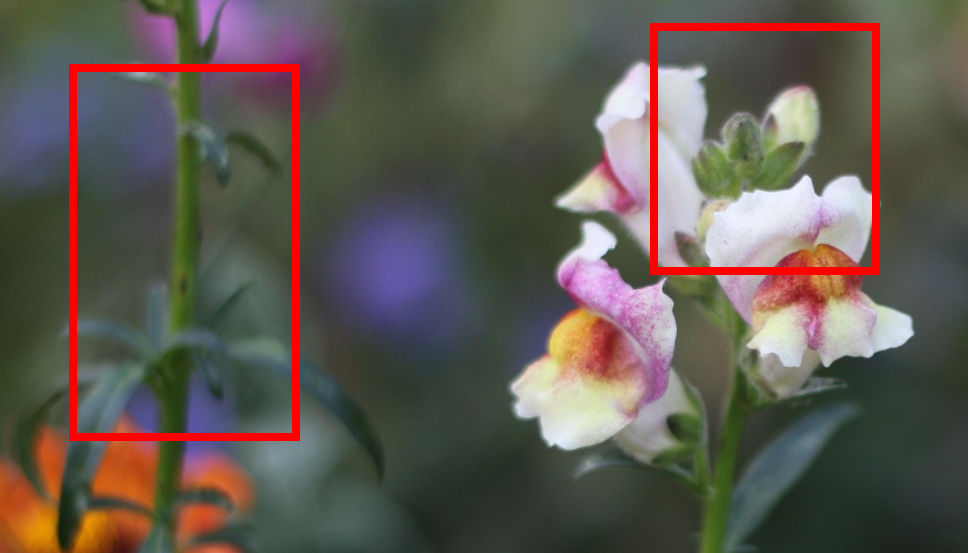} \\
        \includegraphics[width=\widthcomp\linewidth, trim={0 120px 0 25px},clip]{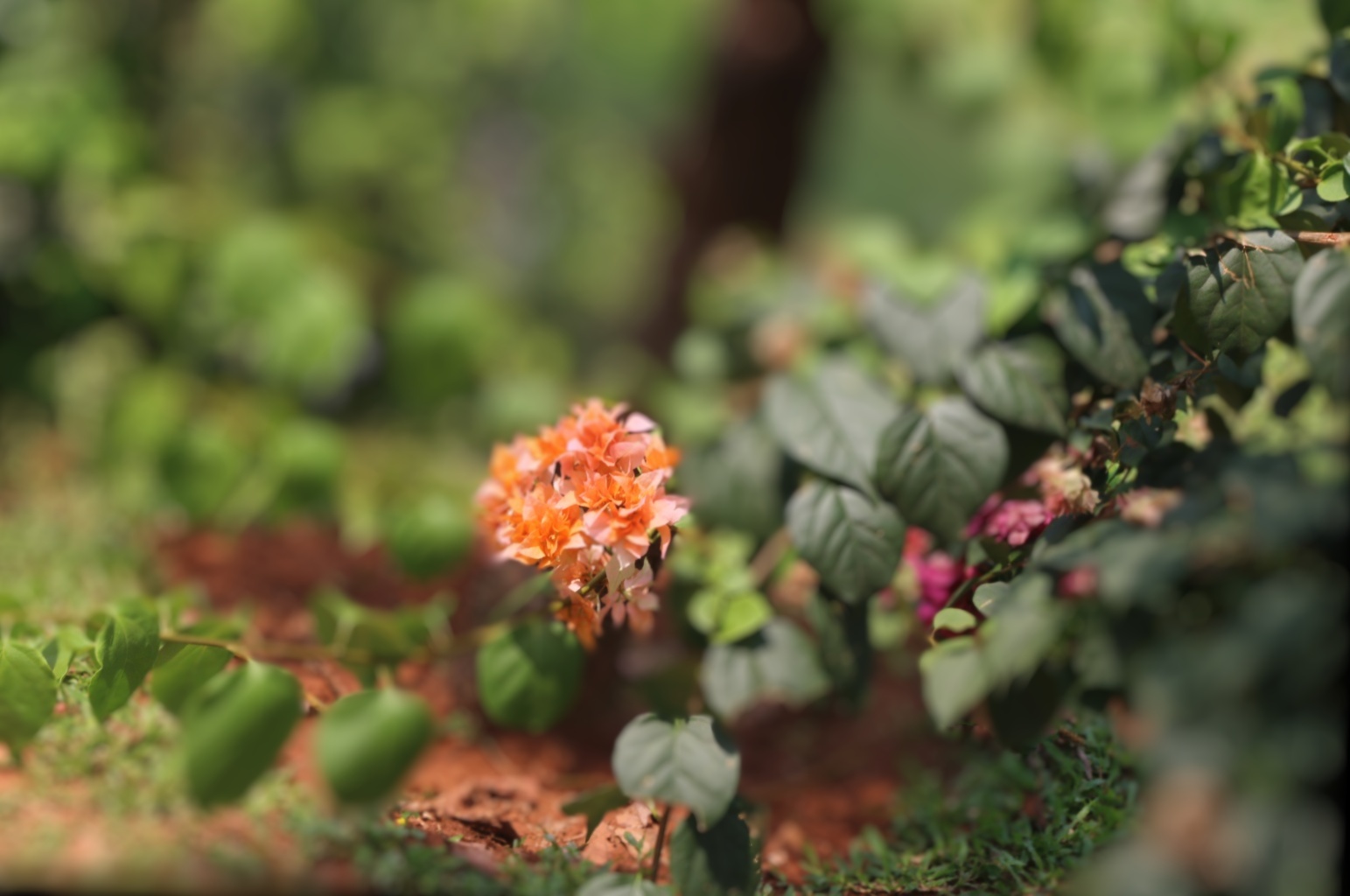} &
        \includegraphics[width=\widthcomp\linewidth, trim={0 120px 0 25px},clip]{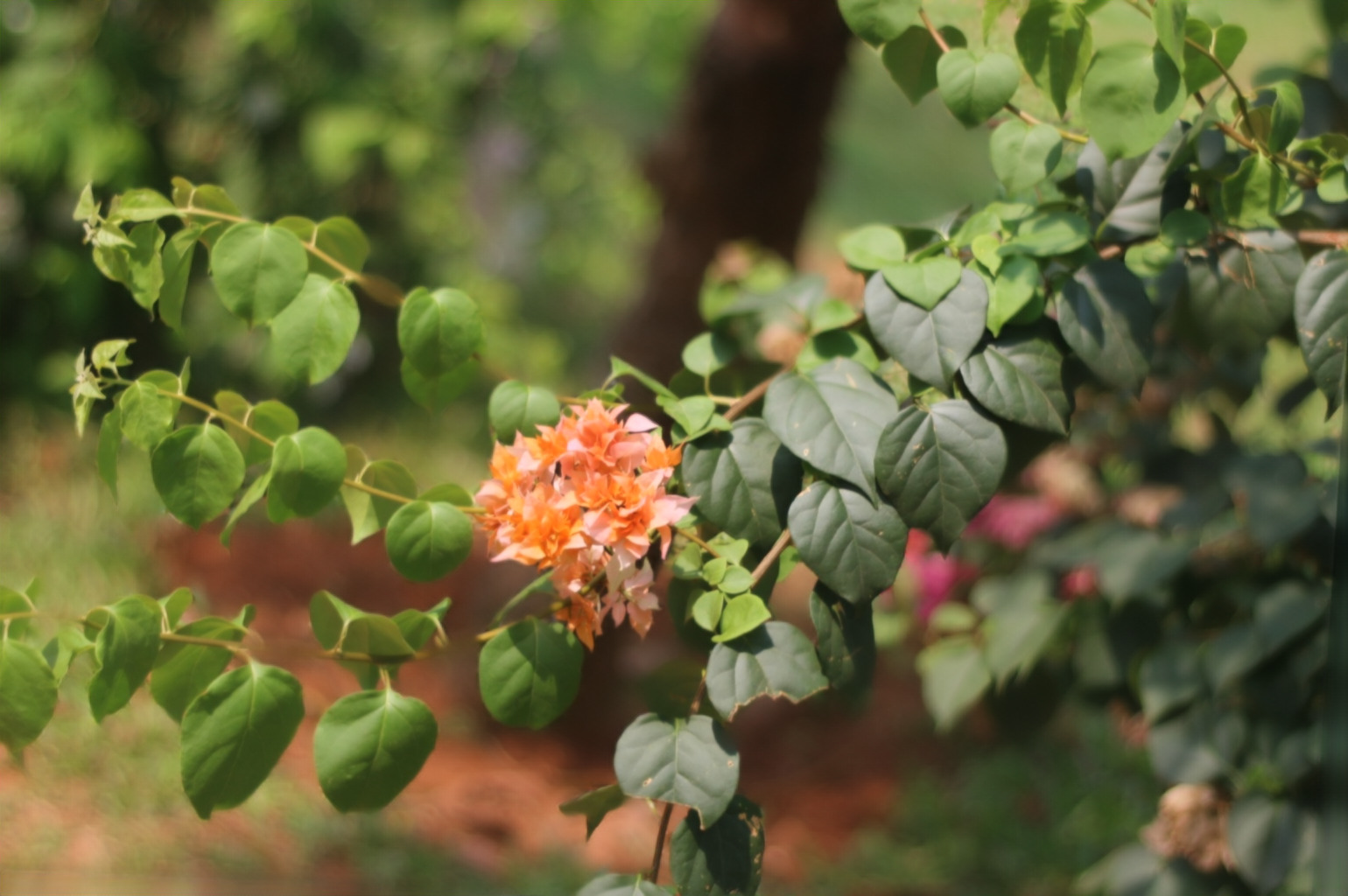} &
        \includegraphics[width=\widthcomp\linewidth, trim={0 120px 0 25px},clip]{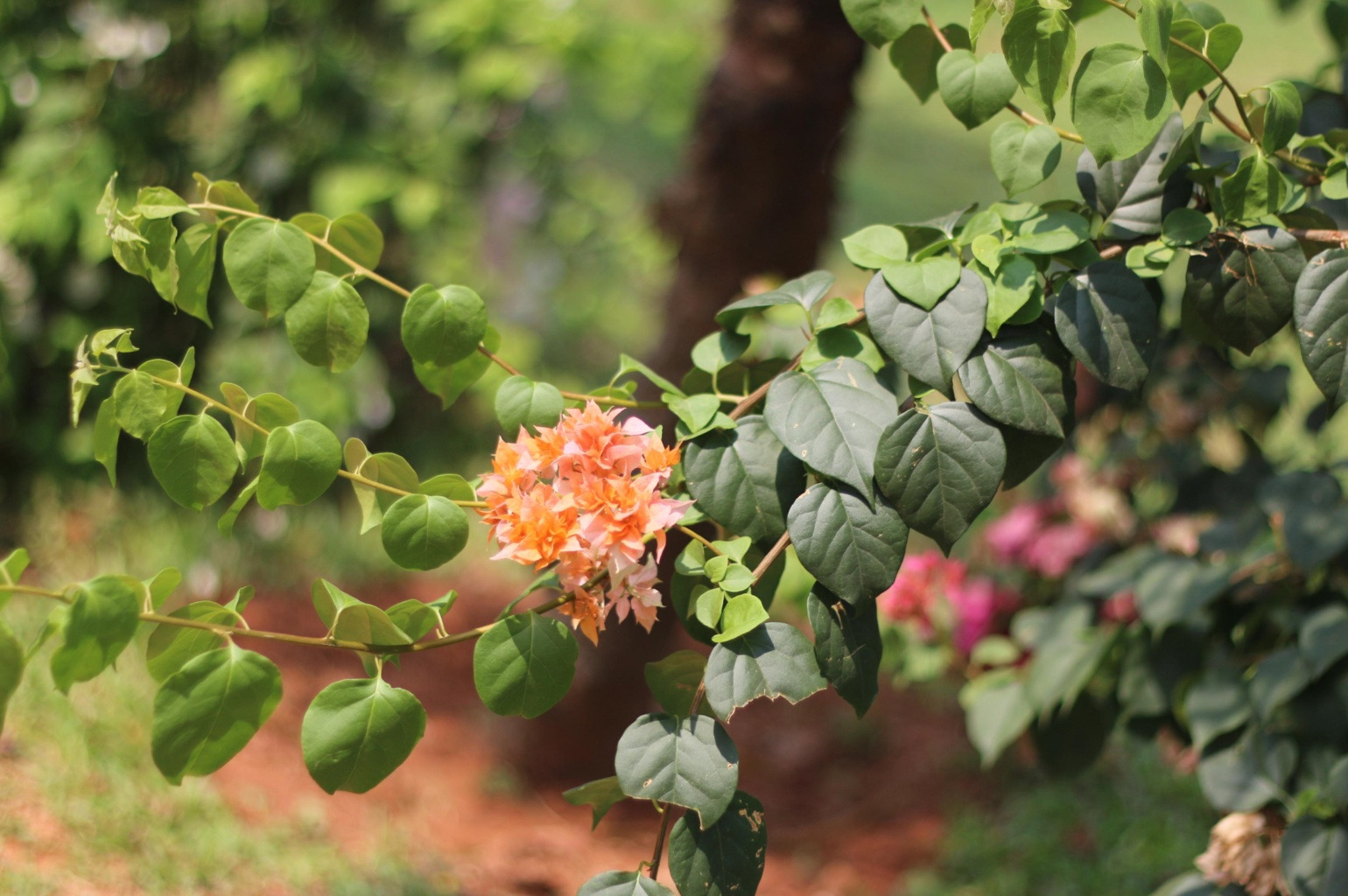} \\
    \end{tabular}
    \vspace{-2mm}
    \caption{Comparison on EBB400~\cite{BokehMeHybrid}. Note how our method suffers from fewer depth artifacts, particularly in the marked areas.}
    \label{fig:EBB400_vis}
    \vspace{-2mm}
\end{figure}

\noindent \textbf{Performance on EBB400:}
We further validate our strong results by applying our method to the established EBB400 benchmark~\cite{BokehMeHybrid}, where we compare against additional Bokeh rendering methods.
We train our method from \cref{tab:bokn_comp} on the remaining \textit{EBB!} samples for one epoch with a small learning rate of \textit{1e-6} to calibrate our lens model.
In \cref{tab:EBB400} our approach that purely \textit{learns} the Bokeh effect from \textit{real} images significantly outperforms the competing methods that are built around hand-crafted classical Bokeh synthesis by a large margin.
In \cref{fig:EBB400_vis} we achieve a visibly better foreground-background separation and a higher-fidelity Bokeh simulation than BoMe~\cite{BokehMeHybrid}.

\begin{table}[]
    \footnotesize
    \centering
    \setlength{\tabcolsep}{1.5pt}
    \renewcommand{\arraystretch}{0.8}
    \begin{tabular}{l|c|c|c|c|c|c}
	Method & RVR~\cite{zhang2019synthetic} & VDSLR~\cite{yang2016virtual} & DF~\cite{xiao2018deepfocus} & BoMe~\cite{BokehMeHybrid} & \textbf{Ours}-M \\
	\midrule
	PSNR$\uparrow$ & 23.56  & 23.78  & 23.81  & \textit{23.85}  & \textbf{24.47} \\
	SSIM$\uparrow$ & 0.8690 & 0.8738 & 0.8754 & \textit{0.8770} & \textbf{0.8829} \\
    \end{tabular}
    \vspace{-2mm}
    \caption{Quantitative results on EBB400~\cite{BokehMeHybrid}.}
    \vspace{-3.5mm}
    \label{tab:EBB400}
\end{table}

\subsection{Network Ablations}

\begin{table}[]
        \centering
        \setlength{\tabcolsep}{3pt}
        \renewcommand{\arraystretch}{0.6}
        \footnotesize
            \begin{tabular}{rl|cccc|c}
            \toprule
            & &
            \fnum{2.0} &
            \fnum{2.8} &
            \fnum{4.0} &
            \fnum{8.0} &
            all \\
            &
            Method & 
            PSNR$\uparrow$ & 
            PSNR$\uparrow$ & 
            PSNR$\uparrow$ & 
            PSNR$\uparrow$ &
            PSNR$\uparrow$ \\
            \midrule
            \multicolumn{2}{l|}{\textbf{\textit{Attentions}}} & \multicolumn{4}{c|}{} \\
            a) & Swin~\cite{liu2021swin} & 
            28.570 & 
            29.801 & 
            30.694 & 
            33.707 & 
            32.242 \\ 
            b) & NAT~\cite{Hassani_2023_CVPR} &
            29.569 & 
            30.715 & 
            31.399 & 
            34.450 & 
            32.940 \\ 
            \midrule
            \multicolumn{2}{l|}{\textbf{\textit{AAA Ablation}}} & \multicolumn{4}{c|}{} \\
            1) & maskless & 
            27.557 & 
            28.202 & 
            29.042 & 
            32.807 & 
            31.271 \\ 
            2) & single-mask &
            29.222 & 
            29.953 & 
            30.740 & 
            34.419 & 
            32.708 \\ 
            3) & multi-mask &
            \textit{29.790} & 
            \textit{30.665} & 
            \textit{31.659} & 
            \textit{34.830} & 
            \textit{33.224} \\ 
            4) & \textit{f}-aware & 
            \textbf{29.794} & 
            \textbf{31.101} & 
            \textbf{31.790} & 
            \textbf{34.994}  & 
            \textbf{33.300} \\ 
            \bottomrule
            \end{tabular}
            \vspace{-2mm}
            \caption{Results of ablation on attention choice and design.}
            \vspace{-3mm}
        \label{tab:ablation}
\end{table}

\begin{figure}
    \centering
    \footnotesize
    \setlength{\tabcolsep}{0.5pt}
    \renewcommand{\arraystretch}{0.3}
    \def\widthcomp{0.328}
    \def\widthcompp{0.1618}
    \begin{tabular}{cccccc}
        \multicolumn{2}{c}{Input} & \multicolumn{2}{c}{Syn-DoF~\cite{wadhwa2018synthetic}} & \multicolumn{2}{c}{\textbf{Ours \fnum{3.2}}} \\
        \addlinespace[0.5pt]
        \multicolumn{2}{c}{\includegraphics[width=\widthcomp\linewidth, trim={0 15px 0 8px},clip]{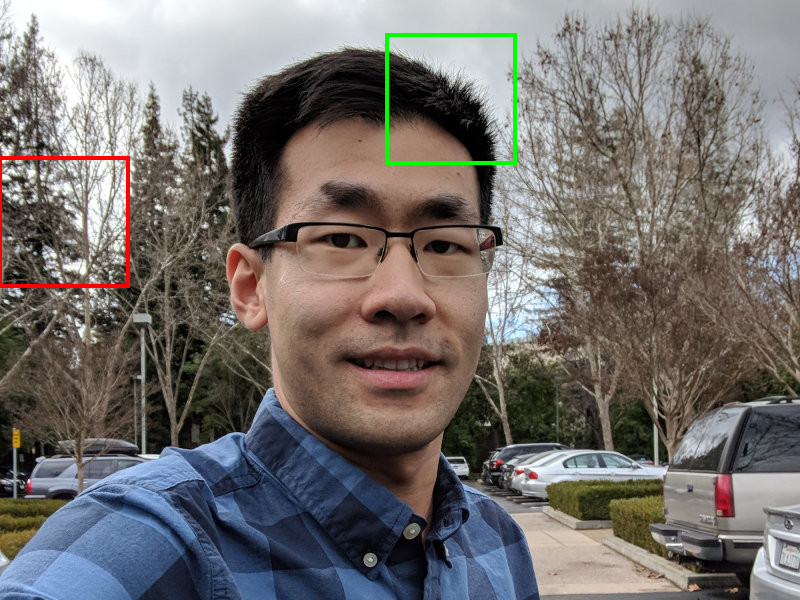}} &
        \multicolumn{2}{c}{\includegraphics[width=\widthcomp\linewidth, trim={0 15px 0 8px},clip]{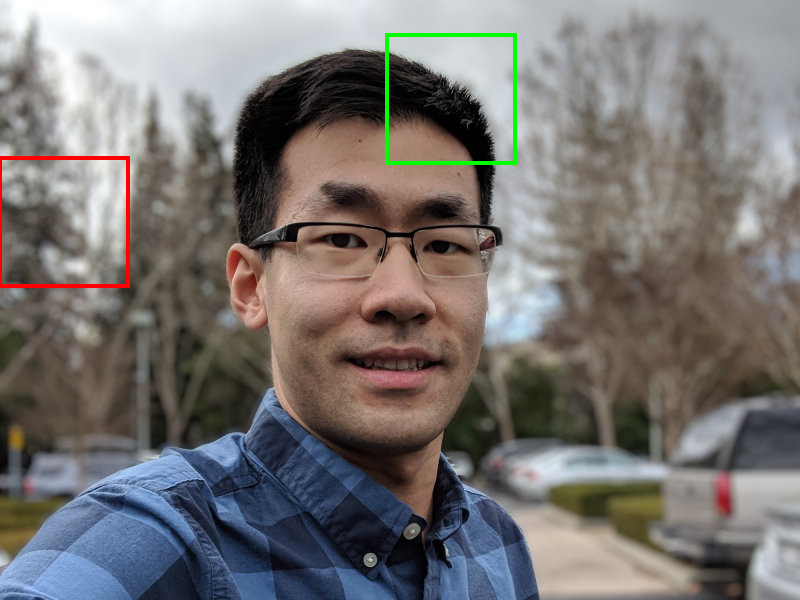}} &
        \multicolumn{2}{c}{\includegraphics[width=\widthcomp\linewidth, trim={0 15px 0 8px},clip]{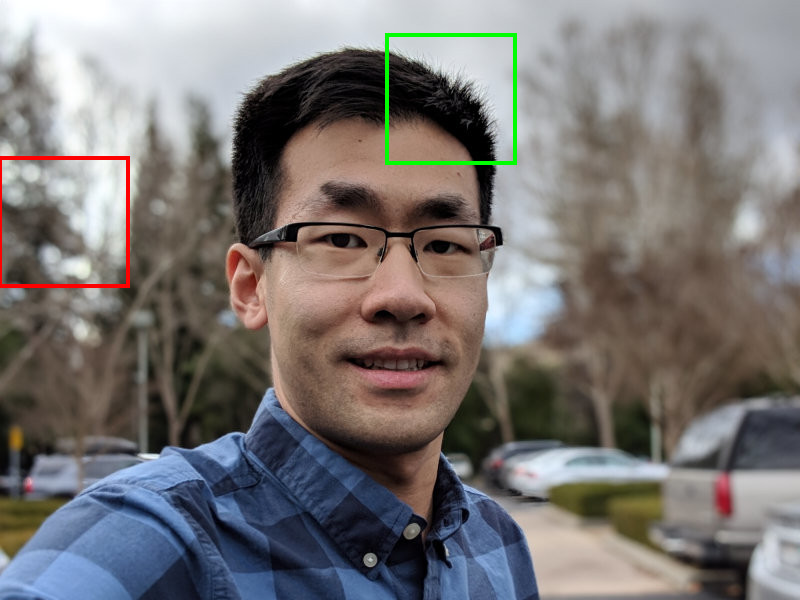}} \\
        \includegraphics[width=\widthcompp\linewidth, trim={0 7px 0 5px},clip]{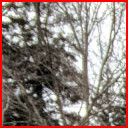} &
        \includegraphics[width=\widthcompp\linewidth, trim={0 7px 0 5px},clip]{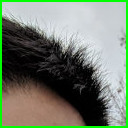} &
        \includegraphics[width=\widthcompp\linewidth, trim={0 7px 0 5px},clip]{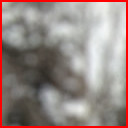} &
        \includegraphics[width=\widthcompp\linewidth, trim={0 7px 0 5px},clip]{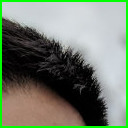} &
        \includegraphics[width=\widthcompp\linewidth, trim={0 7px 0 5px},clip]{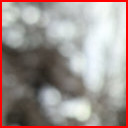} &
        \includegraphics[width=\widthcompp\linewidth, trim={0 7px 0 5px},clip]{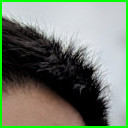} \\
    \end{tabular}
    \vspace{-2mm}
\caption{Comparison on \textbf{real-world smartphone portrait photography~\cite{wadhwa2018synthetic}}. Ours successfully retains fine details such as hair.}
\label{fig:zero-shot-portrait}
\vspace{-4mm}
\end{figure}

\noindent\textbf{Choice of Attention Mechanism: }
We assessed two attention mechanism alternatives in \cref{tab:ablation}.

\noindent \textbf{a)} Swin attention~\cite{liu2021swin} induces local bias through attention on fixed grid windows and shift operations for inter-window dependencies. However, it appears to be unable to model the long-range dependencies needed for Bokeh rendering.

\noindent \textbf{b)} NAT~\cite{Hassani_2023_CVPR} uses adaptive windowed attention based on the position of each query token.
Yet, due to its lack of spatial and aperture awareness, it fails to outperform our AAA.

\noindent\textbf{Aperture Aware Attention Ablation: }
Each ablation phase in \cref{tab:ablation} adds a mechanism to our attention model.

\noindent \textbf{1)} Lack of spatial awareness in the attention drastically reduces performance in the Bokeh rendering task.

\noindent \textbf{2)} A unified decay mask for all heads enhances spatial context and biases attention toward local image regions.

\noindent \textbf{3)} Varying the scale of the attention mask for each head biases them toward specific blur intensities, enhancing performance, particularly at lower \fstops.

\noindent \textbf{4)} Aperture awareness allows bias to be tuned to the required \fstop, leading to a improvement for narrow apertures.
Without this awareness, a trade-off arises regarding the choice of an optimal decay to render varied \fstops.

\subsection{Zero-Shot Evaluation}
\label{subsec:realworld}

Our method further generalizes in the \emph{zero-shot} setting on unseen smartphone images.
In \cref{fig:zero-shot-portrait}, we demonstrate its application to a portrait image.
Our method clearly outperforms Syn-DOF~\cite{wadhwa2018synthetic} (Google Pixel Portrait mode), which was tailored to use the DualPixel data from the Google Pixel Phone.
Ours achieves more accurate subject separation, without using auxiliary data or models. 
Additional comparisons are provided in the supplemental material.

\subsection{Further Applications}
\label{subsec:beyond}

\begin{figure}
    \centering
    \footnotesize
    \setlength{\tabcolsep}{0.5pt}
    \renewcommand{\arraystretch}{0.3}
    \def\widthcomp{0.328}
    \def\widthcompp{0.1618}
    \begin{tabular}{cccccc}
        \multicolumn{2}{c}{Input} & \multicolumn{2}{c}{DRBNet~\cite{ruan2022learning}} & \multicolumn{2}{c}{\textbf{Ours}} \\
        \addlinespace[0.5pt]
        \multicolumn{2}{c}{\includegraphics[width=\widthcomp\linewidth, trim={0 100px 0 20px},clip]{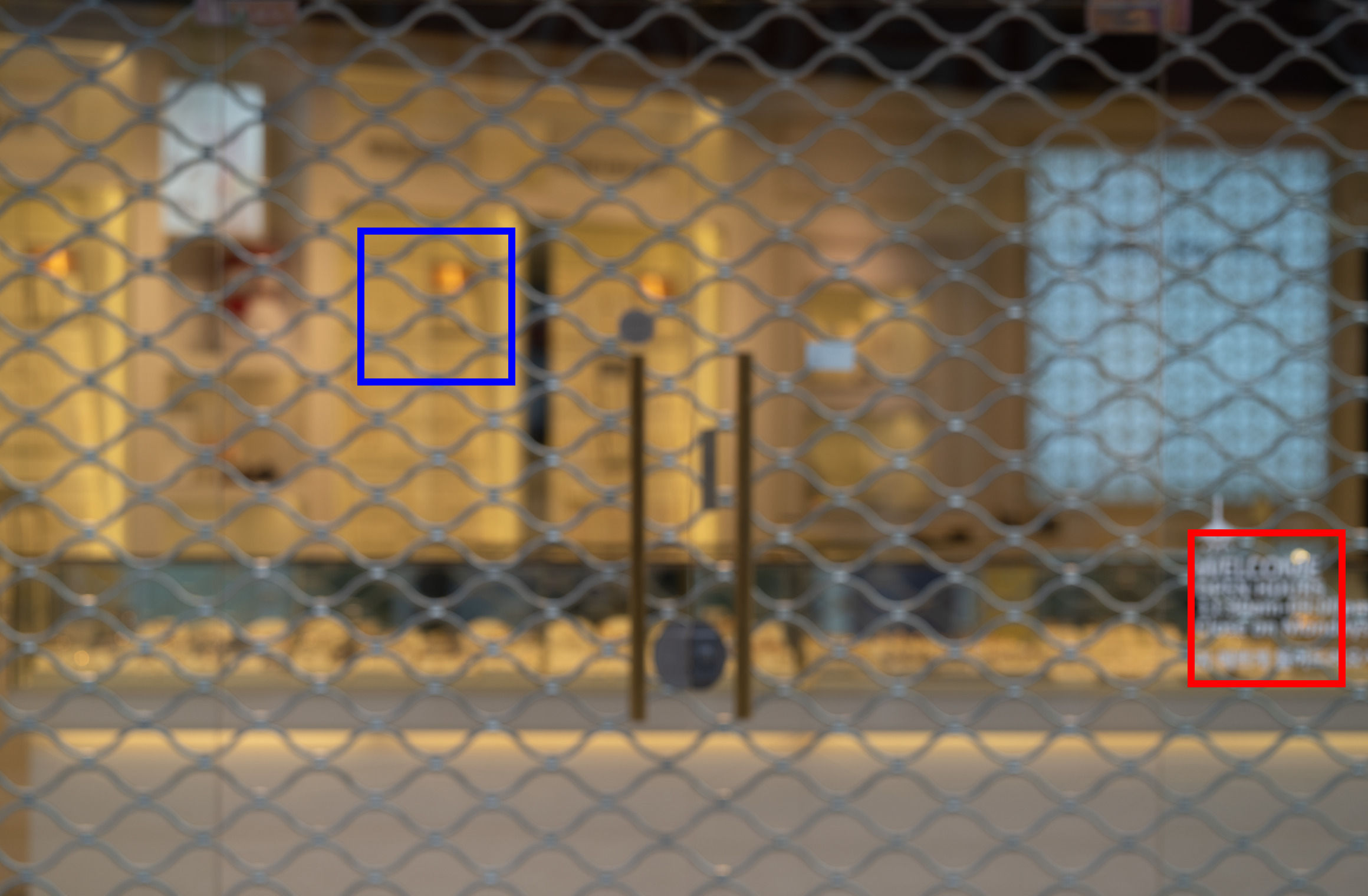}} &
        \multicolumn{2}{c}{\includegraphics[width=\widthcomp\linewidth, trim={0 100px 0 20px},clip]{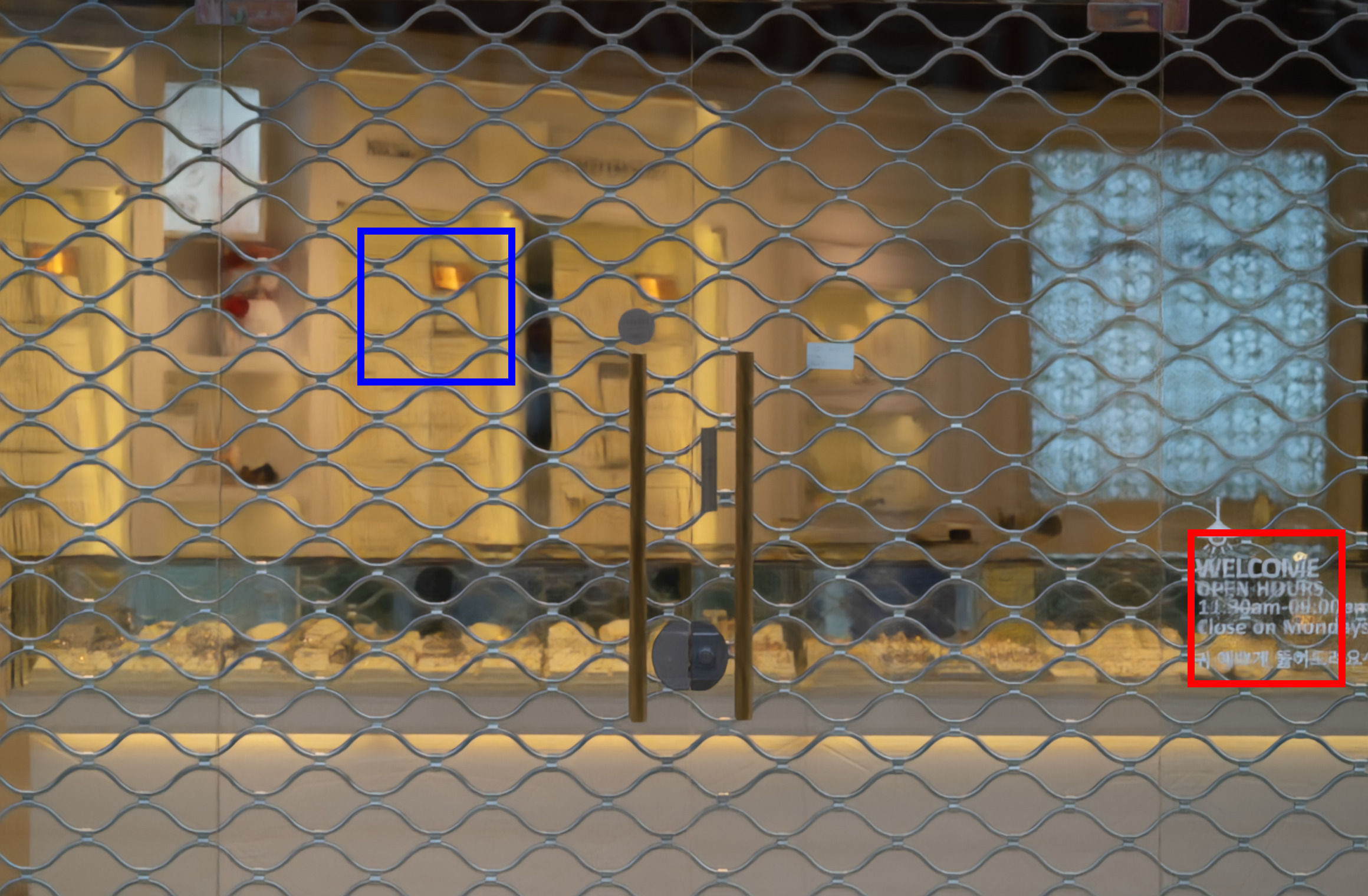}} &
        \multicolumn{2}{c}{\includegraphics[width=\widthcomp\linewidth, trim={0 100px 0 20px},clip]{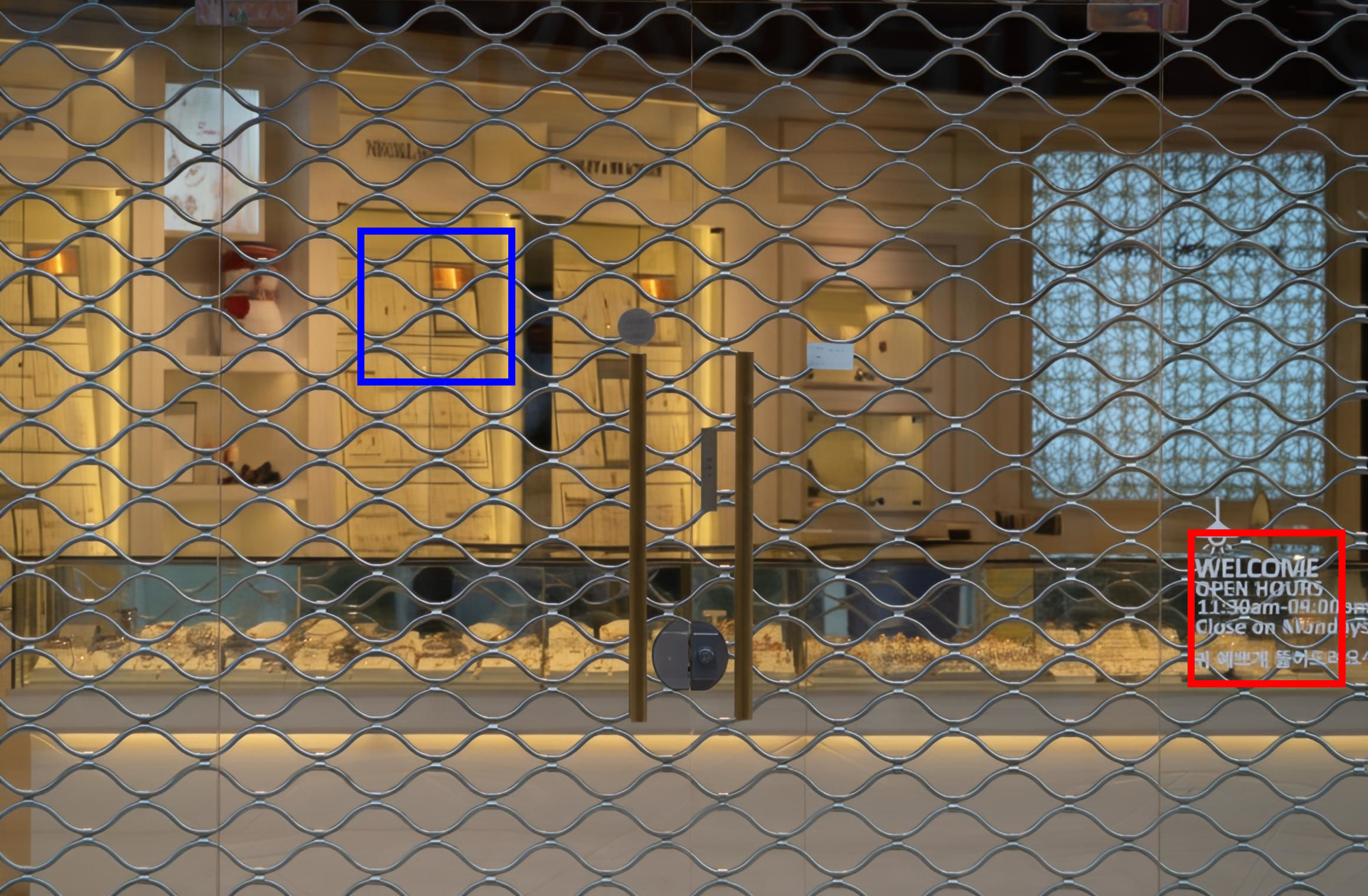}} \\
        \includegraphics[width=\widthcompp\linewidth]{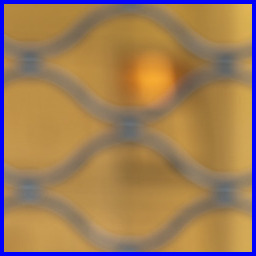} &
        \includegraphics[width=\widthcompp\linewidth]{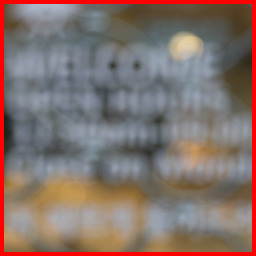} &
        \includegraphics[width=\widthcompp\linewidth]{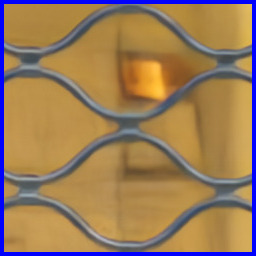} &
        \includegraphics[width=\widthcompp\linewidth]{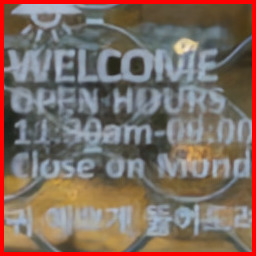} &
        \includegraphics[width=\widthcompp\linewidth]{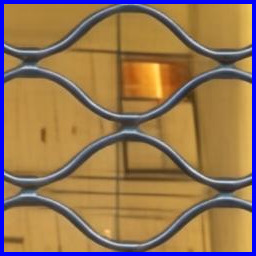} &
        \includegraphics[width=\widthcompp\linewidth]{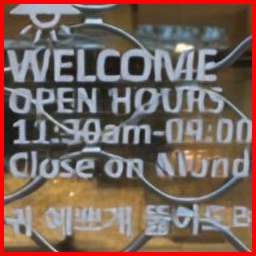} \\
    \end{tabular}
    \vspace{-2mm}
\caption{Visual Comparison on Real DOF~\cite{abuolaim2020defocus}. Ours achieves higher visual clarity while reconstructing fine background details.}
\vspace{-2mm}
\label{fig:RealDOFQual}
\end{figure}

In single view \textit{defocus deblurring}, the goal is to reconstruct sharp image areas blurred by the Bokeh effect.
Intuitively, this is the inverse problem to Bokeh rendering.
Similarly, large-scale public training datasets for defocus deblurring are not available, the largest contender being DPDD~\cite{abuolaim2020defocus} with only 500 scenes.
Conveniently, we can inverse \dataset to create RealDefocus with 4400 scenes, additional defocus intensity variation boosting its total size to 23,000 pairs.
Our proposal notably contains samples with a much stronger defocus effect than DPDD, greatly increasing the difficulty of the reconstruction process.
Trained on RealDefocus, our architecture performs competitively in zero-shot evaluation on the RealDOF~\cite{abuolaim2020defocus} defocus deblurring benchmark \cref{tab:RealDOF}. 
The visual comparison with the top performing DRBNet~\cite{ruan2022learning} method in \cref{fig:RealDOFQual} shows that we are able to render a much clearer image.

Although a deeper study of RealDefocus, its impact on single-view \textit{defocus deblurring} methods, and the link between defocus deblurring and Bokeh Rendering would be of interest, it is beyond our current scope.

\begin{table}[]
        \centering
        \renewcommand{\arraystretch}{0.8}
        \footnotesize
            \begin{tabular}{l|ccc}
            \toprule
            Method                                  & PSNR$\uparrow$ & SSIM$\uparrow$ & LPIPS$\downarrow$ \\
            \midrule
            Input                                   & 22.333         & 0.633          & 0.524  \\
            \midrule
            DPDNet$_S$~\cite{abuolaim2020defocus}   & 22.870         & 0.670          & 0.425 \\
            AIFNet~\cite{ruan2021aifnet}            & 23.093         & 0.680          & 0.413  \\
            IFANet~\cite{lee2021iterative}          & 24.709         & 0.749          & 0.306 \\
            KPAC~\cite{son2021single}               & 23.984         & 0.716          & 0.336 \\
            DRBNet~\cite{ruan2022learning}          & \textit{25.745}& \textit{0.771} & \textit{0.257} \\
            \midrule
            Ours-L (Deblur)                     & \textbf{25.858} & \textbf{0.797} & \textbf{0.205} \\
            \bottomrule
            \end{tabular}
            \vspace{-1mm}
            \caption{Results on RealDOF~\cite{abuolaim2020defocus}. Note that training data varies.}
            \vspace{-2mm}
        \label{tab:RealDOF}
\end{table}

\subsection{Limitations}

The main limitation of our Bokeh rendering method is that its focus is always aligned with the optical focus point of the input image, similar to previous one-step neural approaches~\cite{ignatov2020aim, ignatov2020rendering, nagasubramaniam2023BEViT, dutta2021stackedbokeh, BokehGlassGAN}.
This means that unlike some previous multi-step rendering frameworks, introduced in \cref{sec:related}, our method can not change the focus subject.

\begin{figure}
    \centering
    \footnotesize
    \setlength{\tabcolsep}{0.2pt}
    \renewcommand{\arraystretch}{0.15}
    \def\widthcomp{0.248}
    \def\widthcompp{0.1235}
    \begin{tabular}{cccccccc}
        \multicolumn{2}{c}{Input} & \multicolumn{2}{c}{BoMe~\cite{BokehMeHybrid}} & \multicolumn{2}{c}{\textbf{Ours+BoMe~\cite{BokehMeHybrid}}} & \multicolumn{2}{c}{DC2~\cite{alzayer2023dc2}} \\
        \addlinespace[0.5pt]
        \multicolumn{2}{c}{\includegraphics[width=\widthcomp\linewidth]{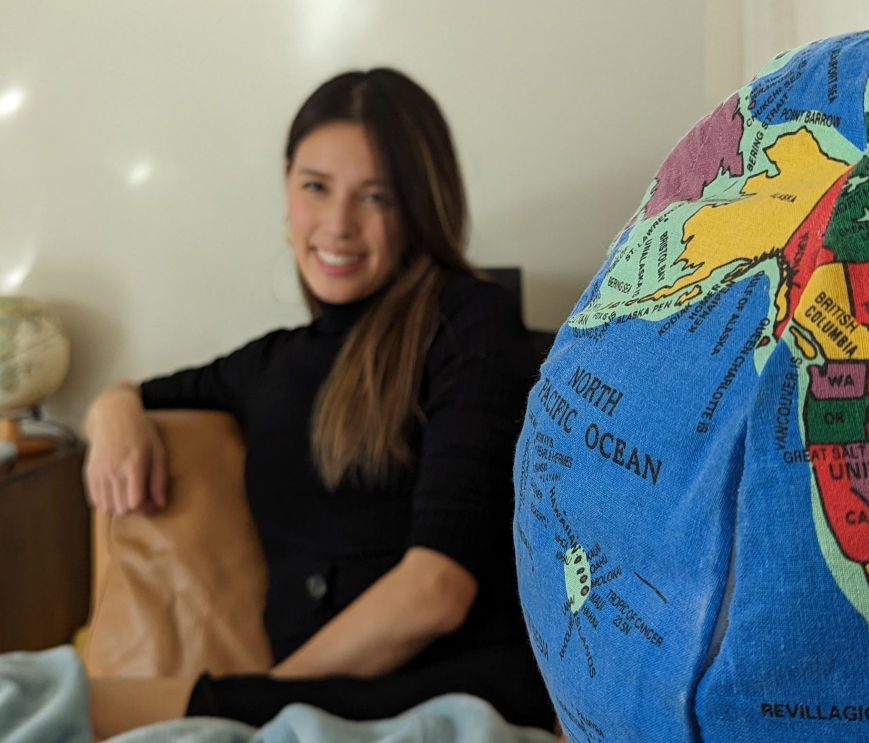}} &
        \multicolumn{2}{c}{\includegraphics[width=\widthcomp\linewidth]{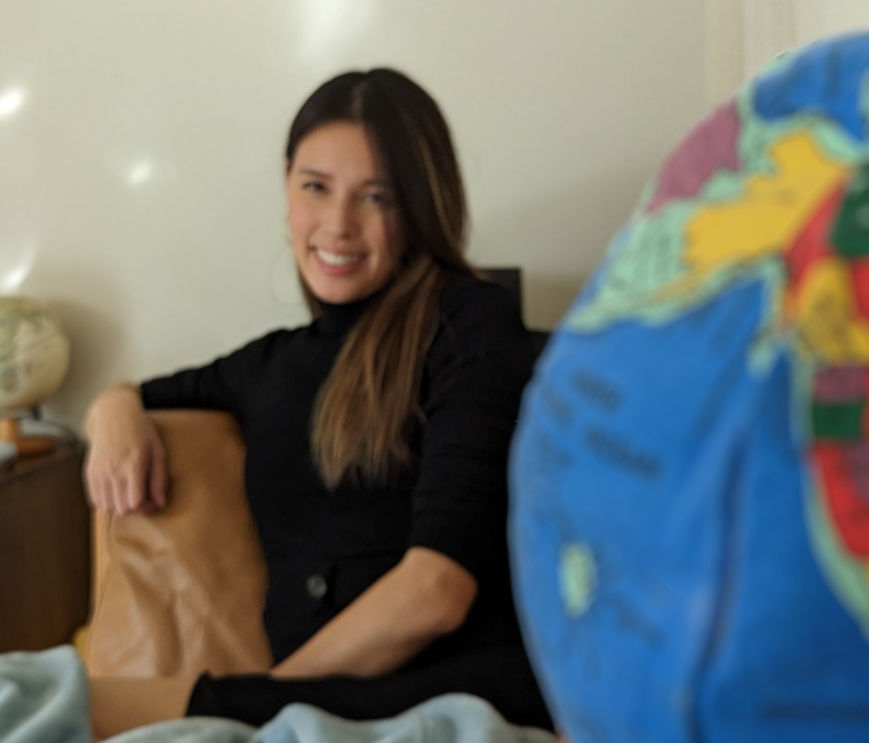}} &
        \multicolumn{2}{c}{\includegraphics[width=\widthcomp\linewidth]{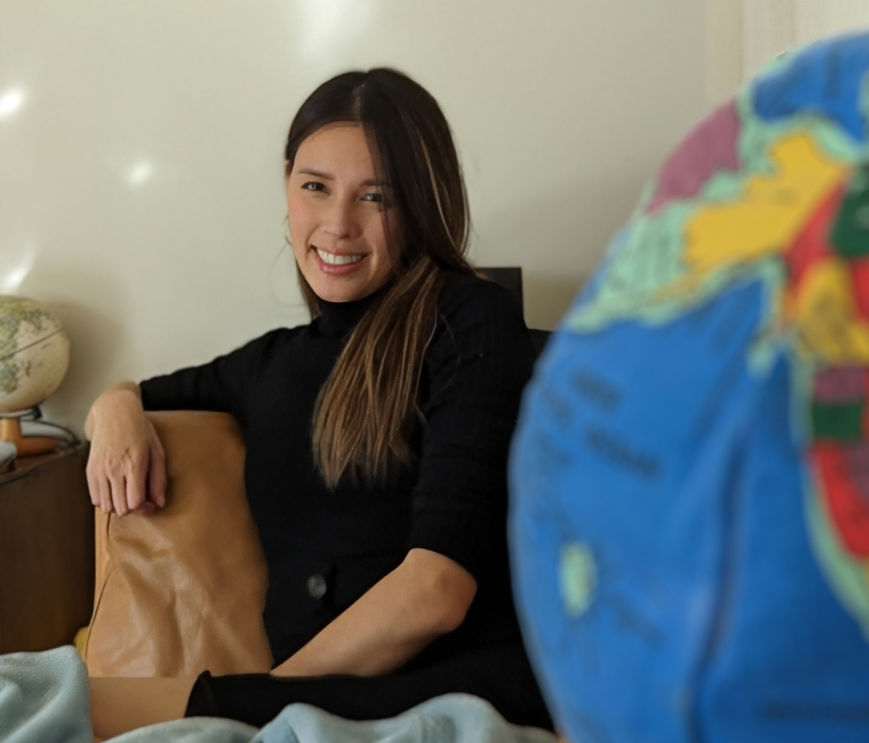}} &
        \multicolumn{2}{c}{\includegraphics[width=\widthcomp\linewidth]{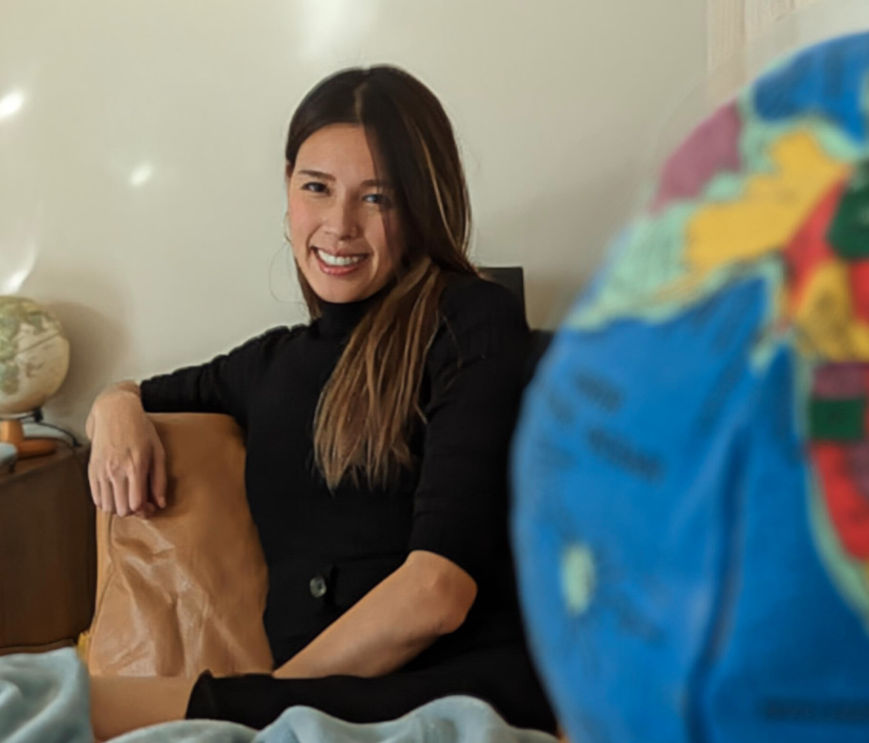}} \\
        \includegraphics[width=\widthcompp\linewidth, ]{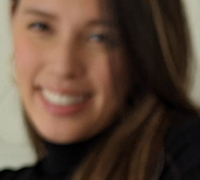} &
        \includegraphics[width=\widthcompp\linewidth, ]{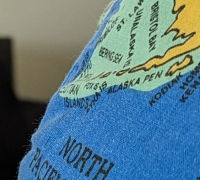} &
        \includegraphics[width=\widthcompp\linewidth, ]{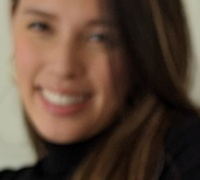} &
        \includegraphics[width=\widthcompp\linewidth, ]{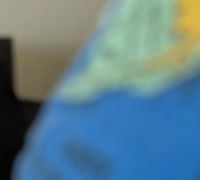} &
        \includegraphics[width=\widthcompp\linewidth, ]{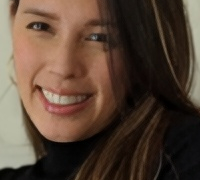} &
        \includegraphics[width=\widthcompp\linewidth, ]{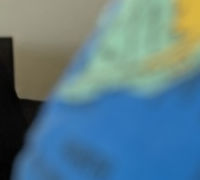} &
        \includegraphics[width=\widthcompp\linewidth, ]{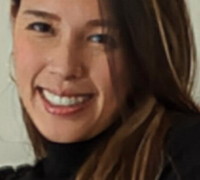} &
        \includegraphics[width=\widthcompp\linewidth, ]{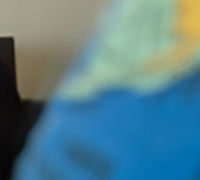} \\
    \end{tabular}
    \vspace{-2mm}
\caption{Demonstration of \textit{refocusing}. Note that our approach shows less unnatural sharpening artifacts in the face while showing less color bleeding artifacts around the blurred image region.}
\vspace{-4mm}
\label{fig:refocus}
\end{figure}

Although these methods control the focus of Bokeh synthesis, they are incapable of actually \textit{refocusing}, limiting their application in the real image domain~\cite{alzayer2023dc2}.
So, if given an image with two subjects, such as \cref{fig:refocus}, a method like BoMe~\cite{BokehMeHybrid} can render Bokeh according to a subject, but not recover its sharpness.
In \cref{fig:refocus} we integrate our \textit{defocus deblurring} model into BoMe as an additional pre-processing step to obtain a true image \textit{refocusing} framework.
The result compares favorably with the DC2~\cite{alzayer2023dc2} framework that relies on multi-camera fusion.
Unfortunately, a deeper comparison is impossible since both data and code for DC2 remain private~\cite{alzayer2023dc2}.

A new open-source real-world dataset and method for this under-explored refocusing task would be helpful to the community.
However, as the focus of our work is on the Bokeh rendering task, this is outside our current scope.

\section{Conclusion}
This work addresses important gaps in Bokeh rendering through two major contributions. 
First, our RealBokeh dataset, comprising a significant number of high-resolution images with diverse aperture and focal length settings, providing a robust foundation for training Bokeh rendering models. 
Second, our Bokehlicious architecture, featuring an innovative aperture-aware attention mechanism, demonstrates that efficient Bokeh rendering with aperture control is achievable without compromising quality or relying on auxiliary data. 
Our strong performance across multiple benchmarks, zero-shot generalization, and defocus deblurring validates both the utility of our dataset and our architectural choices. 
By making our code and dataset publicly available, we aim to promote progress in Bokeh rendering and enable new applications in computational photography.
{
    \small
    \bibliographystyle{ieeenat_fullname}
    \bibliography{main}
}

\end{document}